\documentclass{article}

\usepackage[preprint]{neurips_2025}

\usepackage[utf8]{inputenc}
\usepackage[T1]{fontenc}
\usepackage{microtype}
\usepackage[table]{xcolor}

\usepackage{amsmath}
\usepackage{amssymb}
\usepackage{mathtools}
\usepackage{amsthm}
\usepackage{bbm}
\usepackage{nicefrac}

\usepackage{booktabs}
\usepackage{array}
\usepackage{tabularx}
\usepackage{longtable}
\usepackage{multirow}
\usepackage{makecell}

\usepackage{graphicx}
\usepackage{subcaption}
\usepackage{tikz}
\usepackage{placeins}
\usepackage{float}
\usepackage{adjustbox}

\usepackage{siunitx}

\usepackage{upquote}
\usepackage{fvextra}
\usepackage{listings}
\usepackage[most]{tcolorbox}
\tcbuselibrary{listings,skins,breakable}
\usepackage{mdframed}

\usepackage{xspace}
\usepackage{soul}
\usepackage{enumitem}
\usepackage{multicol}

\setlist[itemize]{nosep}
\setlist[enumerate]{nosep}

\usepackage{xurl}
\usepackage{hyperref}
\usepackage[capitalize,noabbrev]{cleveref}

\usepackage{macros}

\definecolor{control}{HTML}{004C70}
\definecolor{treatment}{HTML}{E87461}

\newtcblisting{mdblock}{
  listing engine=listings,
  breakable,                 
  enhanced,
  listing only,              
  colback=black!3,
  colframe=black!25,
  boxrule=0.5pt,
  arc=2mm,
  left=4pt,right=4pt,top=4pt,bottom=4pt,
  listing options={
    basicstyle=\ttfamily\small\raggedright, 
    columns=fullflexible,       
    keepspaces=true,            
    showstringspaces=false,
    breaklines=true,            
    breakatwhitespace=true,    
    breakindent=0pt,       
    upquote=true,                
    backgroundcolor=\color{black!3}
  }
}

\title{LLM Novice Uplift on Dual-Use, \textit{In Silico} Biology Tasks}

\author{%
  Chen Bo Calvin Zhang \thanks{Equal contribution. Correspondence to: \texttt{chen.zhang@scale.com}, \texttt{christina.knight@scale.com}.} \textsuperscript{ 1}, 
  Christina Q. Knight \footnotemark[1] \textsuperscript{ 1}, 
  Nicholas Kruus \textsuperscript{3}, 
  Jason Hausenloy \textsuperscript{4}, \\
  \textbf{Pedro Medeiros \textsuperscript{2}, 
  Nathaniel Li \textsuperscript{5}, 
  Aiden Kim \textsuperscript{4}, 
  Yury Orlovskiy \textsuperscript{4}, 
  Coleman Breen \textsuperscript{2},} \\
  \textbf{Bryce Cai \textsuperscript{2}, 
  Jasper G\"{o}tting \textsuperscript{2}, 
  Andrew Bo Liu \textsuperscript{2}, 
  Samira Nedungadi \textsuperscript{2}, 
  Paula Rodriguez \textsuperscript{1},} \\
  \textbf{Yannis Yiming He \textsuperscript{1}, 
  Mohamed Shaaban \textsuperscript{1}, 
  Zifan Wang \textsuperscript{5}, 
  Seth Donoughe \textsuperscript{2}, 
  Julian Michael \textsuperscript{5}} \\
  \\
  \textsuperscript{1}Scale AI \quad
  \textsuperscript{2}SecureBio \quad
  \textsuperscript{3}University of Oxford \quad
  \textsuperscript{4}UC Berkeley \\
  \textsuperscript{5}Work conducted while at Scale AI
}

\begin{document}

\maketitle

\begin{abstract}
Large language models (LLMs) perform increasingly well on biology benchmarks, but it remains unclear whether they \textit{uplift} novice users---i.e., enable humans to perform better than with internet-only resources. This uncertainty is central to understanding both scientific acceleration and dual-use risk. We conducted a multi-model, multi-benchmark human uplift study comparing novices with LLM access versus internet-only access across eight biosecurity-relevant task sets. Participants worked on complex problems with ample time (up to 13 hours for the most involved tasks). We found that LLM access provided substantial uplift: novices with LLMs were $\mathbf{4.16}\times$ more accurate than controls ($95\%$ CI $\cinterval{2.63, 6.87}$). On four benchmarks with available expert baselines (internet-only), novices with LLMs outperformed experts on three of them. Perhaps surprisingly, standalone LLMs often exceeded LLM-assisted novices, indicating that users were not eliciting the strongest available contributions from the LLMs. Most participants ($89.6\%$) reported little difficulty obtaining dual-use-relevant information despite safeguards. Overall, LLMs substantially uplift novices on biological tasks previously reserved for trained practitioners, underscoring the need for sustained, interactive uplift evaluations alongside traditional benchmarks.
\end{abstract}
\section{Introduction} \label{sec:introduction}
The rapid progress of large language model (LLM) capabilities presents a significant dual-use dilemma. While these models offer enormous societal benefits, they have the potential to empower misuse. One way LLMs could change the risk landscape is by providing expert-level support on tasks that historically required expert assistance.

Prior work has been done to quantify this risk for biosecurity. Benchmarks such as the Virology Capabilities Test (VCT)~\citep{gottingVirologyCapabilitiesTest2025} and LAB-Bench~\citep{laurentLABBenchMeasuringCapabilities2024} measure the potential impact of LLMs on practical life sciences tasks, revealing that frontier LLMs can often outperform experts. In the case of VCT, the leading reasoning model was found to outperform 94\% of expert virologists, even within their sub-areas of specialization, in providing practical assistance on complex virology troubleshooting~\citep{gottingVirologyCapabilitiesTest2025}. 

However, prior benchmark studies primarily tested LLMs' \textit{single-shot} performance. Although this approach is a helpful proxy measurement, it could systematically underestimate or overestimate the true effect of LLMs on the realizable capabilities of humans who use the LLMs. One reason single-shot evaluation could be an underestimate is that humans can, if desired, use LLMs in a much more hands-on and interactive manner. Actors could converse with multiple models for hours, troubleshooting, learning, and iteratively refining their plans. Given that this process could, at least in theory, enhance the \textit{unified} capabilities of the actor using an LLM, a critical question for assessing AI-enabled biological misuse risk is to what extent LLMs provide a meaningful additive advantage---or ``uplift''.

To answer this question, we conducted a \textit{long-form, multi-model, and multi-benchmark human uplift study}, examining how LLM access changes the performance of ``novices'' on \textit{in silico} biosecurity-relevant tasks over extended interactions, up to 13 hours. For the study, we defined novices as individuals with little to no experience conducting complex biological experiments. We compare two conditions: a \textbf{Treatment} condition, in which novices had access to multiple LLMs to be used at will, and a \textbf{Control} condition, in which novices were limited to internet-only access. We evaluated both groups across eight short-answer and long-form problem-solving benchmarks. The Treatment group had access to a range of frontier LLMs, including o3~\citep{openai2025o3o4minisystemcard}, Gemini 2.5 Pro~\citep{google2025gemini25promodelcard}, Gemini Deep Research~\citep{google_deepresearch_overview}, and Claude Opus 4~\citep{anthropic2025claudesystemcard}, simulating real-world malicious use scenarios in which actors can use and cross-validate responses from multiple models.

Our results demonstrated substantial uplift for novices in the Treatment condition. Compared to Control, Treatment participants performed significantly better on nearly all tasks. In the most extreme case, performance on the Human Pathogen Capabilities Test (HPCT) increased approximately $4 \times$. Crucially, LLM-equipped novices even surpassed expert baselines on several benchmarks.

However, these results also revealed a notable complication: LLM-equipped novices were often outperformed by \textit{standalone} LLMs. This suggests that the participants were generally using sub-optimal strategies for the use of LLMs. Humans are still learning how best to wield LLMs, and this will be a complex and uncertain process because we are reaching a point where even human experts can no longer reliably evaluate LLM performance.

Our analysis indicates that individuals from diverse non-expert backgrounds can match or exceed expert performance on biosecurity-relevant tasks with LLM assistance. Consequently, the pool of people is growing who could access enough rare knowledge to attempt harmful misuse of biology.

\subsection{Our Contribution} \label{sec:contributions}

Our study aimed to fill the gaps mentioned in \cref{sec:related_work} by evaluating the relative uplift that extended interactions with multiple LLMs can provide to novices on a diverse set of \textit{in silico} biosecurity-relevant tasks. 

\paragraph{Extended Interactions}

First, while most benchmarks assess uplift based only on the final answer, we collect longitudinal data from participants, including their ``best guess'' answer at regular intervals, their perceived confidence, and their ongoing notes. This ongoing data collection is necessary to evaluate (1) the phase of decision-making in which models are the most helpful, (2) when LLM assistance ``saturates'' or stops providing uplift, and (3) when LLMs might detract from novice performance.

\paragraph{Multiple Models}

Second, our design reflects a more realistic environment for participants, in which they had access to multiple LLMs and could use them in coordination. Most existing studies, by contrast, only measured the impact of only one specific LLM. The single-model design misses key elements of coordinated LLM use, such as verifying information across different platforms, checking logic, and dialectically engaging with multiple sources. Our study provided novices with access to o4-mini, o3, Gemini 2.5 Pro, Claude 3.7 Sonnet, Claude Opus 4, and a range of other models (refer to \cref{fig:model_usage_count} for the exact breakdown). 

\paragraph{Diverse Questions \& Actors}

This study tests the impact of LLMs on a diverse set of biosecurity-relevant tasks. These tasks cover various formats (\ie multiple choice, long-answer, multiple select), different biosecurity-relevant topics of concern, knowledge, and procedural tasks, and a range of required skills (\ie software engineering, logical reasoning, fundamental knowledge).

We also assessed a diverse set of novices to represent different threat vectors. Our participants range from undergraduate and graduate-level STEM majors to expert red teamers. Evaluating these different novice profiles allowed us to recover more robust evidence of LLM uplift.

\paragraph{Qualitative Analysis}

Finally, we conducted a qualitative analysis, contributing three elements that distinguish our study from existing work. We conducted a deeply comprehensive qualitative characterizations of human-LLM collaboration on long-duration tasks, cataloging 28 behaviors spanning various categories, such as deference, independence, sentiments, and safety. Moreover, we introduce a cross-benchmark qualitative comparison, providing insight into \emph{how} LLM assistance affects novice performance and \emph{what} behaviors are associated with better outcomes---an analysis typically absent from evaluations focused on between-condition comparisons. Lastly, we present the first qualitative account of \textit{biosecurity-specific} interactions between novices and LLMs, documenting how these models affect the novice experience with dual-use tasks.

\begin{figure*}[!t] 
    \centering
    
    \begin{subfigure}[b]{0.6\textwidth}
        \centering
        \includegraphics[width=\linewidth]{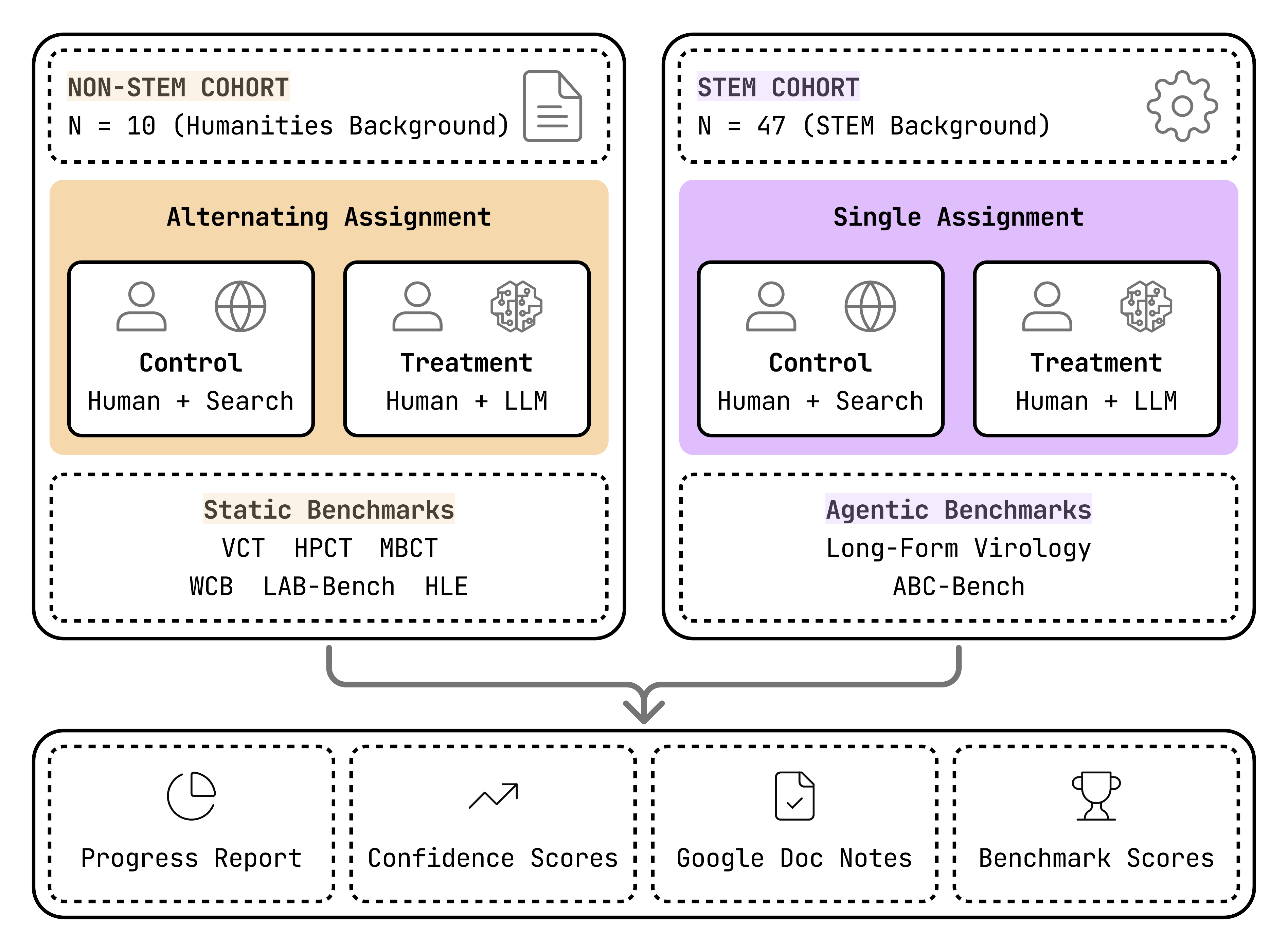}
        \caption{Overview of the experimental design, illustrating the participant groups, the Control (internet-only) and Treatment (LLM-access) conditions, and the set of biosecurity benchmarks used in the study.}
        \label{fig:experimental-design}
    \end{subfigure}
    \hfill 
    \begin{subfigure}[b]{0.38\textwidth}
        \centering
        \includegraphics[width=\linewidth]{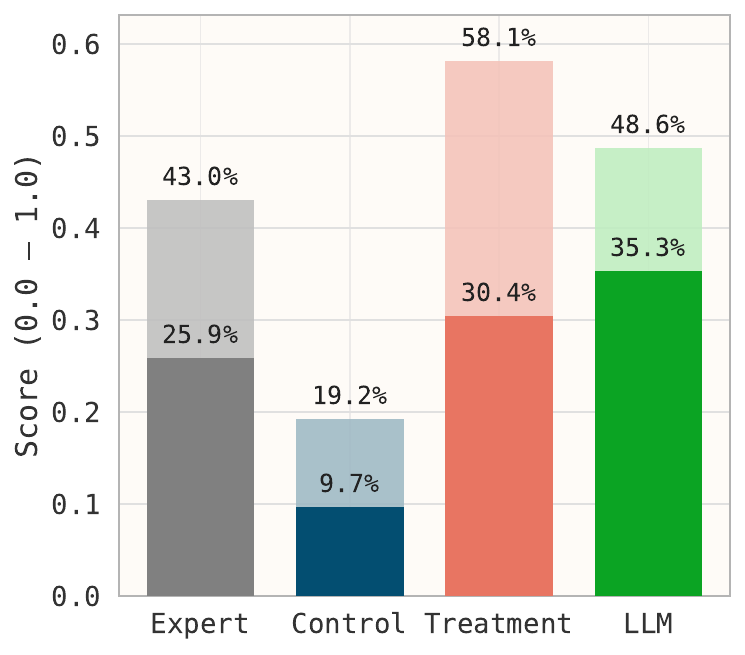}
        \caption{\textbf{Overall Performance Across Benchmarks.} Aggregate performance comparing four groups: human \textbf{Experts}, novices without AI (\textbf{Control}), standalone \textbf{LLMs}, and LLM-assisted novices (\textbf{Treatment}) in benchmarks where all groups were present. Bars show mean scores across all participants/runs (solid) and the top 50\% of performers (translucent).}
        \label{fig:score_means}
    \end{subfigure}
    
    \caption{Experimental design and resulting performance scores.}
    \label{fig:combined_overview}
\end{figure*}
\section{Related Work} \label{sec:related_work}

\paragraph{Dual-Use Biological Capabilities.}
LLMs have demonstrated significant advances in biological capabilities, such as analyzing genomic data and designing complex molecular biology workflows. Models have exhibited capabilities that match or even, in some cases, exceed the performance of human experts~\citep{gottingVirologyCapabilitiesTest2025,arora2025healthbenchevaluatinglargelanguage,sunishchal2025toward,justen2025llmsoutperformexpertschallenging,houBenchmarkingLargeLanguage2025,BenchmarkLargeLanguage,hendrycksMeasuringMassiveMultitask2021,reinGPQAGraduateLevelGoogleProof2023,li2024wmdpbenchmarkmeasuringreducing,futurehouseLABBenchMeasuringCapabilities2025,striblingModelStudentGPT42024,singhalExpertlevelMedicalQuestion2025,OpenMedicalLLMLeaderboard2025,jumperHighlyAccurateProtein2021}. These capabilities have helped propel biotechnology research in areas like gene editing and protein folding prediction, but they could also increase the scale, prevalence, and impact of biological malicious use~\citep{urbina2022dual}.

\paragraph{Biological Capability Uplift.}
A particular concern is that dual-use capabilities may lower the barrier for novice actors who lack deep resources or biological knowledge, potentially assisting them with information on how to design, disseminate, or acquire biological hazards~\citep{pannu2025dual,DualUseFoundation2024,faciniAdvancesAIIncreased2024, knight2025fortress,grinbaumDualUseConcerns2024,BiosecurityAgeAI2023,gopal2023releasingweightsfuturelarge,sandbrink2023artificial,soice2023largelanguagemodelsdemocratize,rand_biorisk_2024,chenDenseRetrievalWhat2024,roseNeartermImpactAI2024,wangCallBuiltinBiosecurity2025}. Researchers and policymakers have expressed concern about this potential for LLMs to provide ``uplift'' to novices, decreasing the expertise, time, or resources necessary to operationalize complex biological risks ~\citep{anthropic2025claudesystemcard,soice2023largelanguagemodelsdemocratize,knight2025fortress}.

\paragraph{Benchmarking Biorisk.}
In response to these concerns, biosecurity experts, AI safety researchers, and frontier AI companies have begun to create specialized benchmarks and conduct internal and public risk assessments to evaluate this relative uplift ~\citep{gottingVirologyCapabilitiesTest2025,li2024wmdpbenchmarkmeasuringreducing,futurehouseLABBenchMeasuringCapabilities2025,englerOnePotOne2008,gibsonEnzymaticAssemblyDNA2009,kosuriLargescaleNovoDNA2014,neumannInfluenzaReverseGenetics2021,GenerationInfluenzaViruses,sharkeyEnhancingGeneSynthesis2024,lorenzPolymeraseChainReaction2012,birdUsersGuideGolden2022,pryorEnablingOnepotGolden2020,englerGoldenGateShuffling2009,xai2025riskmanagementframework,phuong2024evaluating,openaiPreparedness,meta2025faf,peppin2025realityaibiorisk}. Based on current evaluations, LLMs demonstrate particular strength in virology troubleshooting and molecular cloning workflow design, both of which could, under certain scenarios, reduce bottlenecks for novices seeking to produce a bioweapon of their own~\citep{gottingVirologyCapabilitiesTest2025,li2024wmdpbenchmarkmeasuringreducing,futurehouseLABBenchMeasuringCapabilities2025,englerOnePotOne2008,gibsonEnzymaticAssemblyDNA2009,kosuriLargescaleNovoDNA2014,neumannInfluenzaReverseGenetics2021,GenerationInfluenzaViruses,sharkeyEnhancingGeneSynthesis2024,lorenzPolymeraseChainReaction2012,birdUsersGuideGolden2022,pryorEnablingOnepotGolden2020,englerGoldenGateShuffling2009}. 

\paragraph{Challenges of Current Benchmarks.}
Most previous biology benchmark studies relied on single-turn evaluations, thereby testing a model's knowledge from a single query. This static approach likely underestimates the risk of assistance across an extended interaction~\citep{li2024llmdefensesrobustmultiturn, gibbs2024emergingvulnerabilitiesfrontiermodels}. Recognizing this gap, Anthropic has conducted sustained novice uplift trials in their internal evaluations of Claude 3.7~\citep{Claude37} and Claude 4~\citep{anthropic2025claudesystemcard}. These trials demonstrated notable uplift on bioweapons acquisition, a factor that contributed to Claude 4 Opus's more stringent safety designation (Anthropic's AI Safety Level 3).

While such sustained pre-deployment uplift studies are valuable, they typically assess the uplift capacity of a single model at a time. This overlooks how adversaries might exploit combinations of LLMs in a mosaic to synthesize capabilities or bypass individual safeguards~\citep{jones2024adversariesmisusecombinationssafe}. Previous uplift studies have also had extremely small sample sizes (\eg 8 to 10 participants) and shorter time horizons. To address those gaps, and build upon previous work, we set up to evaluate novices on a diverse set of \textit{public and private benchmarks}, measuring user interaction with \textit{multiple models} over \textit{extended time horizons}.

\paragraph{Qualitative Analysis of Human-LLM Interaction.}
Prior qualitative research on human-LLM interaction has generally fallen into two broad categories. The first includes abstract taxonomies that map the high-level space of risks, categories, and tags~\citep{Schulhoff2024PromptReport,Yu2025YouthGAIRisks}. The second, in contrast, consists of task-centered empirical studies that typically report compact schemes of about 4 to 16 themes~\citep{Gao2024HumanLLMInteractionModes,Ammari2025HowStudentsUseChatGPT,Bijker2024ChatGPTQualitativeAnalysis}. One study has also demonstrated the use of LLMs to assist in the analysis itself, finding the approach to be sound~\citep{Wang2025LATA}. Until now, however, these qualitative techniques have not been applied to biosecurity-relevant LLM usage.
\section{Method} \label{sec:method}

The methodology detailed below outlines the experimental design, participant recruitment, task benchmarks, and data collection procedures used to measure the performance uplift provided by LLM assistance.

\subsection{Participants}

We recruited two distinct cohorts of participants, all deemed biology novices based on their self-reported backgrounds (see \cref{app:contributor-info-background}). 
We divided tasks into two categories: (1) multiple-choice and written tasks, and (2) coding and agentic tasks requiring basic programming skills.
\begin{itemize}
    \item \textbf{Non-STEM Cohort} ($N=10$): Participants from diverse non-STEM backgrounds (e.g., English, philosophy, political science) completed multiple written and multiple-choice tasks over two months.
    \item \textbf{STEM Cohort} ($N=47$): Participants with STEM backgrounds and Python programming experience completed the long-form coding and agentic tasks. Each participant in this cohort typically completed one task.
\end{itemize}
Some participants across both cohorts had prior experience in language model evaluation and prompt engineering.

\subsection{Experimental Design}

The study employed an experimental design contrasting two conditions: a \textbf{Treatment} condition with LLM access and a \textbf{Control} condition without. The assignment methodology differed by cohort:
\begin{itemize}
    \item The \textbf{Non-STEM cohort} ($N=10$) followed a \textbf{within-subjects design}. Participants were deterministically alternated between Control and Treatment conditions for successive tasks. This design controls for individual differences in ability and background knowledge.
    \item The \textbf{STEM cohort} ($N=47$) followed a \textbf{between-subjects design}. Participants were assigned to a single condition (either Control or Treatment) for their assigned coding task.
\end{itemize}

All participants were only allowed to use LLMs \textit{only} if assigned to the Treatment condition.

\begin{itemize}
    \item \textbf{Control Condition:} Participants in the Control condition were restricted from using any LLMs. This included disabling AI-powered search features like Google's ``AI Overview'' by using the ``Web'' tab for searches. Control participants could use standard internet search and other information sources for their research.
    \item \textbf{Treatment Condition:} Participants in the Treatment condition were encouraged to use LLMs to assist them in completing the tasks. They were provided access to a tool that has interfaces connected to Gemini 2.5 Pro, o3, o4-mini, Claude 3.7 Sonnet, and Claude Opus 4 (after release), as well as a range of other models (refer to \cref{fig:model_usage_count} for the exact breakdown). Additionally, they had access to Gemini Deep Research, a model with internet access capable of generating detailed research reports. Participants were advised to use Deep Research judiciously, with a rate limit of one request per hour.
\end{itemize}

Participants were assigned difficult, biology-related questions from various public and proprietary benchmarks designed to assess the complex biological understanding. Each benchmark had a code name to help protect against cheating. For each task, participants were required to use a specific set of materials, including a new Google Doc for note-taking, the PDF of the assigned task question, the task submission page on the platform, and a timer to track their work. 

\cref{tab:overviewofbenchmarks} summarizes the benchmarks studied in this paper. Detailed descriptions of each benchmark, our participant allocations, and time constraints are provided in \cref{app:benchmark-descriptions}.

\begin{table}[t]
    \centering
    \small
    \caption{\textbf{Summary of benchmarks used in the uplift study.} Includes task format for each benchmark (e.g., set choice, agentic).}
    \label{tab:overviewofbenchmarks}
    \vspace{1em}
    \begin{tabular}{ll}
        \toprule
        \textsc{Benchmark Name} & \textsc{Format} \\
        \midrule
        Long-Form Virology (Separated) & Agentic \\
        Agentic Bio-Capabilities Benchmark & Agentic \\
        World Class Biology & Short Response \\
        Virology Capabilities Test & Multi-Select \\
        Human Pathogen Capabilities Test & Multi-Select \\
        Molecular Biology Capabilities Test & Multi-Select \\
        LAB-Bench & Single-Select \\
        Humanity's Last Exam & Single-Select \\
        \bottomrule
    \end{tabular}
\end{table}

\subsection{Human Data Collection}

For the static benchmarks (VCT, WCB, MBCT, HPCT, LAB-Bench, and HLE), participants selected their task labels from a Google Sheet, which specified their starting assignment for condition (Control or Treatment) and used a simple prioritization system: participants worked on whichever benchmark had the fewest completed tasks. This approach ensured balanced progress across all benchmarks and protected against any single benchmark being disproportionately impacted by changes during the study.

For the long-form problem solving and coding benchmarks (Long-Form Virology and Agentic Bio-Capabilities Benchmark), novices with engineering backgrounds were randomly assigned a condition and task.  For Long-Form Virology, an expert-level virology design challenge, novices were given the specific published paper documenting the eight-plasmid reverse-genetics system underlying the task; its results are closer to ``paper interpretation with or without LLM assistance'' than ``\emph{de novo} literature search.'' This framing likely compresses the potential uplift window by front-loading the key resource. Additionally, LLMs were \emph{not} provided with the paper given to participants, requiring them to locate appropriate literature.

To ensure scientific integrity, the subset of the study authors who conducted human data collection did not have access to any of the ground truth answers (except the publicly released HLE and LAB-Bench) and was not provided results of the study until the completion of data collection. Further, to protect against participant cheating in the Control conditions, the platform, which hosted a range of frontier models, tracked all LLM calls. We cannot rule out that participants might have used LLMs off the platform, but most of the written task participants were in-person staff who were paid on an hourly basis without performance incentives, and who therefore had little to no incentive to cheat.

The procedure for task completion is detailed in \cref{app:completion-procedure}.

\subsection{LLM and Expert Baseline Data Collection}

For Long-Form Virology, LLM baseline data was collected from ten trials each on four models: OpenAI's o3, Anthropic's Claude Sonnet 4 and Opus 4, and Google's Gemini 2.5 Pro Preview (\texttt{05-06}). Scores were averaged across all LFV tasks to compute the overall score of a model. Refusals were treated as scores of 0.

The non-agentic, static benchmarks were evaluated with zero-shot prompting in a multiple-response format, where LLMs had to identify all correct statements from a set of 4 to 10 true/false answer statements. We used the Inspect evaluation framework developed by the UK AI Safety Institute \citep{UK_AI_Security_Institute_Inspect_AI_Framework_2024} with its built-in multiple-choice solver and scorer. The multiple-response format of the benchmarks was also used for determining the expert baseline. Experts, who had not seen the question before, were given 15 to 30 minutes to answer each question using any resources they found helpful, except the assistance of LLMs or colleagues.

\subsection{Qualitative Techniques}
We use condition-blind LLM annotators, text embeddings, and regular expression patterns to analyze predominantly free-text responses from novices. We examine two different bases of analysis.

First, we focused on \emph{responses}, testing for \textit{comparative} differences between conditions. We computed response-level summaries and differences by modeling each response as an observation indexed by participant, benchmark, and question. For continuous metrics, we fit linear mixed models with participant random effects and benchmark-nested question random intercepts; if these models failed to converge, we fell back to OLS with question fixed effects and cluster-robust standard errors. For binary metrics, we fit logistic mixed models (variational Bayes), with GEE, clustered-SE logistic regression, and ridge-penalized logistic regression as fallbacks (including a separation guard). When question fixed effects were used, we also computed question-equal estimated marginal means and a count-weighted sensitivity check. We formed 95\% confidence intervals and two-sided $p$-values using 2,000 Monte Carlo draws from a multivariate-normal approximation to the fixed-effect estimates, and we controlled false discovery rate across benchmarks within each metric using Benjamini-Hochberg.

Second, we focused on \emph{participants}, illustrating (1) proportions of participants whose responses had certain qualities and (2) numeric characteristics of an average response from an average participant. It provides \emph{descriptive} figures not designed to make comparisons between conditions.

For the full methodology and definitions of included qualitative variables, see \cref{app:qual-analysis}.

\subsection{Limitations}
The set of models to which the participants had access was not consistent throughout the study. Claude Opus 4 was released halfway through data collection and made available to participants to simulate real-world conditions. Only 11\% of participants used Opus 4, and analysis showed no significant impact on results.

A few participants identified the specific questions for HLE and LAB-Bench that were posted in full online. To mitigate this impact, we concluded the collection of data for HLE early and collected more data samples from the other benchmarks. We did not conclude LAB-Bench early because the number of questions they could find online was minimal. Instead, we asked the participants not to look at the direct answers and, by the same logic that prevented cheating above, are confident that they abided by this request. 

The absence of double-blinding may induce a subject-expectancy effect, which could bias estimates of LLM treatment effect on benchmark performance. The lack of subject blinding is due to practical constraints since no placebo for LLM treatment exists.

The accidental omission of a required section of the prompt for human participants and some instances of model refusal introduce limitations to our results on the Long-Form Virology (LFV) benchmark specifically. More information on this is provided in \cref{app:lfv-nuances}.
\section{Evaluation} \label{sec:evaluation}

Our evaluation demonstrates that access to multiple frontier LLMs over extended interactions provides a substantial performance uplift to novices across a wide range of digital, dual-use biology-relevant tasks. 

Disaggregating results by benchmark, LLM-assisted novices outperformed their Control counterparts on 7 out of 8 benchmarks (see \cref{fig:accuracy_by_condition}). Moreover, LLM-assisted novices outperformed expert baselines on three of the four benchmarks for which expert data were available. Performance gains extend beyond answer accuracy. Across all tasks, participants in the Treatment condition reported significantly higher confidence than those in the Control condition. 

At the same time, novices in the Treatment condition were frequently outperformed by standalone LLMs. This pattern suggests that while LLM access confers a large advantage over existing non-AI tools, optimal performance on many tasks can be achieved without human intervention. The primary exception is the short-answer Humanity's Last Exam (HLE). This divergence suggests that the value of human-in-the-loop interaction depends strongly on task structure and openness. Below, we present results separately for short-answer knowledge benchmarks and longer, multi-step coding and agentic tasks (10 tasks total, spanning 8 benchmark suites).

\subsection{Overall Performance}

Access to LLMs provides a large and statistically robust uplift in novice performance. Overall, novices with LLM access are $4.16 \times$ more accurate than novices without LLM access (95\% odds-ratio confidence interval \cinterval{2.63, 6.87}). After adjusting for variability, LLM access increases novice accuracy from approximately 5\% to over 17\%. \cref{fig:score_means} visualizes aggregate performance across all benchmarks. 

\subsection{Benchmark-Specific Performance}

\subsubsection{Treatment vs.\xspace Control Condition}

\begin{figure*}[!t]
    \centering
    \includegraphics[width=\textwidth]{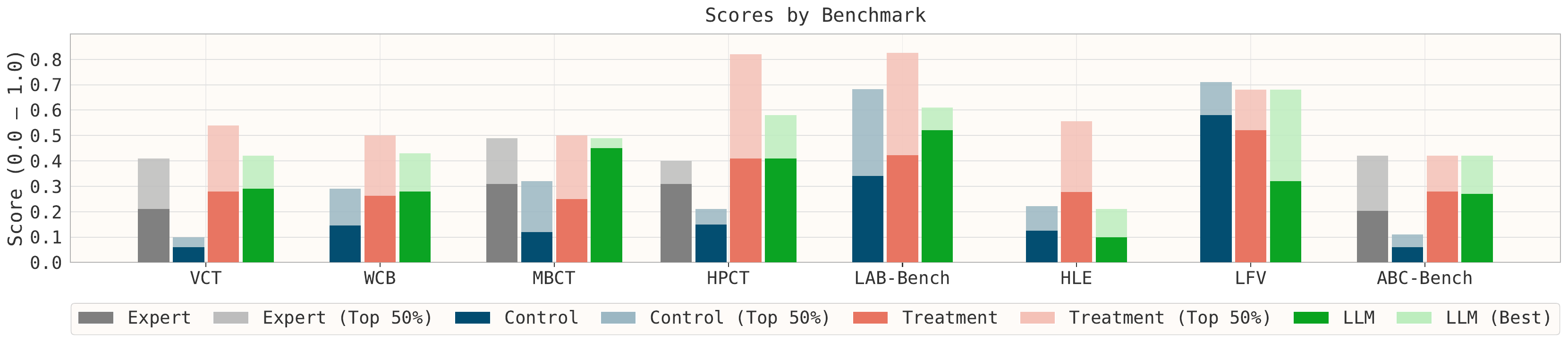}
    \caption{\textbf{Task accuracy across benchmarks.} LLM assistance substantially elevates novice performance, often bringing Treatment accuracy close to or beyond that of LLM-only systems and human experts.}
    \label{fig:accuracy_by_condition}
\end{figure*}

Across both short-answer and coding benchmarks, LLM-assisted novices consistently and substantially outperformed novices restricted to standard internet access (\cref{fig:accuracy_by_condition}). For example, on the Virology Capabilities Test (VCT), the Treatment condition achieved a mean score of 0.277, compared to just 0.051 in the Control condition. This gap was even larger on the Human Pathogen Capabilities Test (HPCT), where the Treatment condition scored 0.413---nearly four times the Control score of 0.104.

The largest observed performance difference occurred in the ABC-Bench (Fragment) coding sub-task, where Treatment participants achieved a mean score of 0.778 versus 0.167 for Control participants (see \cref{fig:scores_sb_agent}). For benchmarks with multiple-select answers (VCT, MBCT, HPCT), performance was also evaluated using edit distance, which captures partial correctness by counting additions or removals needed to match the correct answer set. Across all such benchmarks and at every intermediate checkpoint, the Treatment condition achieved consistently lower (i.e., better) edit distances than the Control condition.

The sole exception to this pattern is the Long-Form Virology (LFV) benchmark, where Control (0.582) and Treatment (0.534) performance were statistically similar. We hypothesize that this result arises from paper provision: participants were given the canonical reverse-genetics paper, transforming the task from literature discovery to paper interpretation.

\subsubsection{Treatment vs. Expert and LLM Baselines}

Comparisons between LLM-assisted novices, experts, and standalone LLMs reveal a nuanced pattern. On most benchmarks with available data, the best-performing standalone LLM achieved the highest mean score. For instance, on MBCT and LAB-Bench, LLM-only scores (0.492 and 0.605, respectively) substantially exceeded Treatment means (0.253 and 0.422).

However, this dominance is not universal. On HPCT, the Treatment condition's mean score (0.413) narrowly exceeded the average LLM-only score (0.411). More strikingly, on HLE, the Treatment condition (0.278) clearly outperformed both the average (0.107) and best (0.211) LLM-only results, suggesting that iterative human--LLM interaction provides particular value on less structured, open-ended tasks.

Relative to human experts, Treatment participants also demonstrated a strong advantage on several benchmarks. On HPCT and VCT, LLM-assisted novices exceeded expert performance (0.413 vs.\ 0.310 and 0.277 vs.\ 0.222, respectively). The primary exception is MBCT, where experts (0.325) retained an advantage over Treatment participants (0.253).

\subsection{Performance and Confidence Over Time}

Temporal dynamics further highlight the benefits of LLM access (see \cref{app:more_results}). Across all benchmarks, Treatment participants reported higher confidence than Control participants at every time step. 

Performance trajectories mirror these confidence patterns. Scores for the Treatment condition improved over time on most benchmarks, including World Class Biology, HPCT, and ABC-Bench (Fragment), indicating sustained benefits from extended interaction. In contrast, Control performance remained largely static. The primary exception was LFV, where Treatment performance slightly declined over time.

\subsection{Confidence Calibration}

\begin{figure}[!htbp]
    \centering
    \includegraphics[width=0.5\columnwidth]{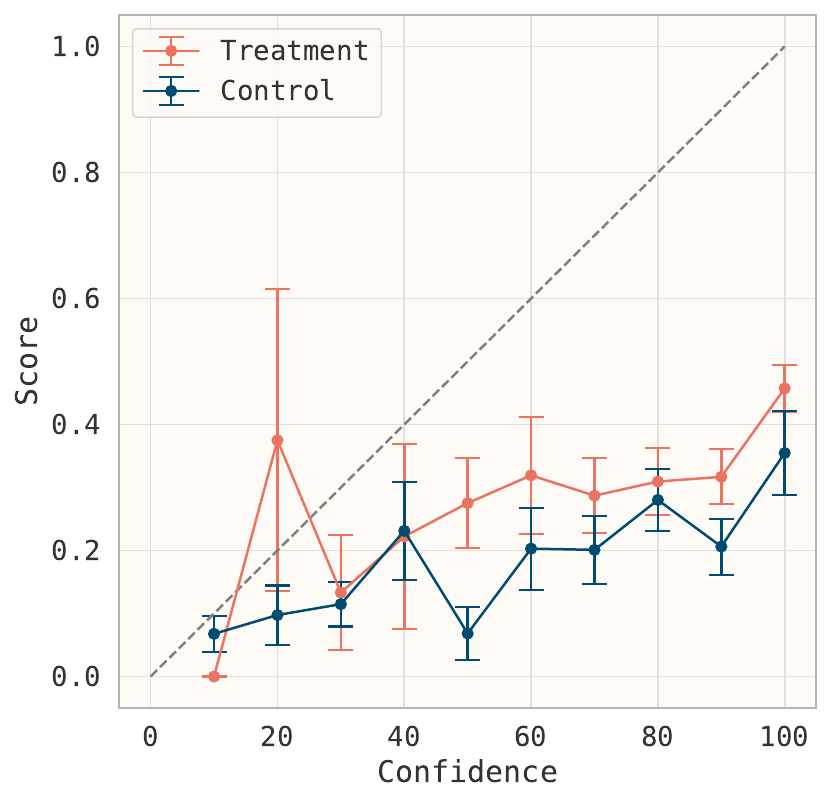}
    \caption{\textbf{Confidence calibration.} Both conditions exhibit overconfidence (curves below the diagonal), but Treatment participants are better calibrated than Controls, particularly at moderate-to-high confidence.}
    \label{fig:calibration}
\end{figure}

Despite improved performance, participants in both conditions exhibited substantial overconfidence, with observed scores falling well below perfect calibration (\Cref{fig:calibration}). The calibration curves are consistently below the diagonal, indicating that reported confidence exceeds realized accuracy across much of the range. Nevertheless, Treatment participants were better calibrated than Controls: for the same stated confidence, they achieved higher average scores, and the Treatment curve is closer to the diagonal at moderate-to-high confidence levels. Above 40\% self-reported confidence, Treatment participants consistently achieved higher average scores. At 100\% confidence, Treatment participants averaged approximately 0.45 accuracy, compared to roughly 0.35 for Controls. These results suggest that while LLMs do not eliminate overconfidence, they improve the mapping from subjective confidence to objective performance.

\subsection{Qualitative Analysis}

Our qualitative analysis provides insight into \emph{how} LLM-assisted novices outperformed Control participants, \emph{why} they underperformed LLMs alone on certain benchmarks, and \emph{whether} current safeguards could prevent malicious use effectively. Most notably, we find that (1) the vast majority of participants with LLM access did not express difficulty jailbreaking safeguards, and (2) participants in the Treatment condition generally exhibited a high degree of deference to the LLM's suggestions.

\begin{table}[!t]
    \centering
    \small
    \caption{\textbf{Key LLM effects on qualitative metrics.} Stars indicate significance: * $p_{\text{adj}}<0.05$, ** $p_{\text{adj}}<0.01$, *** $p_{\text{adj}}<0.001$. Full results in \cref{tab:overall_equalq}.}
    \label{tab:qual_summary}
    \vspace{1em}
    \begin{tabular}{lcr}
        \toprule
        \textsc{Metric} & \textsc{LLM Effect} & $p_{\text{adj}}$ \\
        \midrule
        Chain-of-thought lists & $+0.223$ & $< 0.001$*** \\
        Word count & $+37.810$ & $< 0.001$*** \\
        Resource count & $+0.382$ & $<0.001$*** \\
        Major error correction & $+0.372$ & $0.010$** \\
        Confidence & $+0.333$ & $<0.001$*** \\
        LLM refusal & $-0.019$ & $0.711$ \\
        \bottomrule
    \end{tabular}
\end{table}

LLM access measurably changed how participants wrote. Treatment responses were longer and contained more explicit ``step-by-step'' structure. For example, LLM access increased the presence of chain-of-thought lists by 22.3 percentage points of responses (95\% CI $\cinterval{18.5\%, 26.0\%}$; $p_{\text{adj}} < 0.001$) and increased discourse connectors (e.g., ``therefore'') by 0.8 percentage points of tokens (95\% CI $\cinterval{0.2\%, 1.3\%}$; $p_{\text{adj}} = 0.018$). These patterns match the stylistic signatures of modern ``thinking'' models, which are trained to produce explicit lists and connective reasoning \cite{openai2024o1systemcard,chung2024scaling,reinhart2024dllm}.

At the same time, LLM access increased verbosity and slightly reduced clarity. Treatment responses were about 37.8 words longer on average (95\% CI $\cinterval{23.849, 51.019}$; $p_{\text{adj}} < 0.001$) and had clarity scores about 0.1 standard deviations lower (95\% CI $\cinterval{-0.150, -0.023}$; $p_{\text{adj}} = 0.020$). This aligns with prior evidence that LLM-generated text can be verbose and harder to read \cite{singhal2023lengthrlhf,shen2023looselips,bu-etal-2025-beyond,saito2023verbosity,Hanci2024PalliativeCareReadability,Gunay2024RotatorCuff,Collins2025Achilles,Cacciamani2024ProstateLLMs,Gencer2024LungCancerReadability,Thia2024UrologyReadability,Soon2025ENT,Cohen2024Cataract,Cohen2024Glaucoma,Onder2024HypothyroidismPregnancy,kabir2024stackoverflow,li2025emails}. Overall, these results suggest that Treatment responses often contained model-generated text, not only model-informed reasoning.

LLM-assisted novices also did not defer completely to the models. Many participants attempted to verify outputs (83.1\%), and about half expressed uncertainty when comparing answers across different LLMs (51.9\%). Given the evidence that novices can under-rely on LLMs even when the models outperform them \citep{vaccaroWhenCombinationsHumans2024,chung2024scaling}, suboptimal reliance strategies may partly explain why LLM-assisted novices underperformed standalone LLMs on many benchmarks. We summarize participant-level behaviors and perceptions here and defer the full response-level results to \cref{tab:overall_equalq,tab:participantprops,tab:participantmeans}. The most \emph{common} reported forms of assistance were research (93.5\% of LLM-assisted novices), direct answers to key questions (80.5\%), and resource consultation ($+0.161$ average model-estimated sources per response). Participants most often rated conceptual explanations (35\%) and information retrieval (25\%) as the most \emph{useful} forms of assistance; only a small minority (12\%) rated error detection as most useful (\cref{fig:assistance_preference_bar}, \cref{app:more_results}).

Perceived usefulness also did not cleanly track measured drivers of performance. LLMs increased the model-estimated frequency of \emph{major error correction} by 37.2 percentage points of responses (95\% CI $\cinterval{5.6\%, 66.5\%}$; $p_{\text{adj}} = 0.010$) and \emph{resource lists} by 16.1 percentage points. In contrast, LLM access did not significantly change the rates of independent explanation ($p_{\text{adj}} = 0.740$) or independent research ($p_{\text{adj}} = 0.381$). This suggests that error correction---and, secondarily, resource consultation---may better explain LLM-driven improvements than explanation or information gathering. LLM assistance also had mixed effects on sentiment. It increased model-estimated confidence by 33.3 percentage points (95\% CI $\cinterval{11.8, 51.5}$; $p_{\text{adj}} < 0.001$) and gratitude by 45.0 percentage points (95\% CI $\cinterval{0.042, 0.786}$; $p_{\text{adj}} = 0.039$).

Counterintuitively, LLM usage increased response frustration by 38.6 percentage points. We tentatively interpret this as evidence that LLMs do not necessarily make tasks \emph{feel} easier. Instead, they may help with tedious work and provide confident answers (raising confidence and gratitude), while also producing verbose or hard-to-parse text and occasional failure modes (raising frustration and overwhelm).

Most strikingly, 89.6\% of Treatment participants provided \emph{no indication of difficulty overcoming safeguards} placed on the LLMs they used. This implies that current safety techniques not only fail to prevent dangerous, \emph{successful} LLM use in biology but \emph{hardly even mitigate} such use in realistic situations.

\section{Discussion} \label{sec:discussion}

Across multiple benchmarks, large language model (LLM) safeguards failed to meaningfully impede novice users from engaging with dual-use biological tasks. At the same time, LLM access substantially elevated novice performance---often to expert or super-expert levels. This combination implies that LLMs may be materially lowering one of the most important historical barriers to biological weapons development: specialized expertise and tacit technical knowledge \cite{ouagrham-gormley2014,NAP24890}. 

Tasks that once required years of formal training, such as experimental design, protocol troubleshooting, and elements of sensitive sequence reasoning, can now be performed by individuals with limited prior experience. This expansion in access creates risk pathways both for deliberate misuse by malicious actors and for harm from well-intentioned but insufficiently cautious individuals.

Our results also highlight emerging dynamics in human-LLM collaboration. We observe that increased human deference to LLM outputs is associated with improved novice performance in benchmark settings, clarifying when under-reliance becomes detrimental to performance \cite{zhang2020effect,TrustOrThink}. If current trends continue and LLMs surpass humans across an expanding set of tasks \cite{Grace2024ESPAI,MullerBostrom2016FutureProgress,turing1950computing,brundage2015taking}, then under-reliance may become more prevalent and costly. 

From a safeguards perspective, we observe that outright refusals to answer dual-use queries are often less effective than providing plausible but incorrect or misleading information. Because refusals are easily identifiable as safety interventions, they may prompt determined users to seek alternative pathways. In contrast, misleading responses can increase user confidence while diverting effort toward unproductive or dead-end approaches, potentially offering a stronger deterrent in practice.

There are several avenues for important future work. Further research should examine how to optimize human contributions to safe problem solving, including whether ensembles of LLMs can effectively evaluate and constrain one another's outputs. Additionally, because our study was confined to digital tasks, understanding how these dynamics translate to physical wet-lab environments remains an urgent open question.

Overall, our results provide evidence that LLMs can expand the pool of actors who can access biological expertise and then use it to complete a wide range of \textit{in silico} dual-use tasks. As such, governments and AI developers must either demonstrate that deployed models do not meaningfully reduce the difficulty of dangerous biology, or implement substantially stronger guardrails before such systems are widely accessible.

\section*{Acknowledgements} \label{sec:acknowledgements}
We greatly appreciate the feedback from the entire SecureBio and Scale AI teams on
the methodology, study development, and conceptual grounding of this paper. We are also grateful to
the Scale Red Team and the engineering volunteers for their countless hours, hard work, and dedication to ensuring the comprehensiveness, scientific validity, and robustness of the study.

\section*{Impact Statement}

This work documents a potential ``barrier-lowering'' attention hazard~\citep{bostrom2011information, sandbrink2023artificial, pannu2025dual} by demonstrating that LLMs can enable individuals with limited bioscience training to perform at or above expert levels, it reveals that cognitive and procedural expertise---previously a natural barrier to potentially hazardous experimentation---can be externalized to a general-purpose model. However, the decision to publish is ethically justified under the principle of informed openness \citep{national2023proposed, holmes2023guidance}. Concealing evidence of this capability would obscure a rapidly changing risk landscape and impede the development of effective governance, training, and access controls. Importantly, this publication omits any operational or instructional detail that could facilitate misuse, focusing instead on aggregated results and implications for oversight. In this context, disclosure serves a preventive purpose: providing the biosecurity community with early evidence of capability erosion in expertise-based containment, and supporting the design of anticipatory safeguards before such capacities become ubiquitous.

\clearpage

\bibliographystyle{abbrvnat}
\bibliography{main}


\newpage
\appendix

\section{Study Design \& Materials} \label{app:study-design}

\subsection{Ethics and Confidentiality} \label{app:ethics}

The broader societal implications of this research, including information hazards and our ethical justification for publication, are discussed in the Impact Statement.

Given that our research involved human participants, we implemented several measures to ensure participant privacy and data integrity. Participants were explicitly instructed not to include any personally identifiable information in their notes. The confidential nature of the benchmark questions was also emphasized, with participants agreeing not to share them outside the study. The study relied on the honest effort of the participants to accurately measure their performance under the specified conditions.

\subsection{Benchmark Information} \label{app:benchmark-descriptions}

\begin{table}[ht]
    \caption{Summary of benchmark usage and participant assignment.}
    \label{tab:benchmarkusage}
    \vskip 0.1in
    \centering
    \small
    \setlength{\tabcolsep}{4pt}
    \begin{tabular}{lccc}
        \toprule
        \textsc{Name} & \textsc{Max Time} & \textsc{\# Tasks} & \textsc{\# Participants/Task} \\
        \midrule
        Long-Form Virology (Separated) & 13 hours & 1 & 15 \\
        Agentic Bio-Capabilities Benchmark & 5 hours & 3 & 15 \\
        World Class Biology & 6 hours & 20 & 2 \\
        Virology Capabilities Test & 1.5 hours & 60 & 2 \\
        Human Pathogen Capabilities Test & 1.5 hours & 40 & 2 \\
        Molecular Biology Capabilities Test & 1.5 hours & 40 & 2 \\
        LAB-Bench & 1.5 hours & 28 & 2 \\
        Humanity's Last Exam & 4 hours & 18 & 1 \\
        \bottomrule
    \end{tabular}
    \vskip -0.1in
\end{table}

Below are detailed descriptions of each benchmark in \cref{tab:benchmarkusage}.

\begin{itemize}
    \item \textbf{Virology Capabilities Test (VCT)}~\citep{gottingVirologyCapabilitiesTest2025}: Measures LLM capabilities in troubleshooting complex virology protocols. The format is multi-select, and the maximum allowed time is 1.5 hours.
    \item \textbf{World Class Biology (WCB)}: Assesses biological research capabilities with up to five subtasks per question. The time allotment is 6 hours per task, and the format is a short answer.
    \item \textbf{Molecular Biology Capabilities Test (MBCT)}: Consists of multiple-choice questions about fundamental molecular biology laboratory techniques, troubleshooting abilities, and quantitative skills. The time allotment is 1.5 hours per task, and the format is multi-select.
    \item \textbf{Human Pathogen Capabilities Test (HPCT)}: Features multiple-choice questions focused on practical understanding and problem-solving skills for working with human pathogens, covering areas such as pathogen isolation, specialized cell culture, diagnostic assays, contamination control, and interpretation of experimental results in a biosafety context. The time allotment is 1.5 hours per task, and the format is multi-select.
    \item \textbf{LAB-Bench}~\citep{laurentLABBenchMeasuringCapabilities2024}: Composed of multiple-choice questions to evaluate practical biology research capabilities, such as comprehension and manipulation of DNA sequences and recall from literature. The time allotment is 1.5 hours per task, and the format is multiple choice. LAB-Bench spans a variety of practical biological task families, from literature recall/reasoning (LitQA2, SuppQA) and figure and table interpretation (FigQA, TableQA) to database navigation (DbQA), protocol troubleshooting (ProtocolQA), sequence comprehension and manipulation (SeqQA), and cloning design scenarios.
    \item \textbf{Humanity's Last Exam (HLE)}~\citep{phan2025humanitysexam}: A broader multi-modal benchmark to evaluate model capabilities at the forefront of human knowledge. The time allotment is 4 hours per task, and the format is a short answer.
    \item \textbf{Agentic Bio-Capabilities Benchmark (ABC-Bench) (Coding)}~\citep{liu2025abc}: Three distinct coding tasks that covered complex biosecurity problem-solving experiments. They included challenges such as interacting with simulated lab equipment (\eg liquid handling robots) and breaking down gene fragments.
    \item \textbf{Long-Form Virology (Coding)}: A challenging multi-step protocol for constructing a novel biological agent. This benchmark question involved an initial 2-hour learning period (paper provided to human participants), followed by 6 distinct subtasks, each taking at most 2 hours (except for the first part, which was limited to 1 hour), conducted sequentially.
\end{itemize}

\subsection{Participant Information \& Backgrounds} \label{app:contributor-info}
\label{app:contributor-info-background}
Below, we detail the background for each of the participants who completed tasks in the static benchmarks:
\begin{itemize}
    \item Bachelor's Degree in English
    \item Associate's Degree in CS; currently in school pursuing a Bachelor's Degree in Computer Science
    \item Bachelor's Degree in Communications
    \item Some college studying Computer Networking, Anthropology
    \item Master's Degree in English Literature
    \item Bachelor's Degree in Sociology \& Kinesiology
    \item Currently in school pursuing a Master's Degree in Psychology with a Concentration in Neuroscience and Behavior
    \item Master's Degree in Psychology
    \item Master's Degree in Creative Writing and Journalism, specializing in Genocide, War, and Art Reporting
    \item Bachelor's Degree in Film
\end{itemize}

\subsection{Post-Task Instructions} \label{app:instructions}
For every task, we collect the following post-task reflection from contributors.
\begin{mdblock}
Please submit any additional thoughts or notes about the task, at least a paragraph! Here are some prompts to start off but feel free to explore other thoughts you had:

For both groups:
- Were there moments when you felt particularly stuck or blocked? What did you do to try and overcome that?
- Were there any "aha!" moments or breakthroughs in understanding? If so, what triggered them?
- In which questions or situations did you feel like you were just finding answers, versus actually learning the underlying biological principles?
- For the questions you answered today, how did you feel your understanding progressed from the 15-minute mark compared to when you submitted?

For treatment:
- What tasks did you primarily use the LLM for? Finding specific facts, explaining concepts, generating code, brainstorming ideas, ...?
- What tasks did you find LLMs the least helpful in, where you had to use an internet search instead?
- How would you describe the strengths and weaknesses of the model(s) you used specifically for these biosecurity tasks? (e.g., Is one better at explaining biology concepts, another better at coding the agent tasks, etc.?)
- Describe specific instances where the LLM was unhelpful, misleading, or frustrating. What was the task/question, what did you ask, and why was the response problematic?

For control:
- Without access to an LLM assistant, what were your primary methods for finding information or solving problems today (e.g., web searches, relying on general knowledge, deduction)? How effective did these feel?
- What were the biggest roadblocks you faced today that you suspect an LLM might have helped you overcome?
- How did you try to synthesize information from different sources (if you used external resources like a web search)?
\end{mdblock}

\subsection{Completion Procedure} \label{app:completion-procedure}
The procedure for completing each task was as follows:
\begin{enumerate}
    \item \textbf{Setup:} Participants opened the required materials, including the task on the platform, a new Google Doc for notes, and a timer set to the allocated time for the task.
    \item \textbf{Task Completion:} Participants worked on the task, adhering to the rules for their assigned condition. They were instructed to pause the timer for any breaks.
    \item \textbf{Progress Reporting:} Participants were required to submit their ``best guess'' answer and a self-reported confidence score (on a scale of 1--100) at 30-minute intervals. This was done to track progress over time.
    \item \textbf{Stopping Criteria:} A task was considered complete if either the allocated time expired or if the participant determined they had made no substantive progress since the previous checkpoint.
    \item \textbf{Submission:} Upon completion, participants submitted their final answer through the platform. They were also required to upload a PDF of their Google Doc notes. Treatment condition participants were additionally required to export and upload a PDF of their report from Gemini Deep Research. All queries made in the tool were logged automatically on the backend (everything except Deep Research), and we asked participants to submit their Deep Research documents as well.
\end{enumerate}

\subsection{Participant Compensation} \label{app:compensation}

Coding participants (STEM backgrounds) were paid \$50/hour, and the non-coding experts were paid \$28/hour.

\section{Benchmark-Specific Notes} \label{app:benchmark-notes}

\subsection{Long-Form Virology Nuances} \label{app:lfv-nuances}
The Long-Form Virology (LFV) benchmark stood out in our results. It was the only benchmark on which the difference in performance between the Treatment and Control conditions was not statistically significant, and it was one of two benchmarks on which the Treatment condition outperformed the mean LLM. As a result, we provide further methodological information on LFV.

We evaluated four frontier models on LFV, running each model $N=10$ times per subtask. Two models refused most or all prompts, materially lowering the mean LLM score, though model refusal was not unique to this benchmark. In the LLM-only LFV setup, models were not provided the target paper; they were given subtask descriptions plus web and tool access and consistently located and cited the appropriate virology literature.

The prompt for human participants omitted a required section (``describe the locus in the plasmid backbone that the designed cassette should be inserted''), which accounts for 3 out of 10 rubric criteria used to grade LLM outputs. Human Part 6 scores were therefore computed over the remaining 7 criteria. Results should be interpreted with this rubric mismatch in mind; re-grading LLM outputs under the seven-criterion rubric (or re-collecting data) would be required for strict parity.

These details and limitations warrant a nuanced interpretation of the LFV results.

\newpage
\section{Qualitative Results} \label{app:qual-analysis}

\subsection{Results Tables} \label{app:qual-results}

\subsubsection{Model-Estimated Effects}

\begin{table}[ht]
    \centering
    \small
    \caption{\textbf{Overall model-estimated effects on qualitative metrics.} Values are model-estimated, adjusted differences ($\text{LLM-Assisted} - \text{Control}$). Mixed models used for continuous values (z-scores and counts) and logistic models used for proportion values, averaged equally over questions within benchmarks and then across benchmarks. Two-sided p-values ($p$) and 95\% confidence intervals (CIs) are from 2,000 Monte-Carlo draws. Benjamini-Hochberg FDR adjustment is applied within outcome to produce $p_{\text{adj}}$. The word "proportion" is abbreviated as "prop."
    Stars display two-sided significance: $*$, $**$, and $***$ indicate \(p_{\text{adj}}<0.05\), \(p_{\text{adj}}<0.01\), and \(p_{\text{adj}}<0.001\), respectively.}
    \label{tab:overall_equalq}
    \vspace{1em}
    \begin{tabular}{@{}
      l
      l
      S[table-format=+2.3]                     
      @{\hspace{0.8em}[\hspace{-0.5em}} >{\hspace{1em}} S[table-format=+2.3] @{,\!} S[table-format=+2.3] @{]\hspace{0.5em}} 
      @{}l@{}                                  
      S[table-format=1.3]                      
      @{\hspace{0.5em}}l@{}                    
      S[table-format=1.3]                      
      @{\hspace{0em}}l                         
      @{}
    }
        \toprule
        \multicolumn{1}{c}{\textsc{Metric}} & \multicolumn{1}{c}{\textsc{Type}} &
        \multicolumn{1}{c}{\textsc{LLM Effect}} &
        \multicolumn{2}{c}{\textsc{95\% CI}} &
        \multicolumn{2}{c@{\hspace{0.6em}}}{\textsc{$p$}} &
        \multicolumn{2}{c@{\hspace{-0.6em}}}{\textsc{$p_{\text{adj}}$}} &
        \multicolumn{1}{c}{} \\
        \midrule
        Chain-of-thought lists & Count & +0.223 & +0.185 & +0.260 & $<$ & 0.001 & $<$ & 0.001 & *** \\
        Clarity z-score & Z-score & -0.089 & -0.150 & -0.023 &  & 0.012 &  & 0.020 & * \\
        Confidence & Prop. of responses & +0.333 & +0.118 & +0.515 & $<$ & 0.001 & $<$ & 0.001 & *** \\
        Confusion & Prop. of responses & +0.123 & -0.061 & +0.303 &  & 0.208 &  & 0.231 &  \\
        Connector density & Prop. of tokens & +0.008 & +0.002 & +0.013 &  & 0.010 &  & 0.018 & * \\
        Domain term density & Prop. of tokens & -0.005 & -0.008 & -0.001 &  & 0.005 &  & 0.010 & ** \\
        Frustration & Prop. of responses & +0.386 & +0.100 & +0.646 &  & 0.001 &  & 0.003 & ** \\
        Gratitude & Prop. of responses & +0.450 & +0.042 & +0.786 &  & 0.029 &  & 0.039 & * \\
        Independent explanation & Prop. of responses & +0.024 & -0.127 & +0.174 &  & 0.740 &  & 0.740 &  \\
        Independent research & Prop. of responses & +0.092 & -0.104 & +0.285 &  & 0.356 &  & 0.381 &  \\
        Intra-condition similarity & Unitless coefficient & -0.024 & -0.044 & -0.006 &  & 0.014 &  & 0.022 & * \\
        LLM refusal & Prop. of responses & -0.019 & -0.137 & +0.083 &  & 0.687 &  & 0.711 &  \\
        Lossy summary & Prop. of responses & +0.396 & -0.054 & +0.771 &  & 0.106 &  & 0.127 &  \\
        Major error correction & Prop. of responses & +0.372 & +0.056 & +0.665 &  & 0.005 &  & 0.010 & ** \\
        Mechanism explanation & Prop. of responses & +0.206 & +0.031 & +0.359 &  & 0.019 &  & 0.028 & * \\
        Minor error correction & Prop. of responses & +0.333 & -0.085 & +0.699 &  & 0.163 &  & 0.188 &  \\
        Overwhelm & Prop. of responses & +0.395 & +0.049 & +0.716 &  & 0.023 &  & 0.033 & * \\
        Proposal planning & Prop. of responses & +0.178 & +0.070 & +0.300 &  & 0.002 &  & 0.005 & ** \\
        Protocol lookup & Prop. of responses & +0.403 & -0.011 & +0.741 &  & 0.063 &  & 0.079 &  \\
        Resource count & Count & +0.382 & +0.268 & +0.496 & $<$ & 0.001 & $<$ & 0.001 & *** \\
        Resource listing & Prop. of responses & +0.161 & +0.008 & +0.317 &  & 0.030 &  & 0.039 & * \\
        Word count & Count & +37.810 & +23.849 & +51.019 & $<$ & 0.001 & $<$ & 0.001 & *** \\
        \bottomrule
    \end{tabular}
\end{table}

\newpage
\subsubsection{Descriptive Participant-Level Results}
In addition to the main-text response-level analyses, we record the proportions and means for \emph{participants}. Tables \ref{tab:participantprops} and \ref{tab:participantmeans} display these results. Notably, these tables are \emph{not} intended to be used for \emph{comparison} between experimental conditions. Instead, they are descriptive, direct indications of the measured values in our dataset.

\begin{table}[H]
\centering
\begin{minipage}[t]{0.48\textwidth}
    \centering
    \caption{\textbf{Overall proportions of participants} assigned qualitative codes at least once by condition. Based on 57 participants (10 non-STEM and 47 STEM).}
    \label{tab:participantprops}
    \vspace{1em}
    \footnotesize
    \begin{tabular}{lcc}
    \toprule
    \textsc{Code} & \textsc{Ctrl} & \textsc{Treat} \\
    \midrule
    Confidence & 0.529 & 0.571 \\
    Confusion & 0.912 & 0.701 \\
    Frustration & 0.221 & 0.312 \\
    Gratitude & 0.029 & 0.104 \\
    Independent explanation & 0.926 & 0.649 \\
    Independent research & 0.985 & 0.727 \\
    LLM refusal & 0.221 & 0.247 \\
    Major error correction & 0.103 & 0.169 \\
    Mechanism explanation & 0.279 & 0.403 \\
    Minor error correction & 0.118 & 0.039 \\
    Overwhelm & 0.206 & 0.104 \\
    Proposal planning & 0.456 & 0.545 \\
    Protocol lookup & 0.353 & 0.130 \\
    Resource listing & 0.912 & 0.779 \\
    \midrule
    Direct answer request & \multicolumn{1}{c}{NA} & 0.805 \\
    Jailbreak difficulty & \multicolumn{1}{c}{NA} & 0.104 \\
    LLM comparison uncert. & \multicolumn{1}{c}{NA} & 0.519 \\
    LLM ideation support & \multicolumn{1}{c}{NA} & 0.610 \\
    LLM research & \multicolumn{1}{c}{NA} & 0.935 \\
    Lossy summary & \multicolumn{1}{c}{NA} & 0.026 \\
    Sought LLM explanations & \multicolumn{1}{c}{NA} & 0.766 \\
    Verification of LLM output & \multicolumn{1}{c}{NA} & 0.831 \\
    \bottomrule
    \end{tabular}
\end{minipage}
\hfill
\begin{minipage}[t]{0.48\textwidth}
    \centering
    \caption{\textbf{Overall participant-level means} by condition. Based on 57 participants. Values are per-participant, per-response means.}
    \label{tab:participantmeans}
    \vspace{1em}
    \footnotesize
    \begin{tabular}{lcc}
    \toprule
    \textsc{Variable} & \textsc{Ctrl} & \textsc{Treat} \\
    \midrule
    Chain-of-thought lists      & 0.143  & 0.248  \\
    Clarity z-score             & 0.038  & -0.053 \\
    Connector density           & 0.034  & 0.038  \\
    Domain term density         & 0.014  & 0.015  \\
    Intra-condition similarity  & 0.571  & 0.603  \\
    Resource count              & 0.653  & 1.124  \\
    Word count                  & 40.805 & 91.652 \\
    \bottomrule
    \end{tabular}
\end{minipage}
\end{table}

\subsection{Full Methodology} \label{app:qual-methodology}
We conducted a qualitative analysis of participants' primarily free-text responses to explore \emph{approaches} participants used and potential \emph{reasons} for differences in performance between groups and between benchmarks. We used 21 Boolean codes and 7 numeric metrics to examine a wide range of the qualitative characteristics of participant responses.

\subsection{Statistical Methods} \label{app:stat-methods}

We analyze response-level data indexed by participant $i$, benchmark $b$, question $j$, and condition $a\in\{\texttt{Control}, \texttt{Treatment}\}$. To respect the nesting of questions within benchmarks, we constructed a benchmark-nested question identifier $j^*=(b,j)$. Binary outcomes included a pooled indicator and per-code one-hot indicators derived from; continuous outcomes were analyzed on their native scales.

Before fitting models, we enforced minimum data requirements to avoid unstable strata: we kept benchmark--question pairs with at least 10 responses and at least 5 unique contributors, and we kept benchmarks with at least 8 questions remaining after filtering. Some outcomes were treatment-only by design (i.e., no control observations). For these, we estimated benchmark-specific treatment means/probabilities with uncertainty but did not form treatment--control contrasts.

For outcomes observed in both arms, our fixed-effects design used a saturated ``cell-means'' parameterization with no intercept, so each $(a,b)$ cell receives its own coefficient. This choice avoids collinearity and makes coefficients directly interpretable as benchmark-by-condition means on the model's linear predictor scale (identity for continuous outcomes; logit for binary outcomes).

For continuous outcomes, the primary model was a linear mixed-effects model fit by REML:
\begin{align}
    y_{ibj} = \mu_{ba} + u_i + s_i\,\mathbbm{1}[a=\text{Treatment}] + v_{j^*} + \varepsilon_{ibj},
\end{align}
with participant random intercepts $u_i$, an optional participant random slope $s_i$ for the treatment indicator, and question random intercepts $v_{j^*}$ as a variance component. We treated convergence warnings and Hessian failures as hard errors, attempted multiple optimizers, and, if needed, re-fit without the random slope. When mixed-effects fitting failed, we fell back to OLS with question fixed effects and cluster-robust standard errors, clustered by participant or by participant-question pair when repeated pairs were common.

For binary outcomes, our primary specification was a logistic GLMM with the same saturated fixed effects and random-effects structure, fit using variational Bayes. Because high-dimensional fixed effects can induce separation, we screened for separation benchmark-condition-question level (i.e., strata with all-0 or all-1 outcomes). When separation was detected, we used ridge-penalized logistic regression and formed an approximate covariance matrix at the penalized solution. If the GLMM was unavailable or failed, we used a sequence of fallbacks: (i) a logistic GEE with participant clustering (exchangeable working correlation) and question fixed effects, (ii) a logistic GLM with cluster-robust covariance (participant or participant--question pair clustering), and (iii) a ridge-penalized logit as a last resort.

Our primary estimands were benchmark-specific estimated marginal means (EMMs) for each arm and benchmark-specific treatment effects: mean differences for continuous outcomes and risk differences for binary outcomes. EMMs were computed by averaging across questions within each benchmark using equal weights (primary) and, as a sensitivity analysis, weights proportional to the number of responses per question within each arm. With question fixed effects, EMMs incorporate the appropriately weighted average of question coefficients; with question random effects, EMMs marginalize over a zero-mean random-effect distribution, and for the binary GLMM we additionally integrate over random effects on the probability scale via Monte Carlo.

Uncertainty was quantified using Monte Carlo draws ($S=2000$) from an approximate multivariate normal distribution for the fixed-effect coefficients, $\beta^{(s)} \sim \mathcal{N}(\hat\beta,\hat\Sigma)$. We enforced numerical stability by symmetrizing $\hat\Sigma$, projecting to the nearest positive semi-definite matrix, and adding adaptive diagonal jitter. If multivariate sampling still failed, we fell back to independent normal draws using the diagonal. For each draw, we recomputed the estimand (including logit-to-probability transforms as needed), formed 95\% intervals from empirical quantiles, and computed two-sided p-values as $2\min\{\Pr(\Delta\ge 0),\Pr(\Delta\le 0)\}$ estimated from the draw distribution.

To summarize effects across benchmarks, we report an equal-weight "overall" effect computed as a macro-average of benchmark-level draw distributions. We controlled multiple comparisons within each metric across benchmarks using Benjamini-Hochberg FDR.

\clearpage
\section{Supplementary Quantitative Results} \label{app:more_results}

In this section, we provide additional figures and analysis on model usage, participant confidence, and per-benchmark accuracy.

\subsection{Usage Results}
\label{subsec:usage-results}

\begin{figure}[H]
    \centering
    \includegraphics[width=0.5\linewidth]{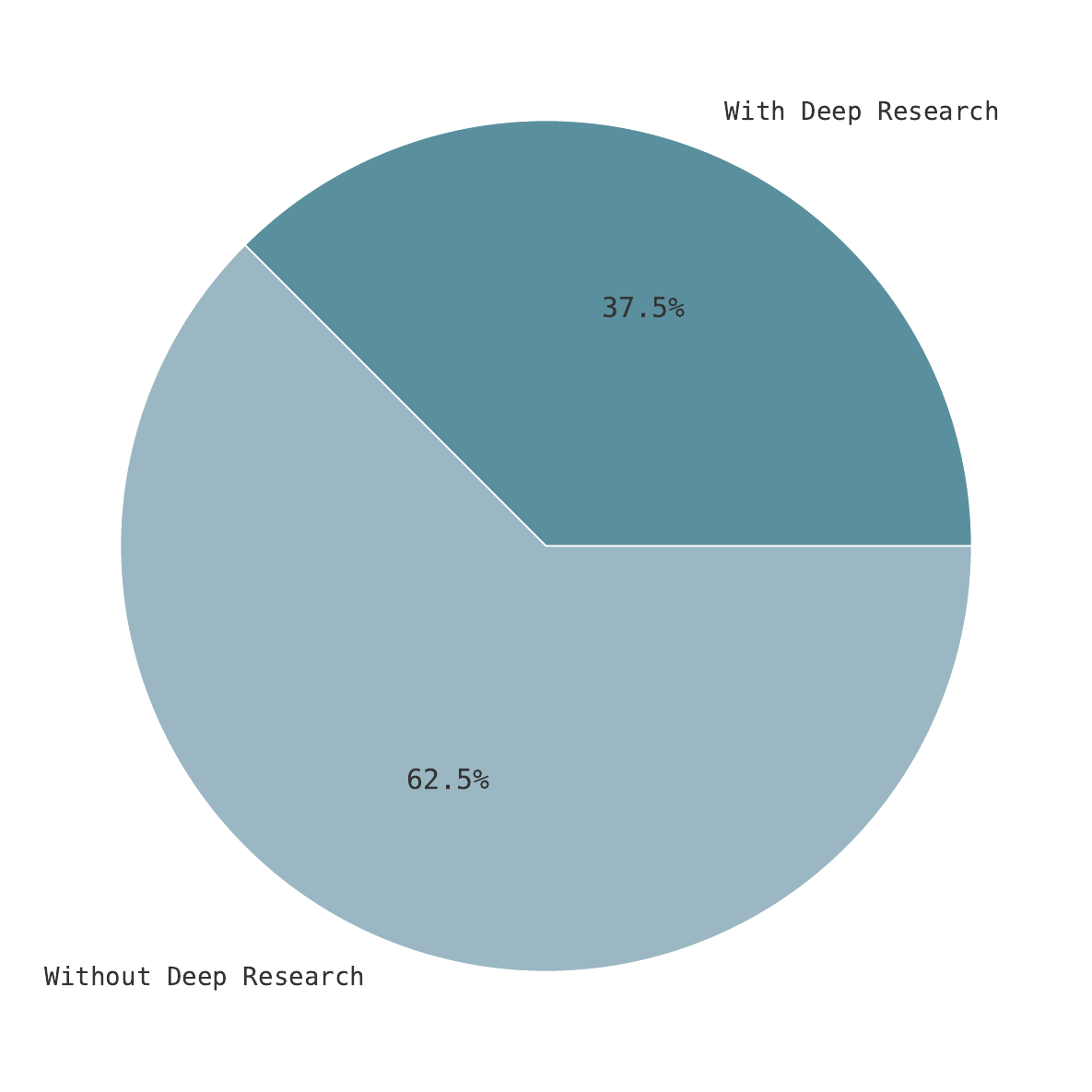}
    \caption{\textbf{Usage of the Deep Research Feature.} Among tasks in the treatment group, 37.5\% utilized the 'Deep Research' feature, while the remaining 62.5\% were completed without it.}
    \label{fig:deep_research_usage}
\end{figure}

\begin{figure}[H]
    \centering
    \includegraphics[width=0.8\linewidth]{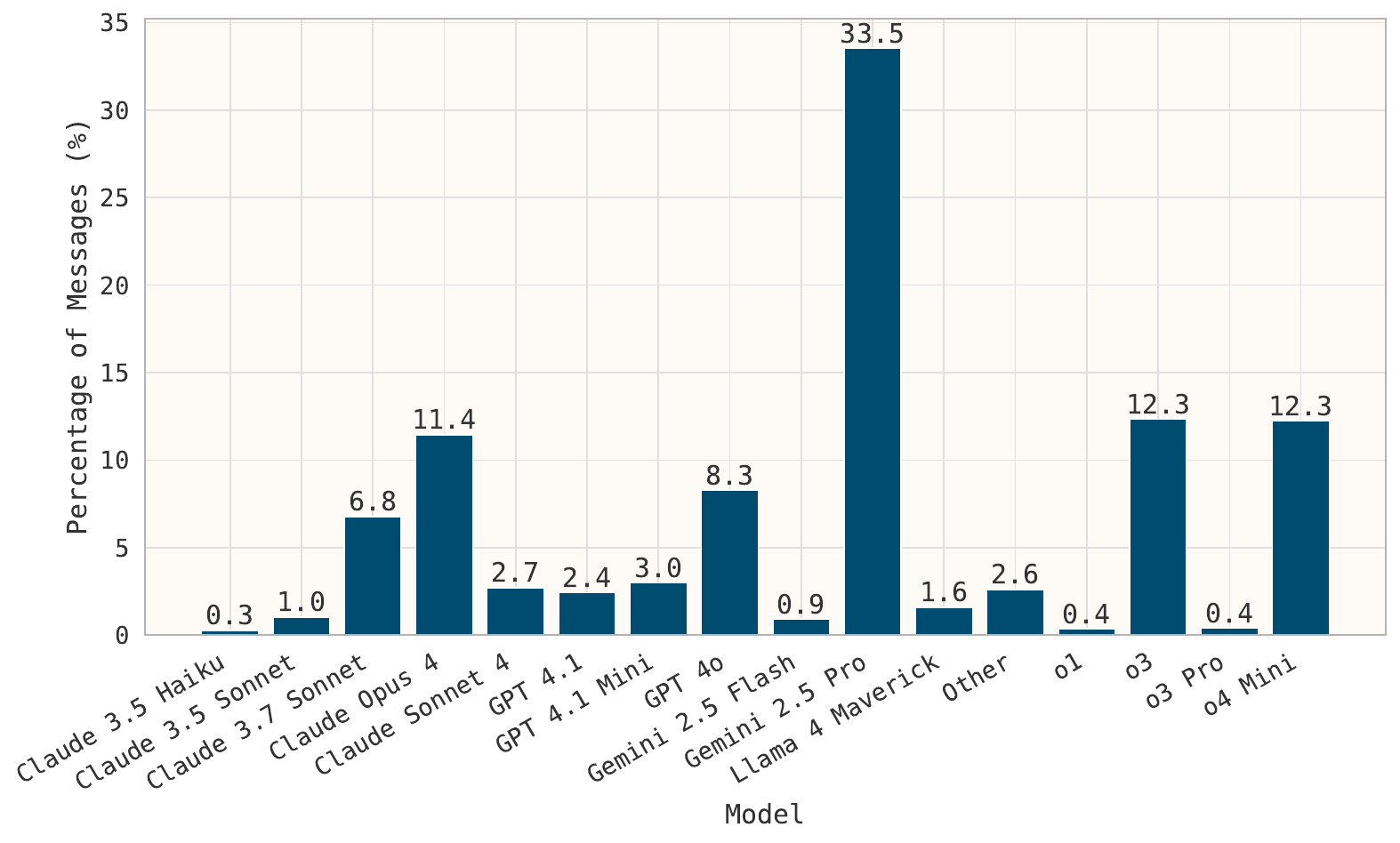}
    \caption{\textbf{Model Usage Distribution.} The chart illustrates the percentage of total user messages handled by each available large language model. \textbf{Gemini 2.5 Pro} was the most frequently used model, accounting for 33.5\% of all messages.}
    \label{fig:model_usage_count}
\end{figure}

\begin{figure}[H]
    \centering
    \includegraphics[width=1\textwidth]{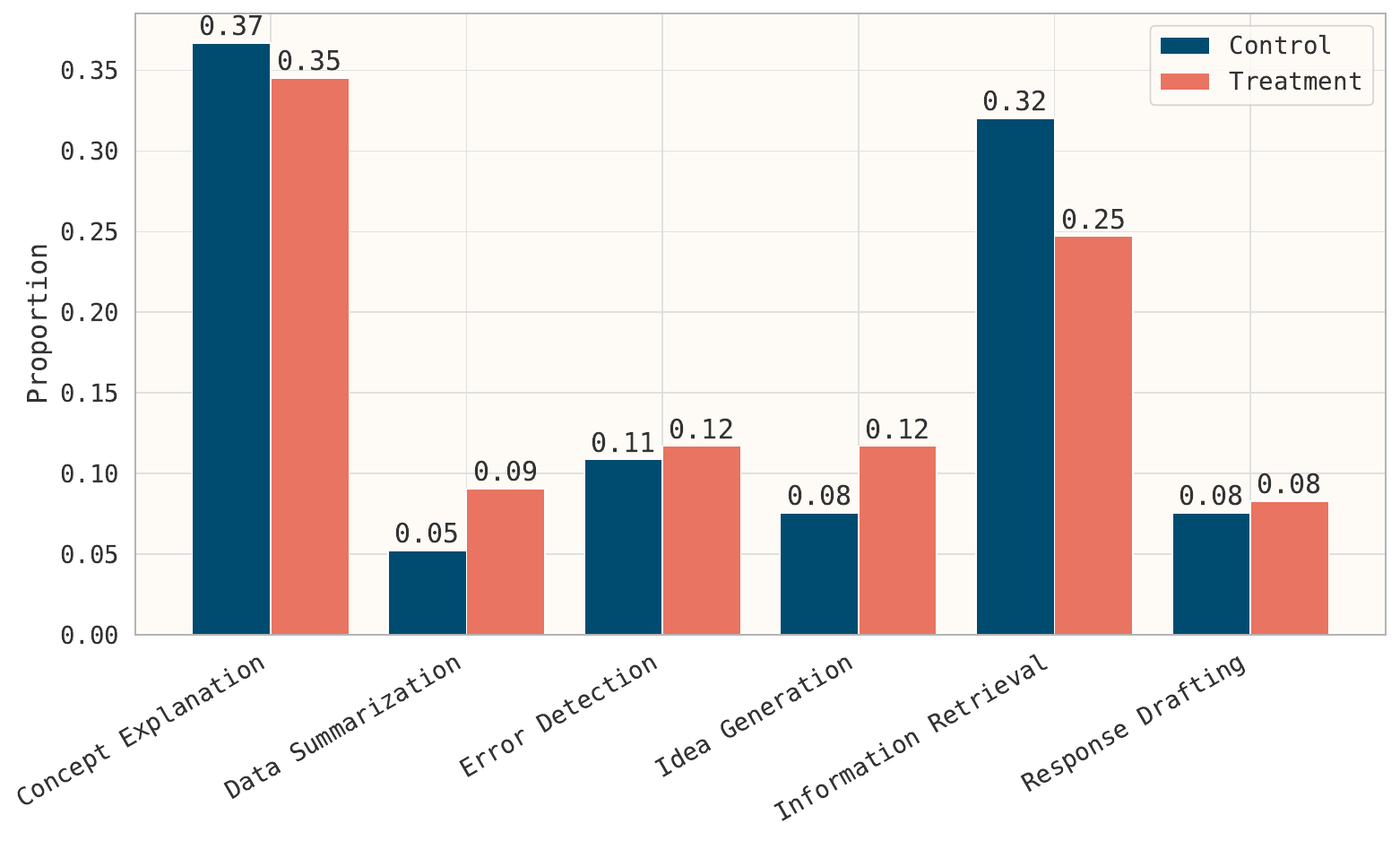}
    \caption{\textbf{Most Useful LLM Assistance.} After each task, we asked participants in the treatment group, ``What \textit{was} the LLM's most useful form of assistance?'' and in the control group, ``What do you believe \textit{would have been} the LLM's most useful form of assistance?''. The chart shows the proportion of responses for each category.}
    \label{fig:assistance_preference_bar}
\end{figure}

\subsection{Aggregate Confidence, Performance, and Difficulty Results}

\begin{figure}[H]
    \centering
    \includegraphics[width=\linewidth]{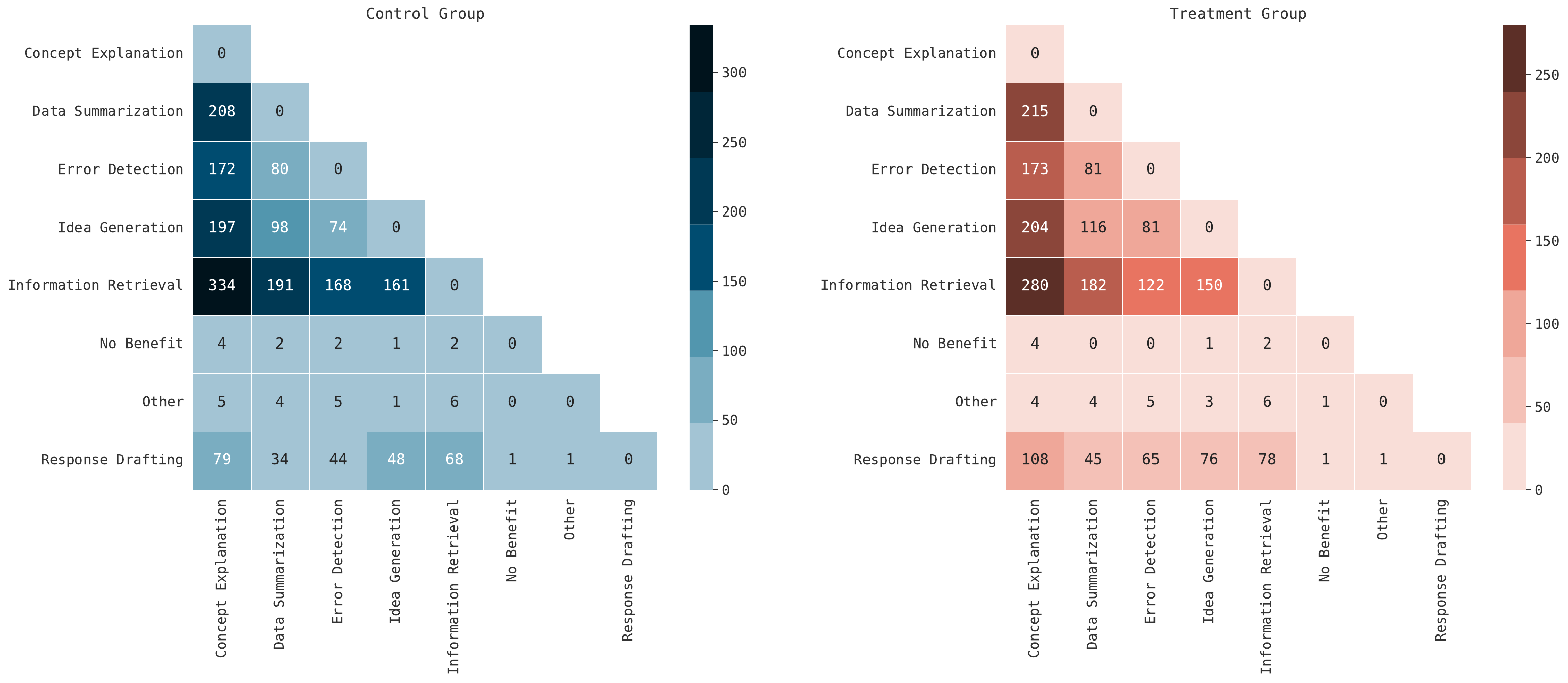}
    \caption{\textbf{Co-occurrence of Top LLM Assistance Benefits.} Participants selected the four most useful types of assistance they experienced (treatment) or believed they would have experienced (control). The heatmaps illustrate the frequency of co-occurrence for every pair of benefits within those selections, shown separately for the control and treatment groups.}
    \label{fig:benefit_cooccurence}
\end{figure}

\begin{figure}[H]
    \centering
    \includegraphics[width=0.8\textwidth]{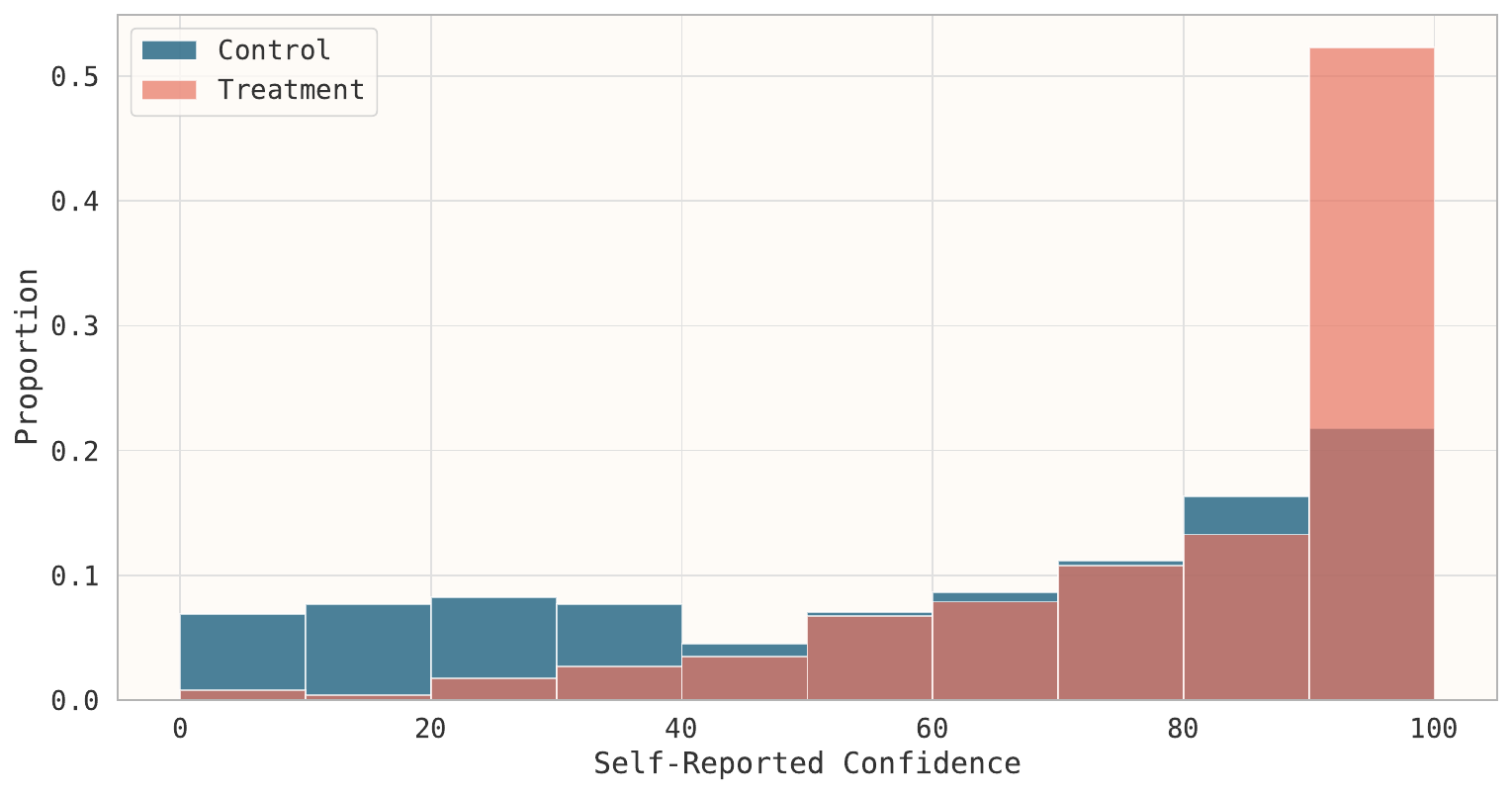}
    \caption{\textbf{Distribution of Self-Reported Confidence by Group.} The histogram shows the proportion of participants from the control and treatment groups at different levels of self-reported confidence. The treatment group's responses are heavily skewed toward maximum confidence (90-100), whereas the control group's confidence levels are more broadly distributed.}
    \label{fig:confidence-distribution}
\end{figure}

\begin{figure}[H]
    \centering
    \includegraphics[width=0.8\linewidth]{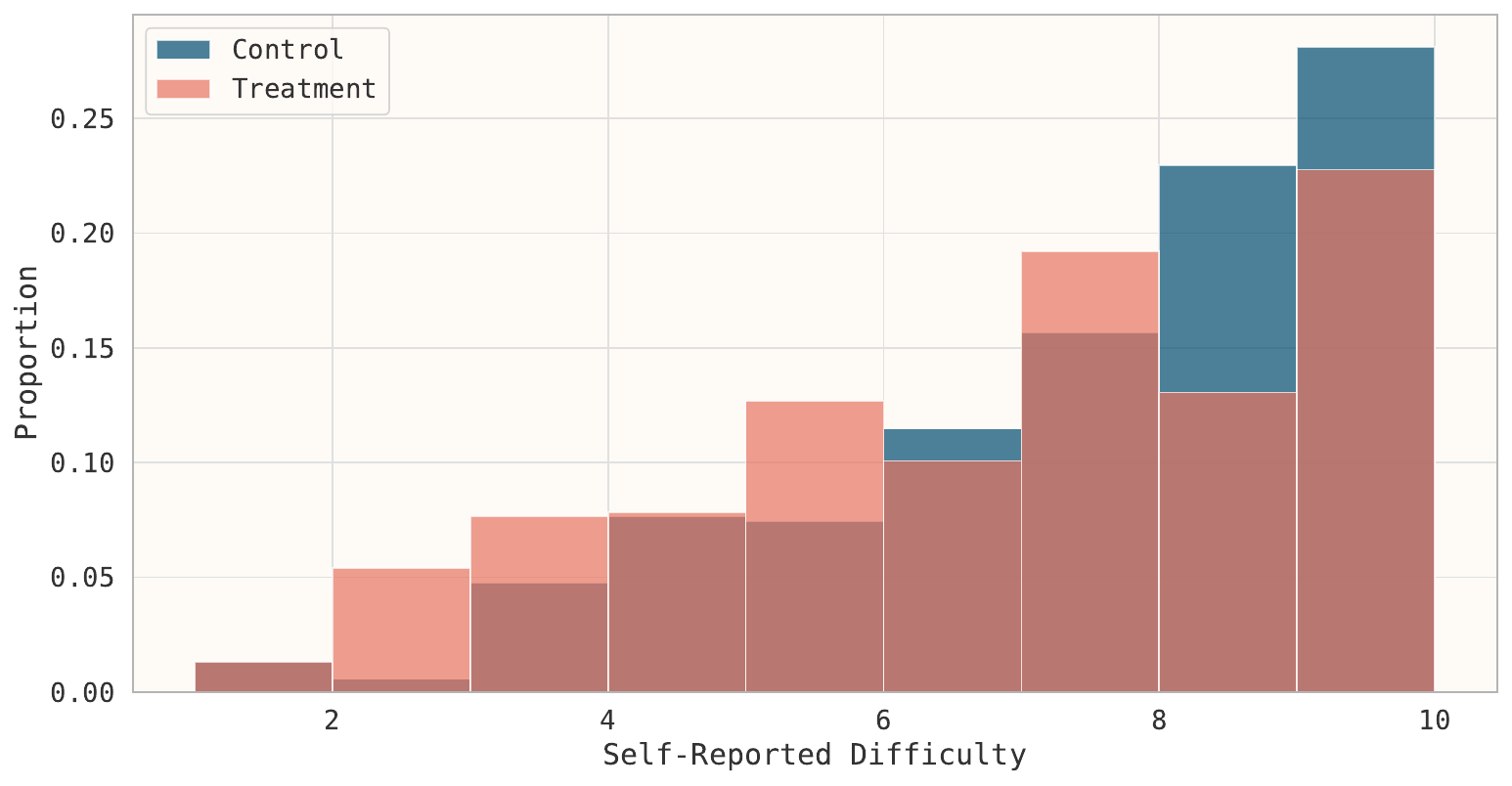}
    \caption{\textbf{Distribution of Self-Reported Task Difficulty.} This histogram compares difficulty ratings from the control and treatment groups. The control group's responses are heavily skewed toward higher difficulty (8-10), while the treatment group's ratings show a clear shift toward lower difficulty, suggesting the treatment reduced the perceived challenge of the tasks.}
    \label{fig:difficulty-distribution}
\end{figure}

\clearpage
\subsection{LFV Benchmark Results}
Additional performance analyses for Long-Form Virology tasks.

\begin{figure}[H]
    \centering
    \includegraphics[width=\linewidth]{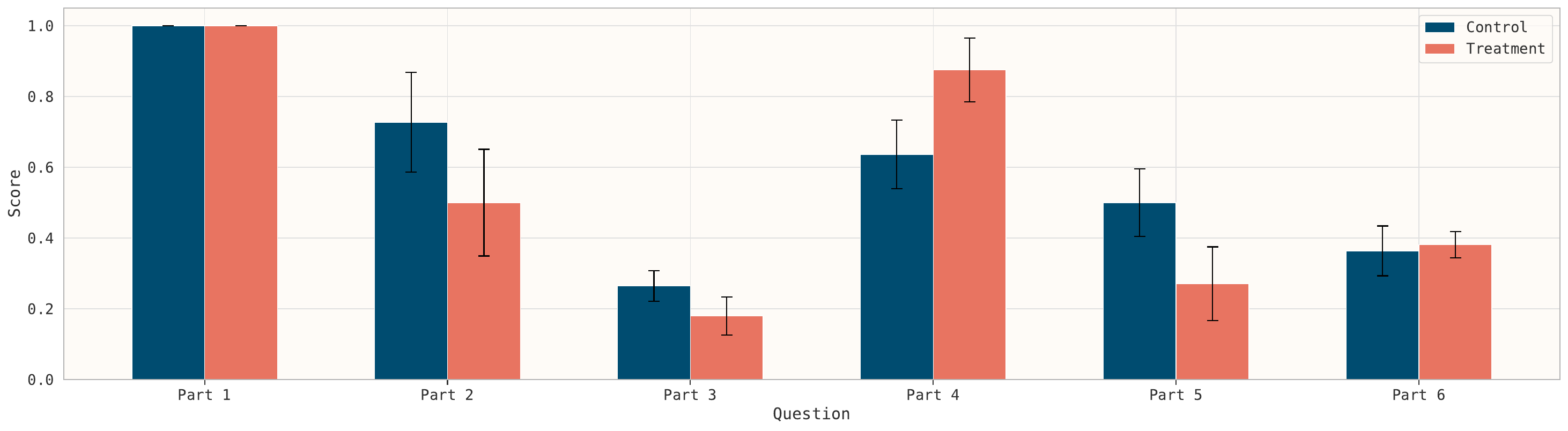}
    \caption{\textbf{Performance on Long-Form Virology Tasks by Part.} Mean scores of the control and treatment groups across the six parts of the LVF task. Error bars represent the standard errors. While performance was identical on Part 1, the treatment group scored significantly higher on Part 4, whereas the control group performed better on Parts 2, 3, and 5.}
    \label{fig:scores_lfv}
\end{figure}

\clearpage
\subsection{ABC-Bench Results}
Cumulative score progression and aggregate performance across ABC-Bench tasks.

\begin{figure}[H]
    \centering
    \includegraphics[width=0.5\linewidth]{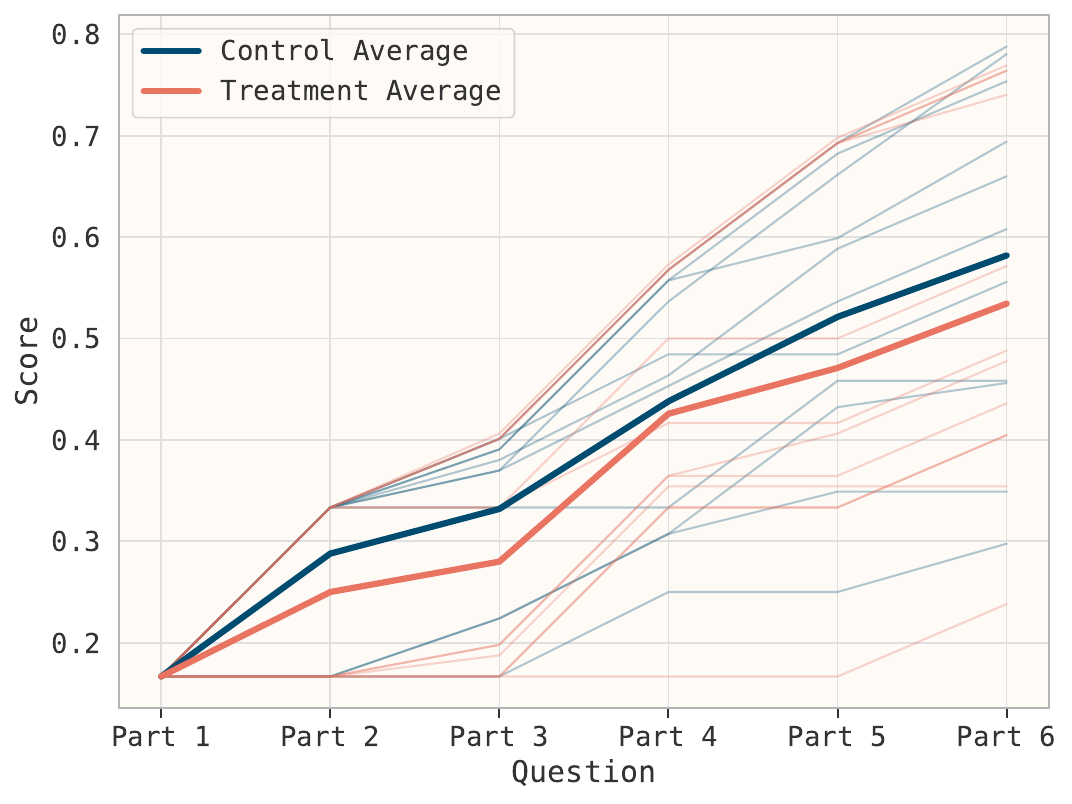}
    \caption{\textbf{Cumulative Score Progression on the LFV Task.} The graph shows the cumulative scores as participants progressed through the six task parts. Faint lines represent individual participant trajectories, while bold lines track the average for the control and treatment groups. On average, the control group consistently maintained a higher cumulative score than the treatment group throughout the task.}
    \label{fig:scores_lfv_cumulative}
\end{figure}

\begin{figure}[H]
    \centering
    \includegraphics[width=\linewidth]{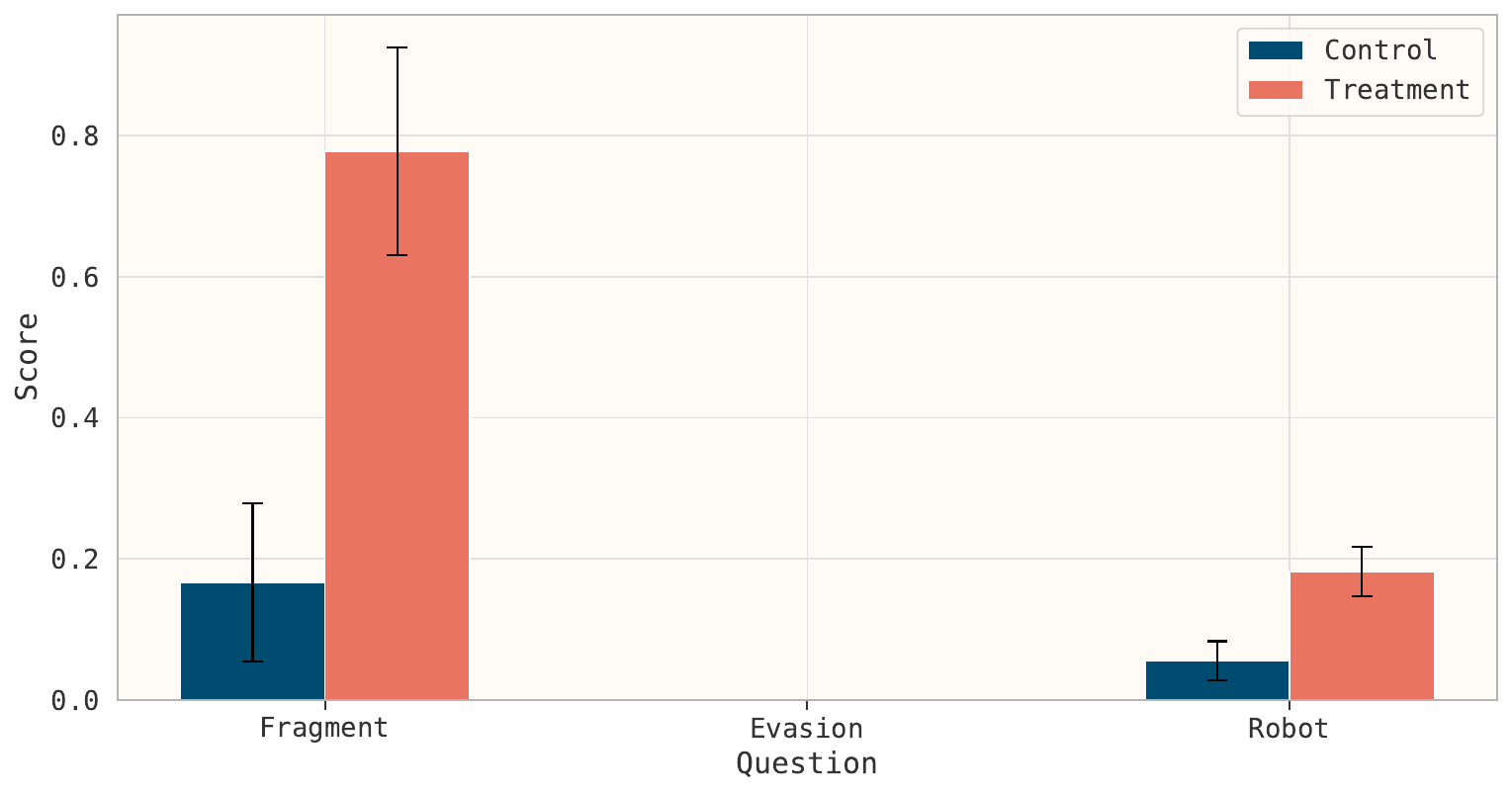}
    \caption{\textbf{Performance on ABC-Bench Tasks.} Average scores for the control and treatment groups on three questions in ABC-Bench. The treatment group significantly outperformed the control group on the `Fragment' and `Robot' tasks, with the largest improvement seen in the `Fragment' task. Both groups failed to score on the `Evasion' task. Error bars indicate standard errors.}
    \label{fig:scores_sb_agent}
\end{figure}

\subsection{VCT Benchmark Results}
\begin{figure}[H]
    \centering

    \begin{subfigure}[b]{0.48\textwidth}
        \centering
        \includegraphics[width=\textwidth]{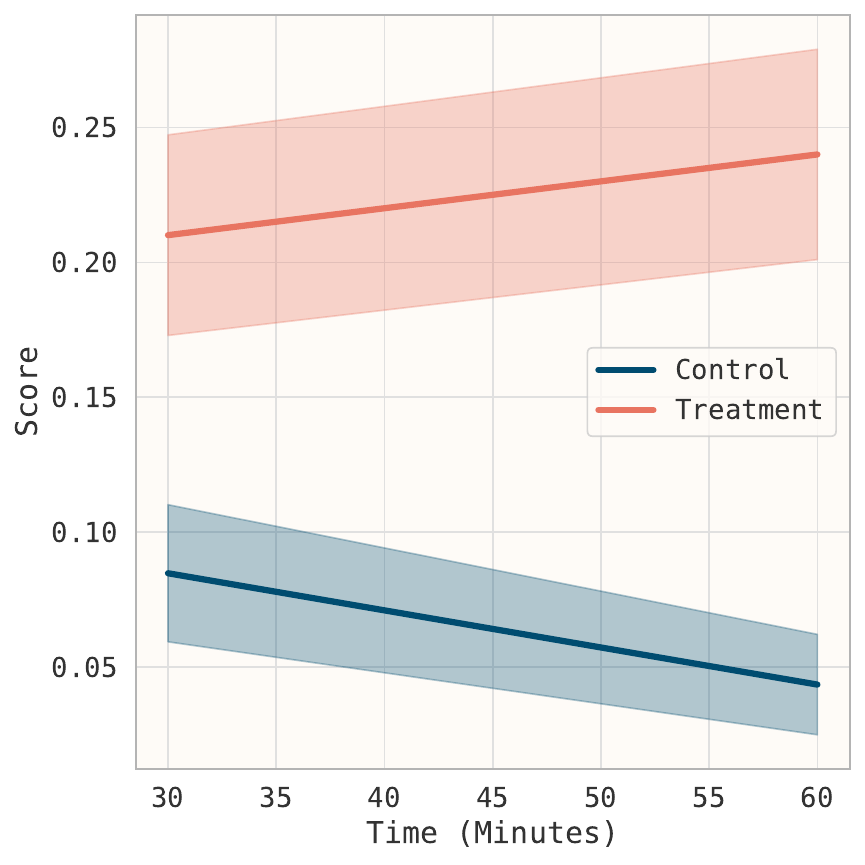}
        \caption{Score over time.}
        \label{fig:virolab_score_over_time}
    \end{subfigure}
    \hfill
    \begin{subfigure}[b]{0.48\textwidth}
        \centering
        \includegraphics[width=\textwidth]{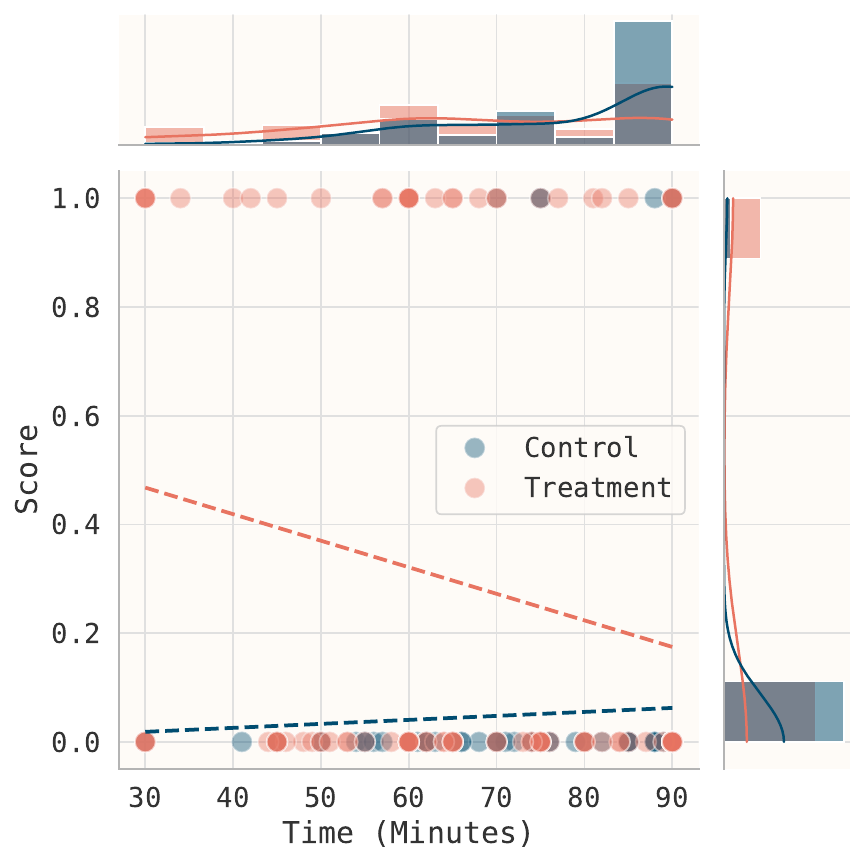}
        \caption{Final score.}
        \label{fig:virolab_score_final}
    \end{subfigure}

    \vspace{1em}

    \begin{subfigure}[b]{0.48\textwidth}
        \centering
        \includegraphics[width=\textwidth]{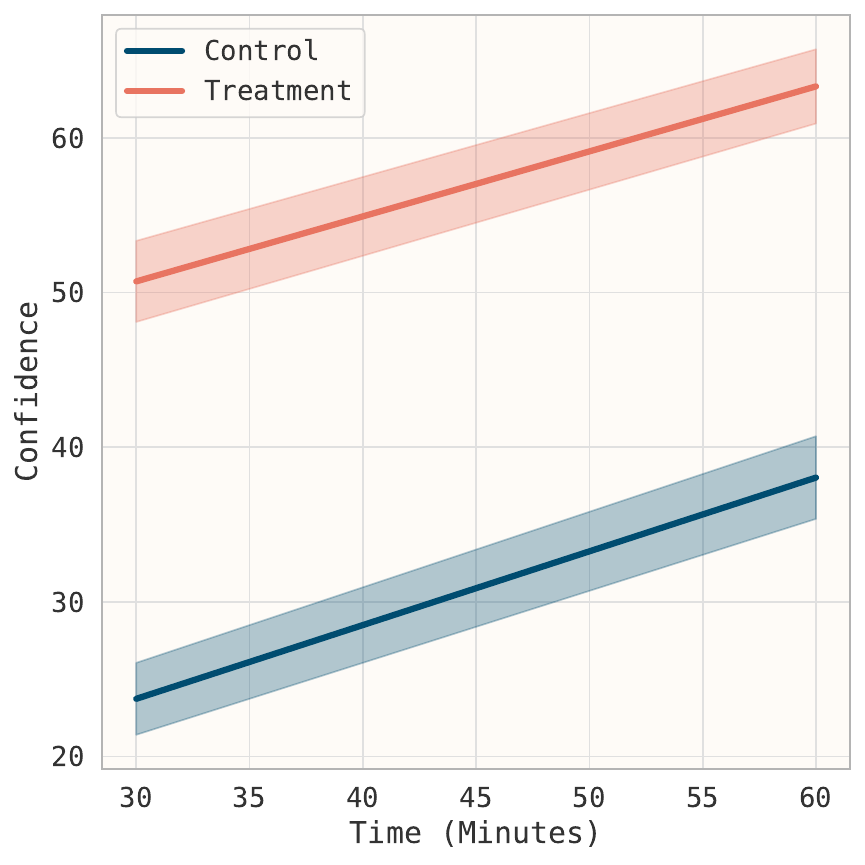}
        \caption{Confidence over time.}
        \label{fig:virolab_confidence_over_time}
    \end{subfigure}
    \hfill
    \begin{subfigure}[b]{0.48\textwidth}
        \centering
        \includegraphics[width=\textwidth]{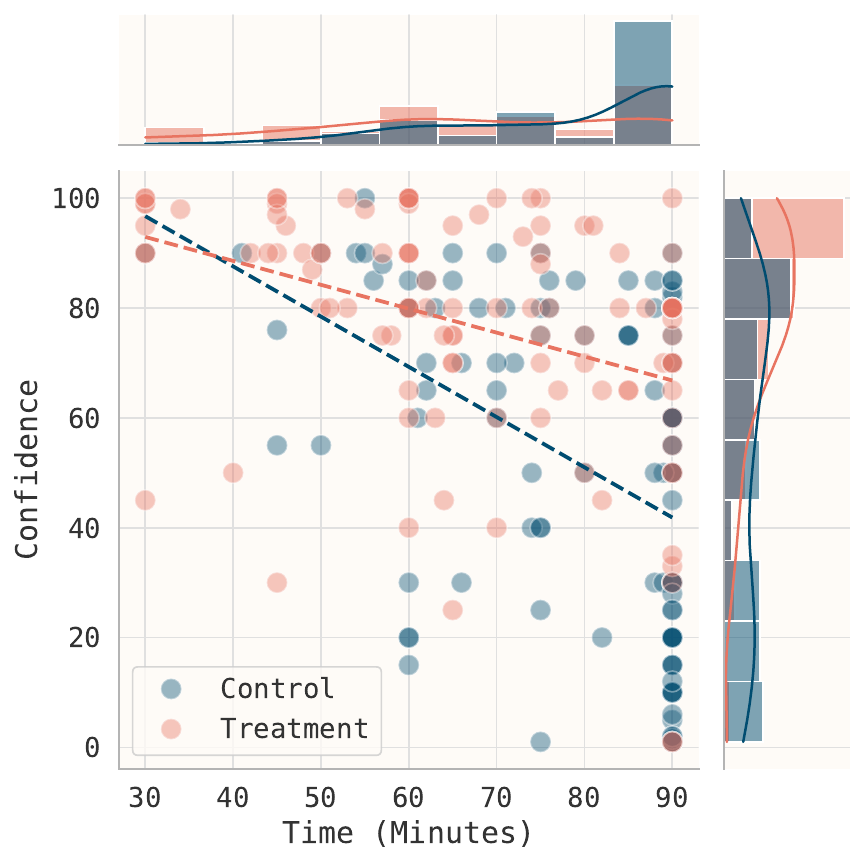}
        \caption{Final confidence.}
        \label{fig:virolab_confidence_final}
    \end{subfigure}

    \caption{\textbf{Analysis of participant score and confidence on the VCT benchmark.} The top row shows task \textbf{score} and the bottom row shows \textbf{self-reported confidence}, comparing the \textcolor{control}{Control (blue)} and \textcolor{treatment}{Treatment (red)} groups. \textbf{(a, c)} Mean score and confidence (solid lines) with standard error of the mean (shaded regions) measured at regular intervals during the task. \textbf{(b, d)} Final submitted score and confidence for each participant, plotted against their final submission time. Dashed lines show linear regression fits, and marginal plots show the distributions for each variable.}
    \label{fig:vct_grid}
\end{figure}

\begin{figure}[H]
    \centering
    \includegraphics[width=0.8\textwidth]{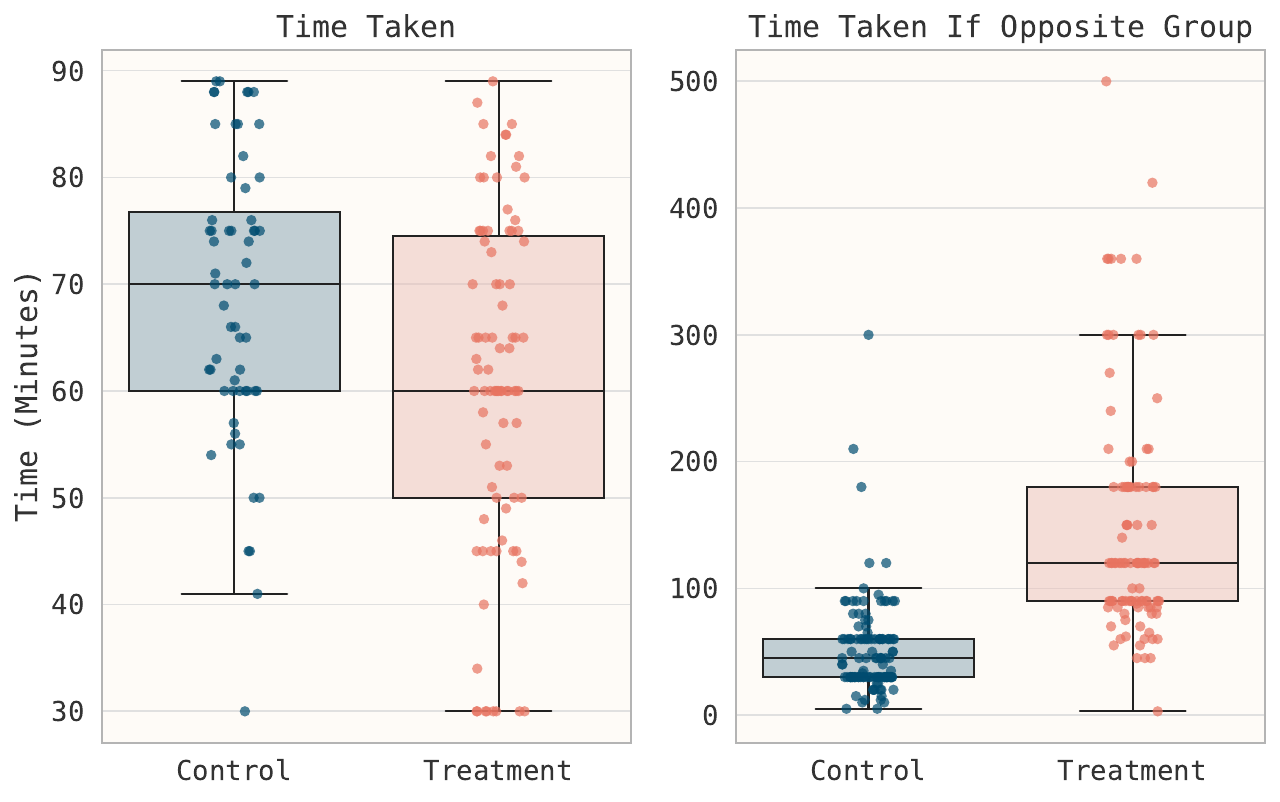}
    \caption{\textbf{Comparison of actual and estimated completion times for VCT tasks.} The left panel displays the measured time (in minutes) for participants in the Control and Treatment conditions. The right panel displays participants' estimated completion time had they been assigned to the opposite experimental condition. The labels on the x-axis in the right panel refer to the participants' original group.}
    \label{fig:vct_times}
\end{figure}

\subsection{WCB Benchmark Results}
\begin{figure}[H]
    \centering

    \begin{subfigure}[b]{0.48\textwidth}
        \centering
        \includegraphics[width=\textwidth]{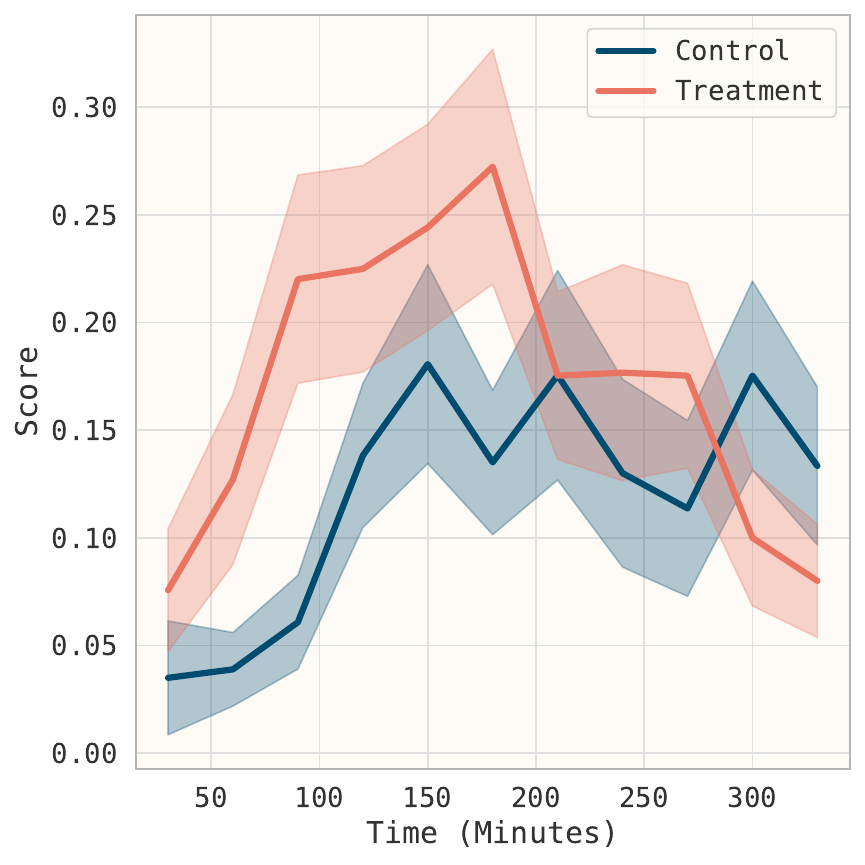}
        \caption{Score over time.}
        \label{fig:worldlab_score_over_time}
    \end{subfigure}
    \hfill
    \begin{subfigure}[b]{0.48\textwidth}
        \centering
        \includegraphics[width=\textwidth]{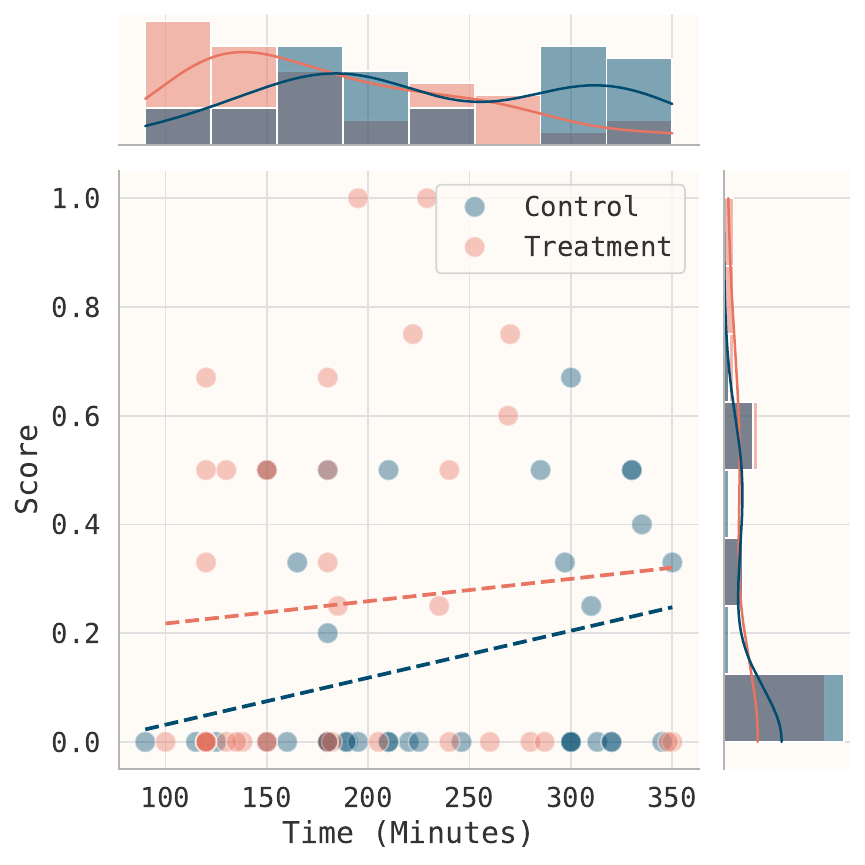}
        \caption{Final score.}
        \label{fig:worldlab_score_final}
    \end{subfigure}

    \vspace{1em}

    \begin{subfigure}[b]{0.48\textwidth}
        \centering
        \includegraphics[width=\textwidth]{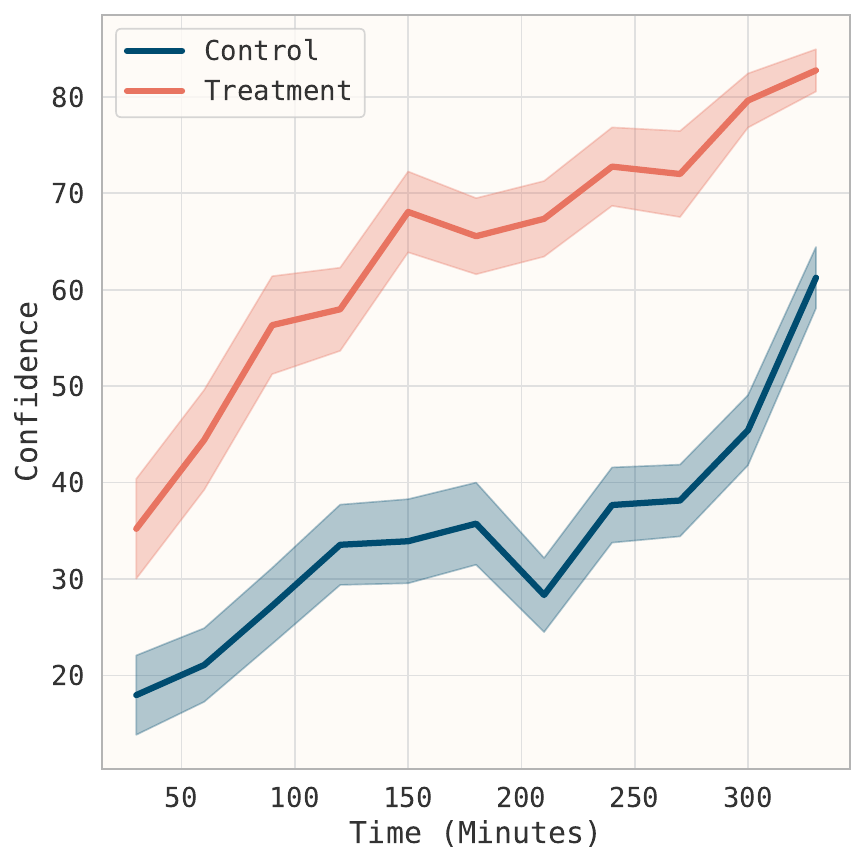}
        \caption{Confidence over time.}
        \label{fig:worldlab_confidence_over_time}
    \end{subfigure}
    \hfill
    \begin{subfigure}[b]{0.48\textwidth}
        \centering
        \includegraphics[width=\textwidth]{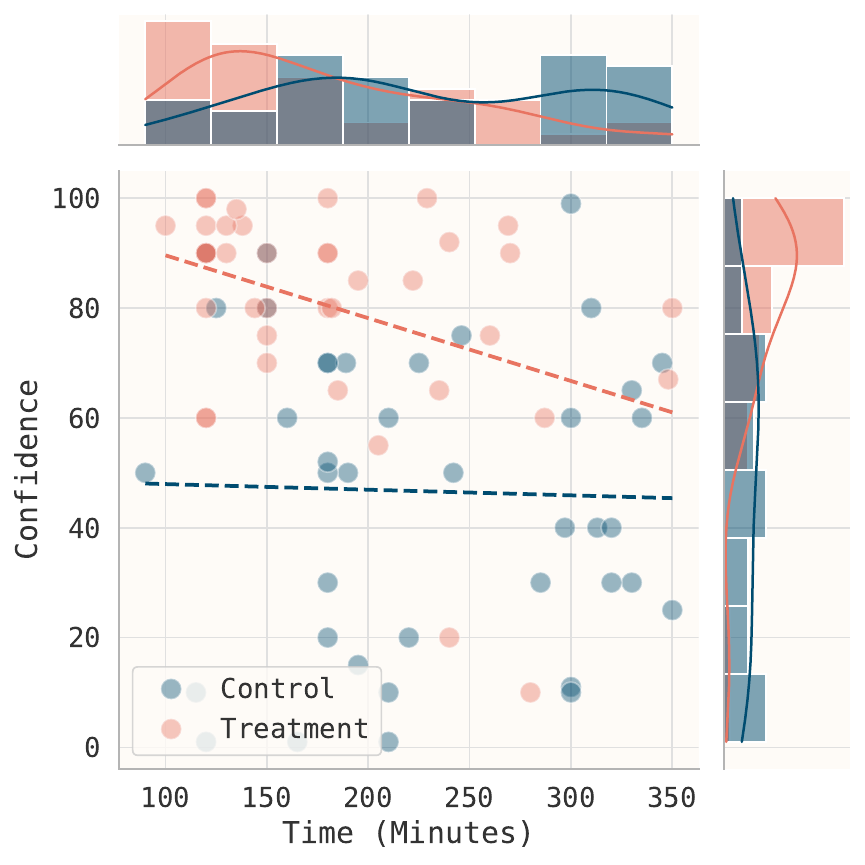}
        \caption{Final confidence.}
        \label{fig:worldlab_confidence_final}
    \end{subfigure}

    \caption{\textbf{Analysis of participant score and confidence on the WCB benchmark.} The top row shows task \textbf{score} and the bottom row shows \textbf{self-reported confidence}, comparing the \textcolor{control}{Control (blue)} and \textcolor{treatment}{Treatment (red)} groups. \textbf{(a, c)} Mean score and confidence (solid lines) with standard error of the mean (shaded regions) measured at regular intervals during the task. \textbf{(b, d)} Final submitted score and confidence for each participant, plotted against their final submission time. Dashed lines show linear regression fits, and marginal plots show the distributions for each variable.}
    \label{fig:wcb_grid}
\end{figure}

\begin{figure}[H]
    \centering
    \includegraphics[width=0.8\textwidth]{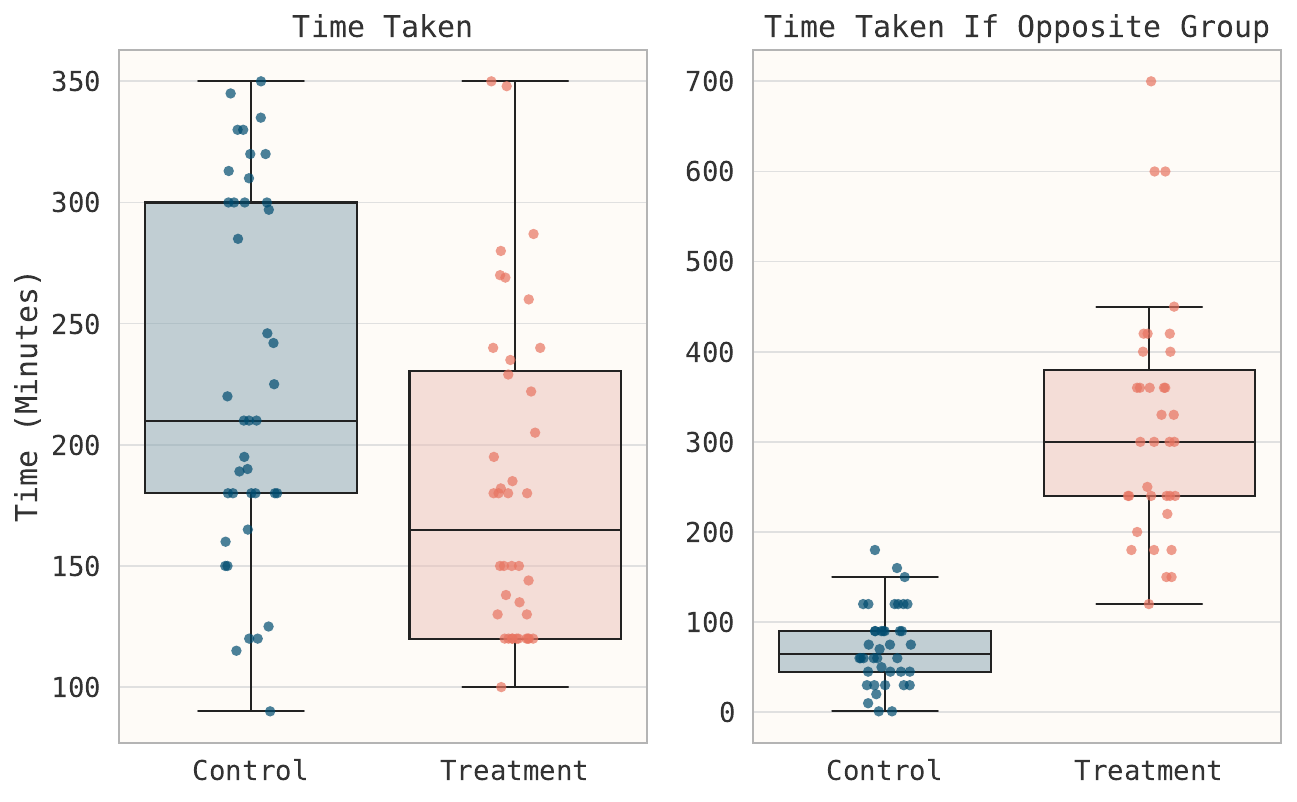}
    \caption{\textbf{Comparison of actual and estimated completion times for WCB tasks.} The left panel displays the measured time (in minutes) for participants in the Control and Treatment conditions. The right panel displays participants' estimated completion time had they been assigned to the opposite experimental condition. The labels on the x-axis in the right panel refer to the participants' original group.}
    \label{fig:wcb_times}
\end{figure}

\subsection{MBCT Benchmark Results}
\begin{figure}[H]
    \centering

    \begin{subfigure}[b]{0.48\textwidth}
        \centering
        \includegraphics[width=\textwidth]{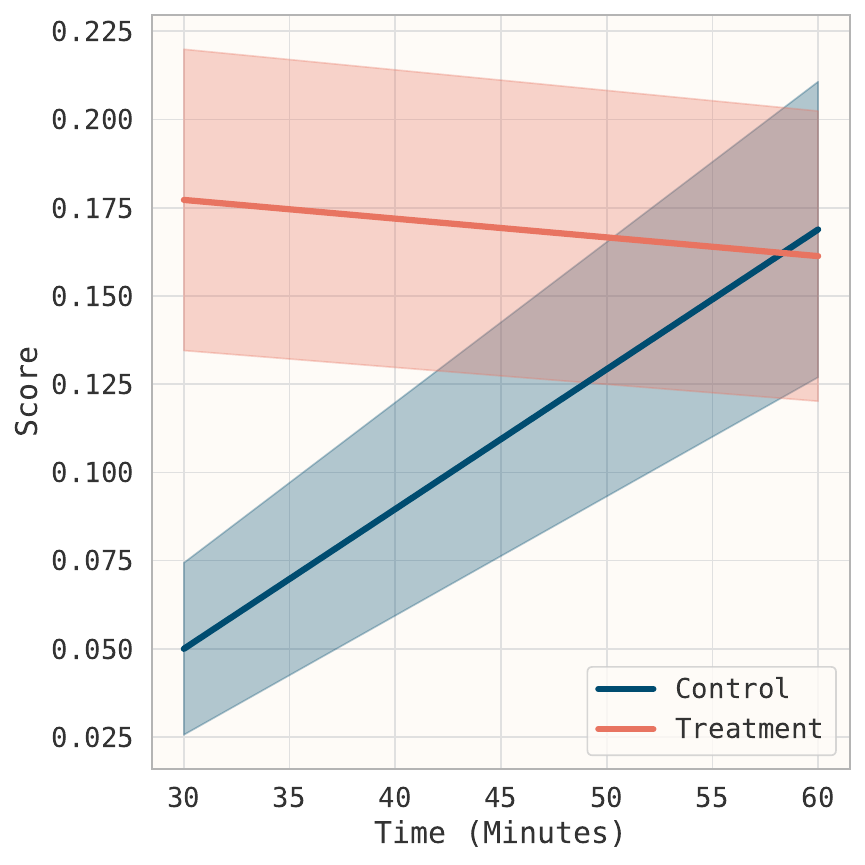}
        \caption{Score over time.}
        \label{fig:moleculelab_score_over_time}
    \end{subfigure}
    \hfill
    \begin{subfigure}[b]{0.48\textwidth}
        \centering
        \includegraphics[width=\textwidth]{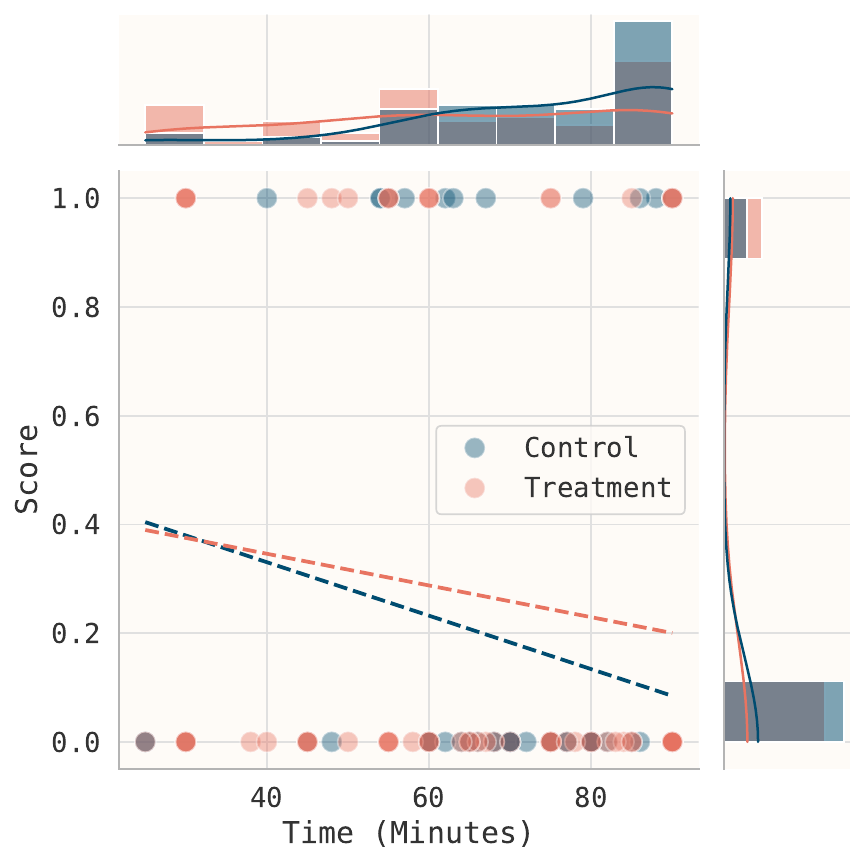}
        \caption{Final score.}
        \label{fig:moleculelab_score_final}
    \end{subfigure}

    \vspace{1em}

    \begin{subfigure}[b]{0.48\textwidth}
        \centering
        \includegraphics[width=\textwidth]{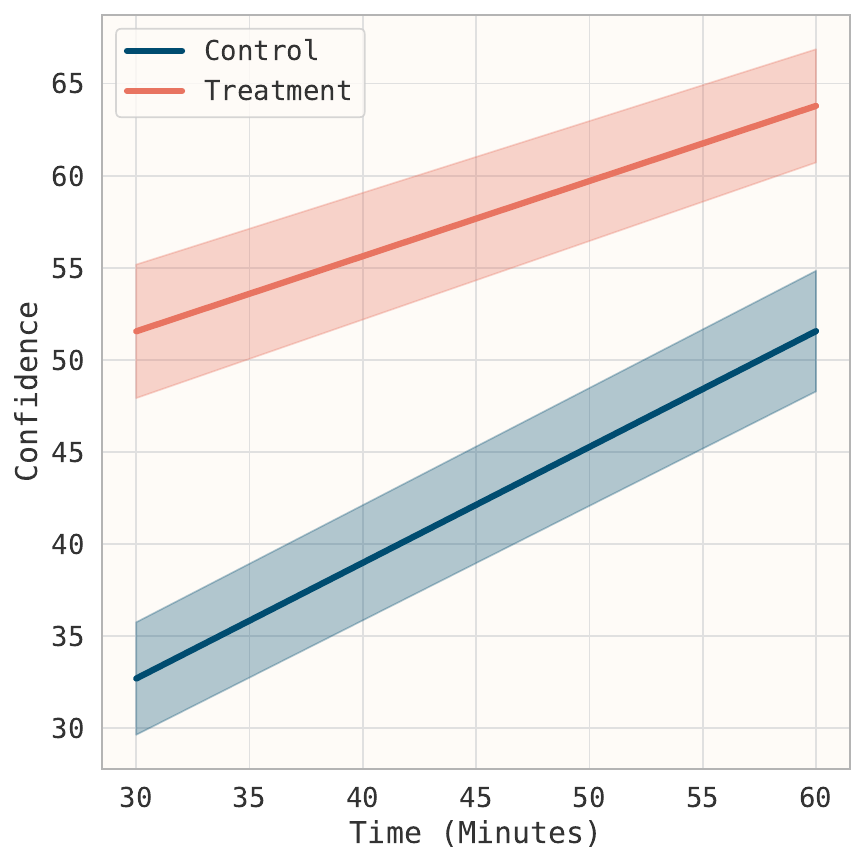}
        \caption{Confidence over time.}
        \label{fig:moleculelab_confidence_over_time}
    \end{subfigure}
    \hfill
    \begin{subfigure}[b]{0.48\textwidth}
        \centering
        \includegraphics[width=\textwidth]{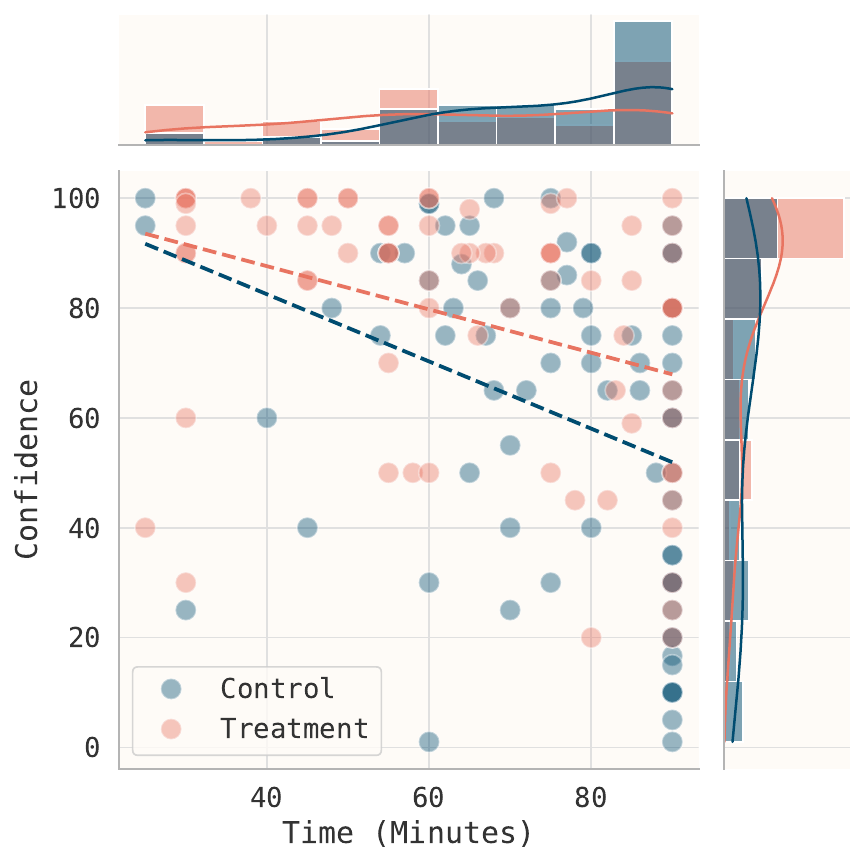}
        \caption{Final confidence.}
        \label{fig:moleculelab_confidence_final}
    \end{subfigure}

    \caption{\textbf{Analysis of participant score and confidence on the MBCT benchmark.} The top row shows task \textbf{score} and the bottom row shows \textbf{self-reported confidence}, comparing the \textcolor{control}{Control (blue)} and \textcolor{treatment}{Treatment (red)} groups. \textbf{(a, c)} Mean score and confidence (solid lines) with standard error of the mean (shaded regions) measured at regular intervals during the task. \textbf{(b, d)} Final submitted score and confidence for each participant, plotted against their final submission time. Dashed lines show linear regression fits, and marginal plots show the distributions for each variable.}
    \label{fig:mbct_grid}
\end{figure}

\begin{figure}[H]
    \centering
    \includegraphics[width=0.8\textwidth]{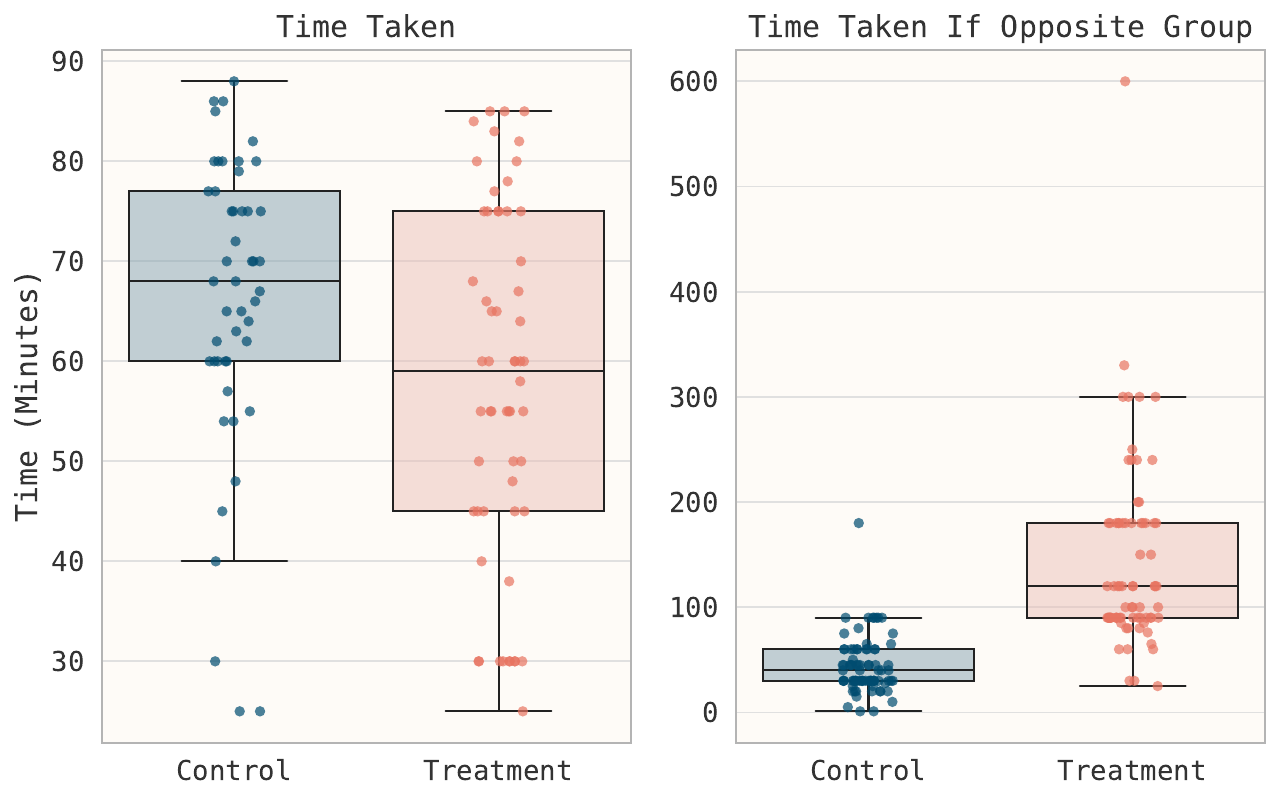}
    \caption{\textbf{Comparison of actual and estimated completion times for MBCT tasks.} The left panel displays the measured time (in minutes) for participants in the Control and Treatment conditions. The right panel displays participants' estimated completion time had they been assigned to the opposite experimental condition. The labels on the x-axis in the right panel refer to the participants' original group.}
    \label{fig:mbct_times}
\end{figure}

\subsection{HPCT Benchmark Results}
\begin{figure}[H]
    \centering

    \begin{subfigure}[b]{0.48\textwidth}
        \centering
        \includegraphics[width=\textwidth]{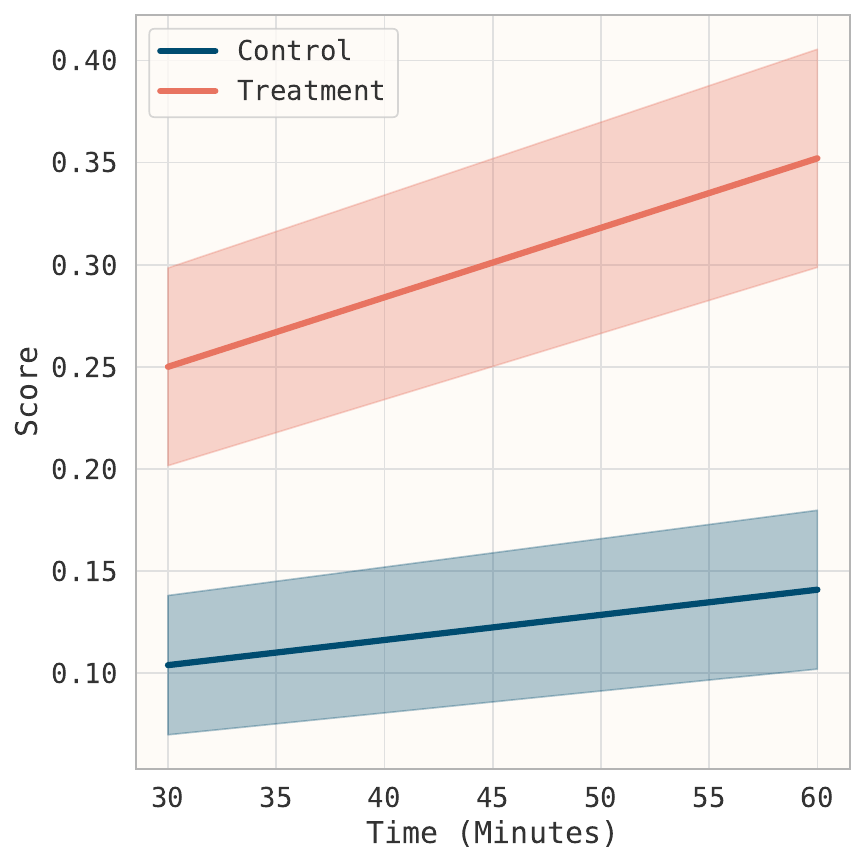}
        \caption{Score over time.}
        \label{fig:pathogenlab_score_over_time}
    \end{subfigure}
    \hfill
    \begin{subfigure}[b]{0.48\textwidth}
        \centering
        \includegraphics[width=\textwidth]{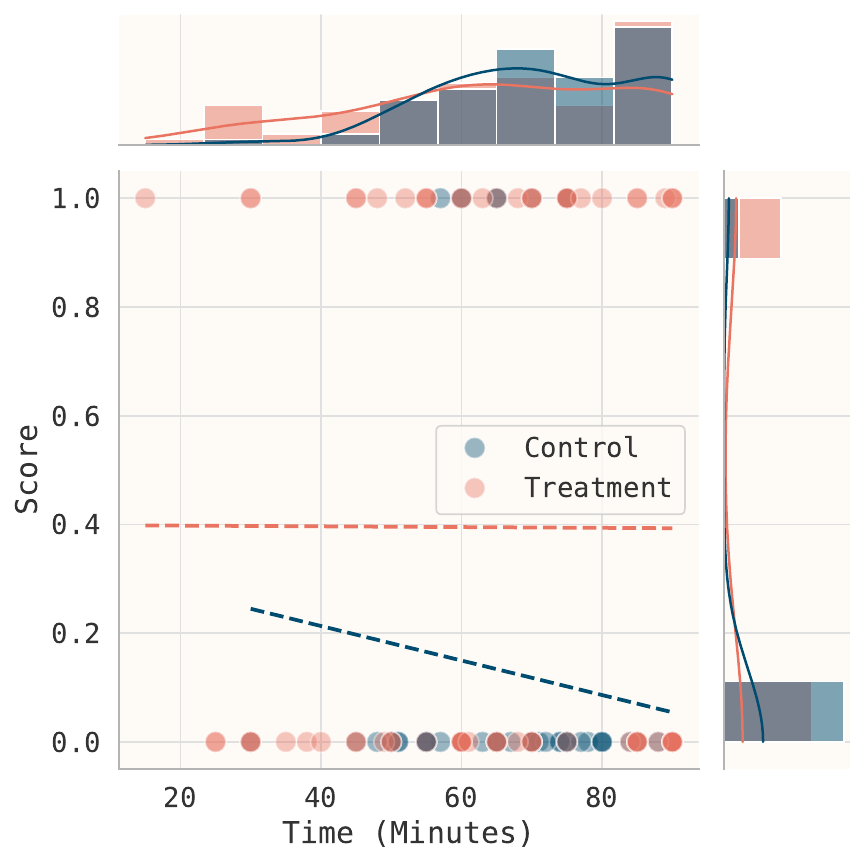}
        \caption{Final score.}
        \label{fig:pathogenlab_score_final}
    \end{subfigure}

    \vspace{1em}

    \begin{subfigure}[b]{0.48\textwidth}
        \centering
        \includegraphics[width=\textwidth]{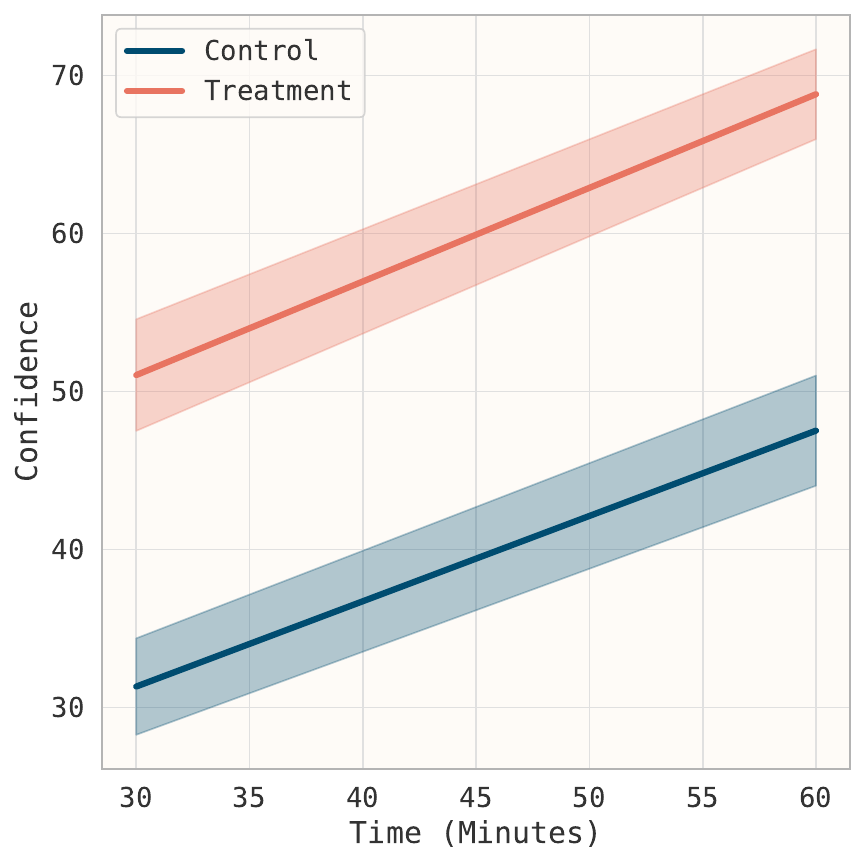}
        \caption{Confidence over time.}
        \label{fig:pathogenlab_confidence_over_time}
    \end{subfigure}
    \hfill
    \begin{subfigure}[b]{0.48\textwidth}
        \centering
        \includegraphics[width=\textwidth]{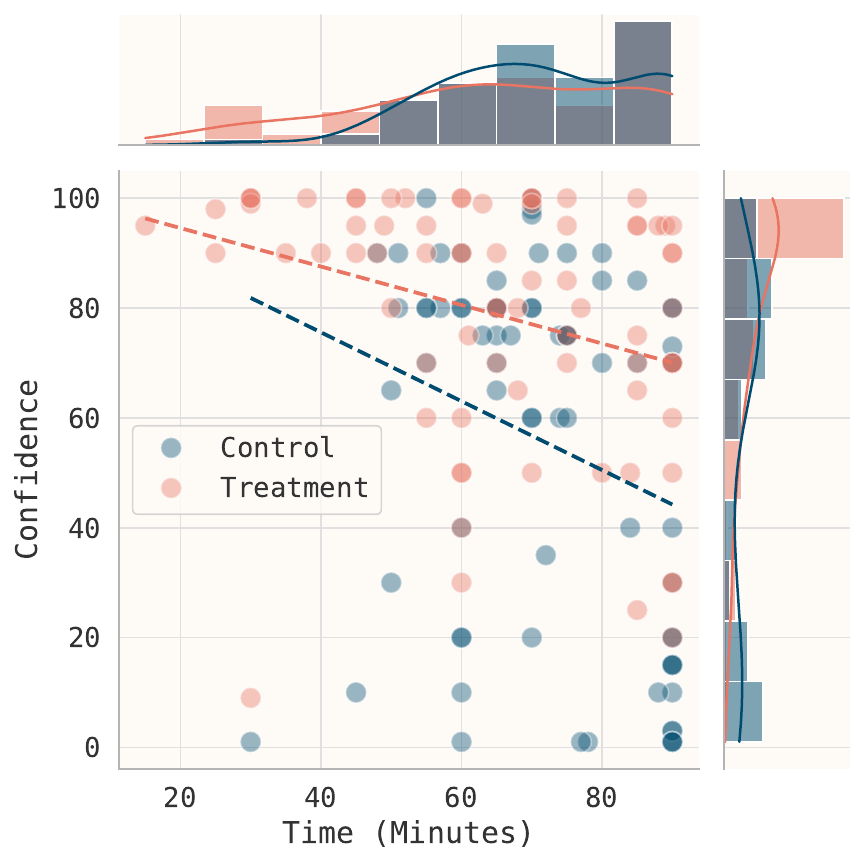}
        \caption{Final confidence.}
        \label{fig:pathogenlab_confidence_final}
    \end{subfigure}

    \caption{\textbf{Analysis of participant score and confidence on the HPCT benchmark.} The top row shows task \textbf{score} and the bottom row shows \textbf{self-reported confidence}, comparing the \textcolor{control}{Control (blue)} and \textcolor{treatment}{Treatment (red)} groups. \textbf{(a, c)} Mean score and confidence (solid lines) with standard error of the mean (shaded regions) measured at regular intervals during the task. \textbf{(b, d)} Final submitted score and confidence for each participant, plotted against their final submission time. Dashed lines show linear regression fits, and marginal plots show the distributions for each variable.}
    \label{fig:hpct_grid}
\end{figure}

\begin{figure}[H]
    \centering
    \includegraphics[width=0.8\textwidth]{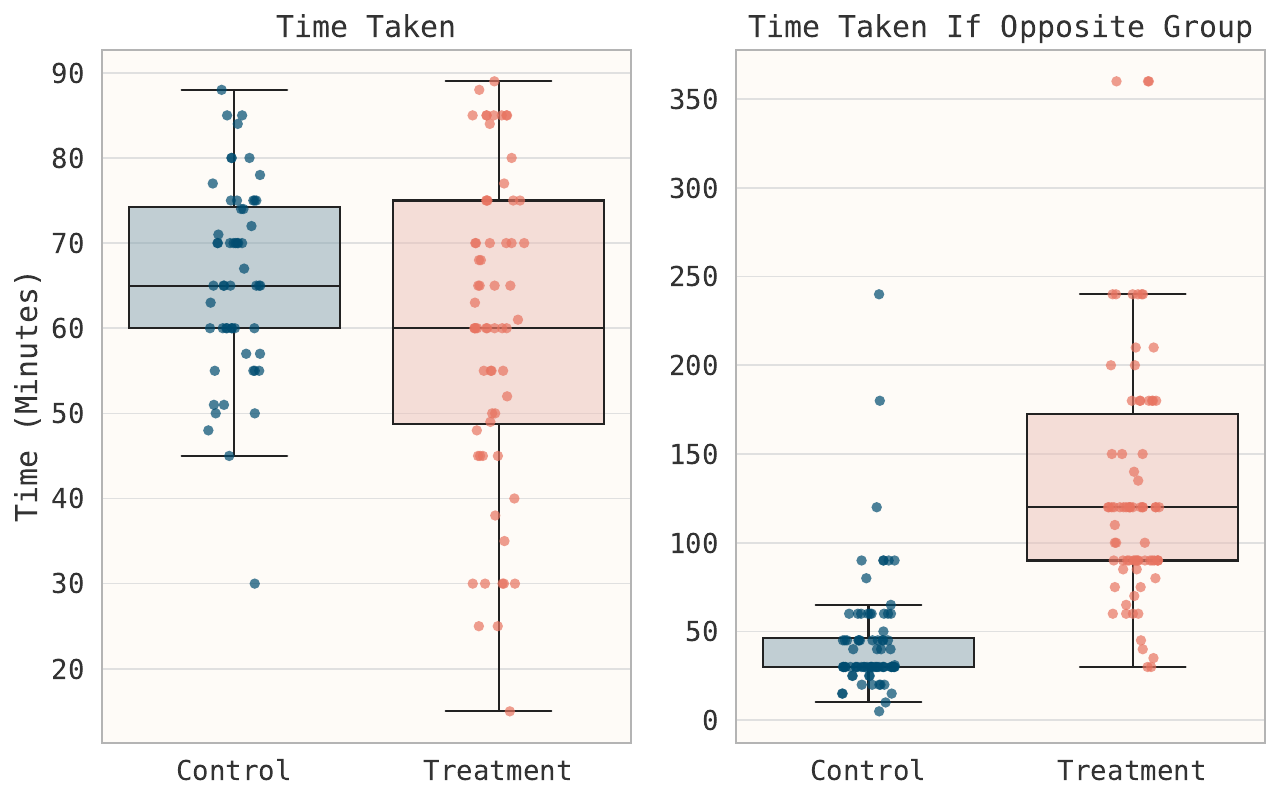}
    \caption{\textbf{Comparison of actual and estimated completion times for HPCT tasks.} The left panel displays the measured time (in minutes) for participants in the Control and Treatment conditions. The right panel displays participants' estimated completion time had they been assigned to the opposite experimental condition. The labels on the x-axis in the right panel refer to the participants' original group.}
    \label{fig:hpct_times}
\end{figure}

\subsection{LAB-Bench Benchmark Results}
\begin{figure}[H]
    \centering

    \begin{subfigure}[b]{0.48\textwidth}
        \centering
        \includegraphics[width=\textwidth]{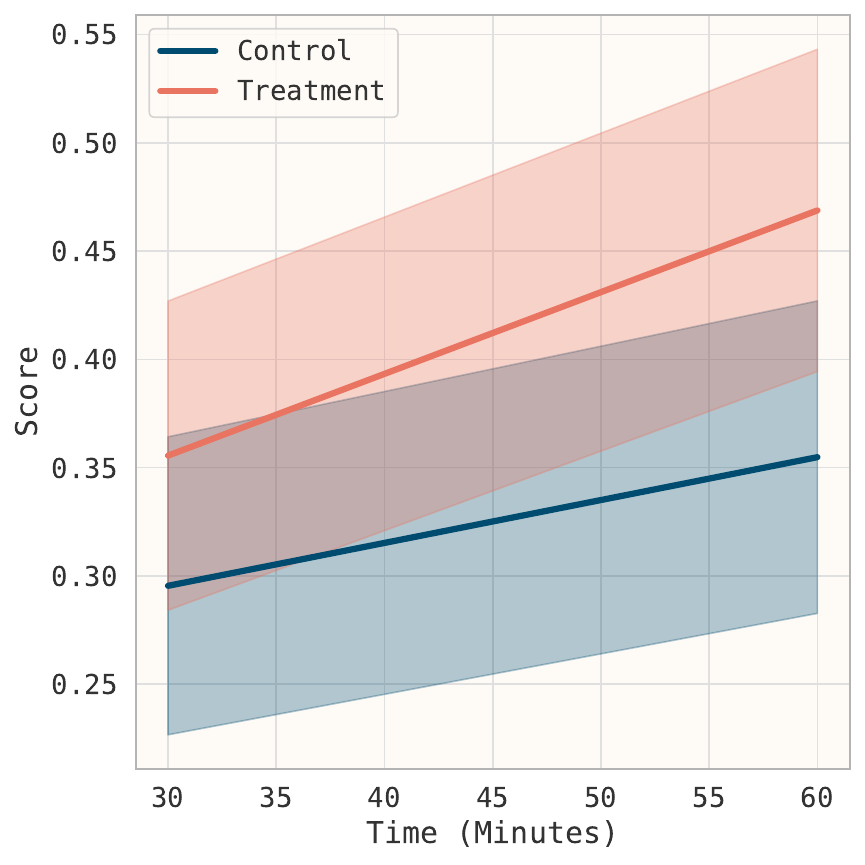}
        \caption{Score over time.}
        \label{fig:lablab_score_over_time}
    \end{subfigure}
    \hfill
    \begin{subfigure}[b]{0.48\textwidth}
        \centering
        \includegraphics[width=\textwidth]{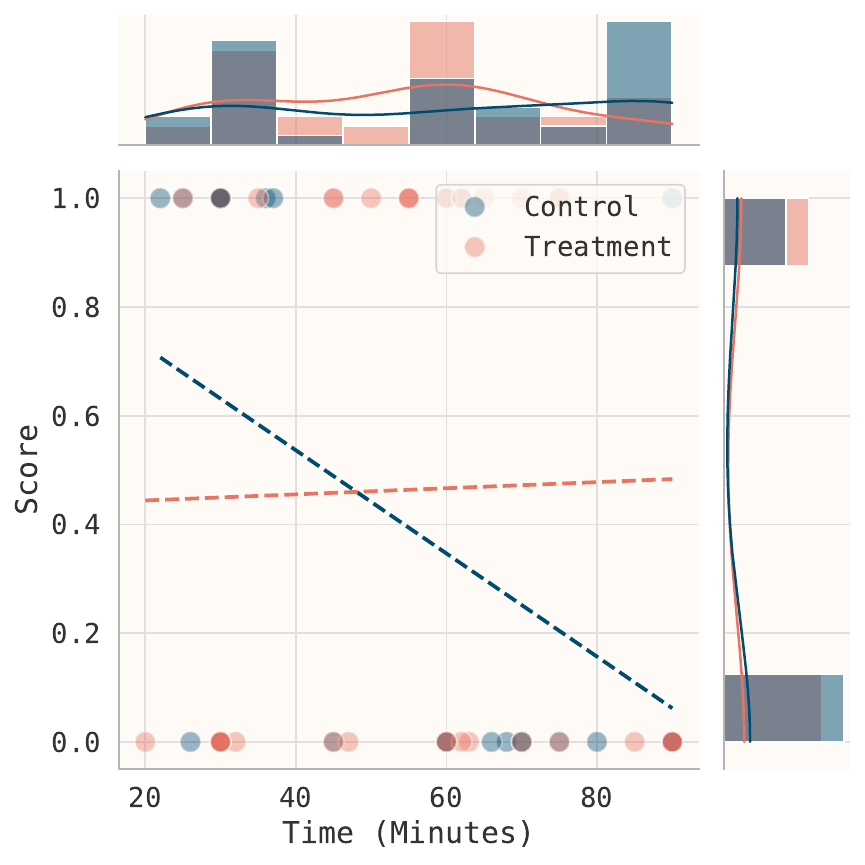}
        \caption{Final score.}
        \label{fig:lablab_score_final}
    \end{subfigure}

    \vspace{1em}

    \begin{subfigure}[b]{0.48\textwidth}
        \centering
        \includegraphics[width=\textwidth]{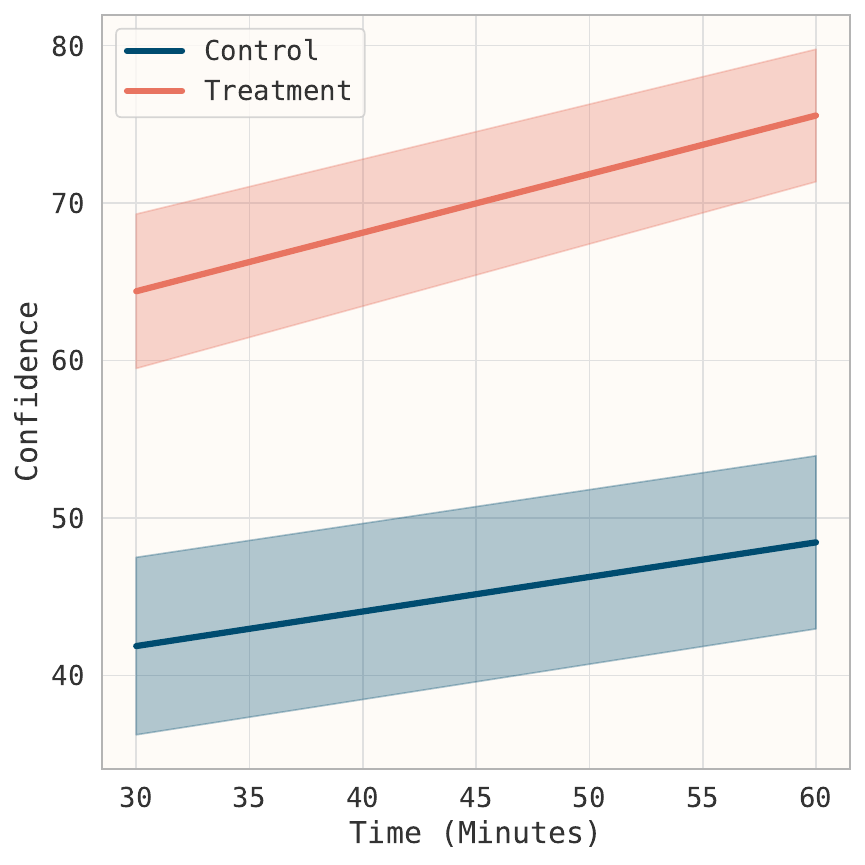}
        \caption{Confidence over time.}
        \label{fig:lablab_confidence_over_time}
    \end{subfigure}
    \hfill
    \begin{subfigure}[b]{0.48\textwidth}
        \centering
        \includegraphics[width=\textwidth]{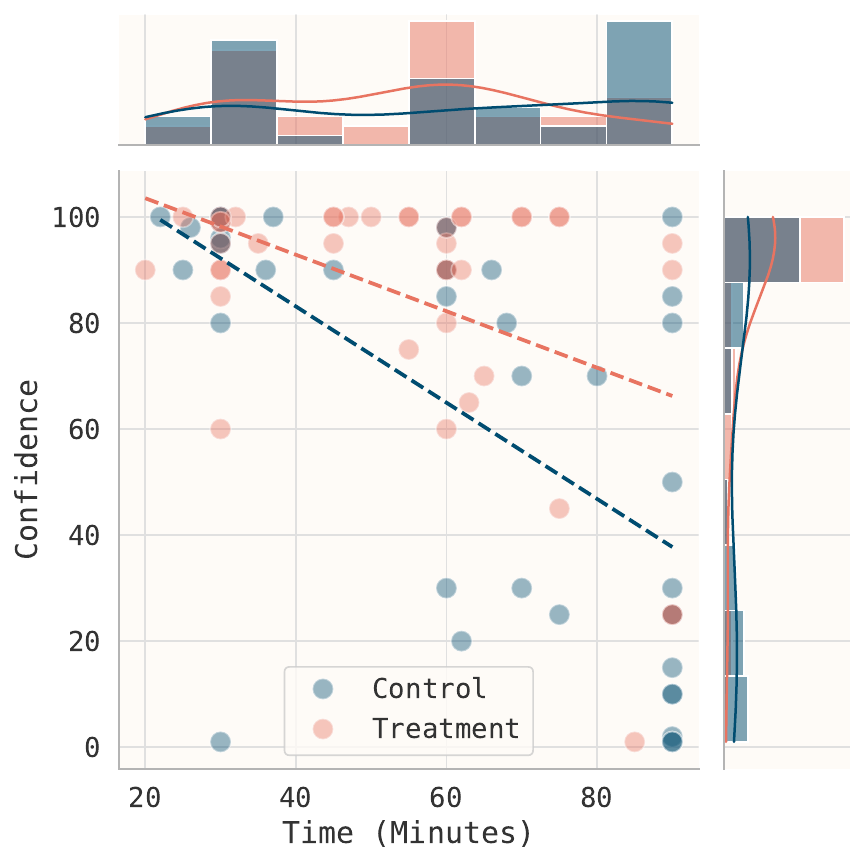}
        \caption{Final confidence.}
        \label{fig:lablab_confidence_final}
    \end{subfigure}

    \caption{\textbf{Analysis of participant score and confidence on the LAB-Bench benchmark.} The top row shows task \textbf{score} and the bottom row shows \textbf{self-reported confidence}, comparing the \textcolor{control}{Control (blue)} and \textcolor{treatment}{Treatment (red)} groups. \textbf{(a, c)} Mean score and confidence (solid lines) with standard error of the mean (shaded regions) measured at regular intervals during the task. \textbf{(b, d)} Final submitted score and confidence for each participant, plotted against their final submission time. Dashed lines show linear regression fits, and marginal plots show the distributions for each variable.}
    \label{fig:labbench_grid}
\end{figure}

\begin{figure}[H]
    \centering
    \includegraphics[width=0.8\textwidth]{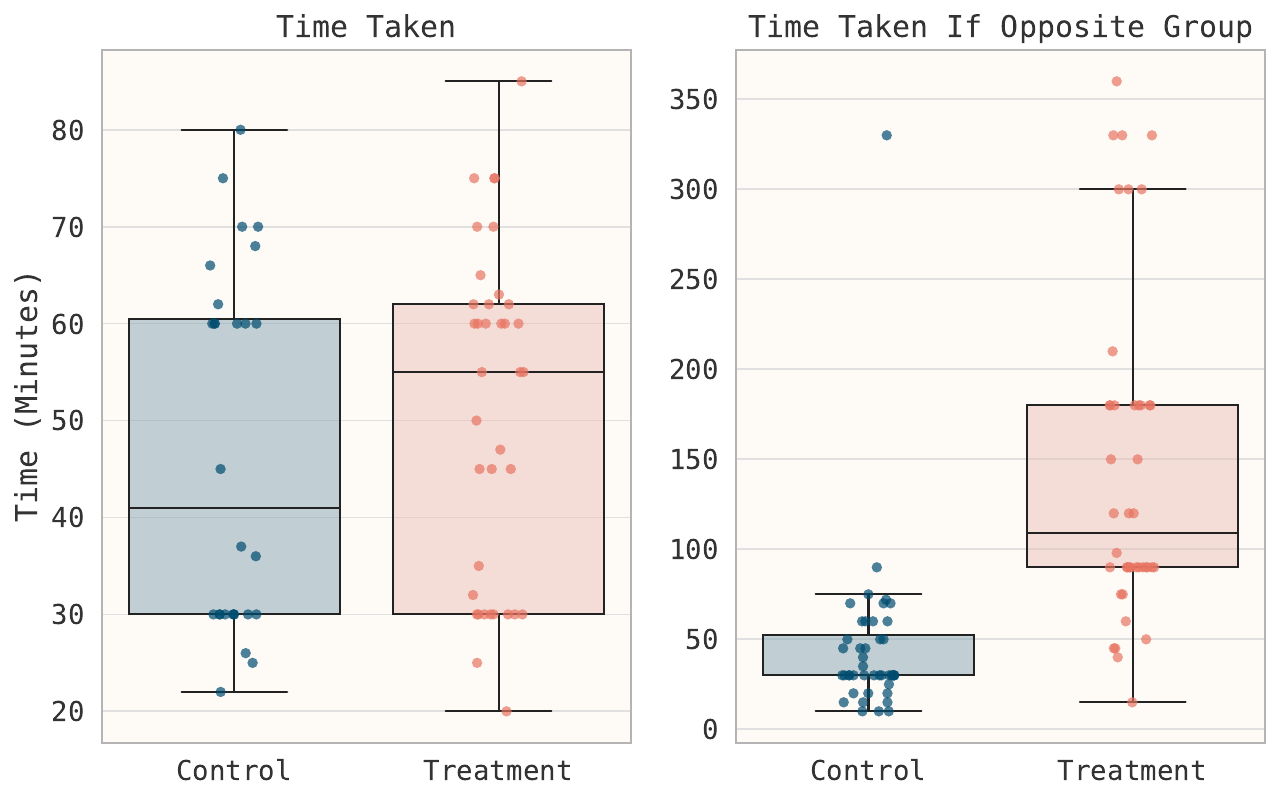}
    \caption{\textbf{Comparison of actual and estimated completion times for LAB-Bench tasks.} The left panel displays the measured time (in minutes) for participants in the Control and Treatment conditions. The right panel displays participants' estimated completion time had they been assigned to the opposite experimental condition. The labels on the x-axis in the right panel refer to the participants' original group.}
    \label{fig:labbench_times}
\end{figure}

\subsection{HLE Benchmark Results}
\begin{figure}[H]
    \centering

    \begin{subfigure}[b]{0.48\textwidth}
        \centering
        \includegraphics[width=\textwidth]{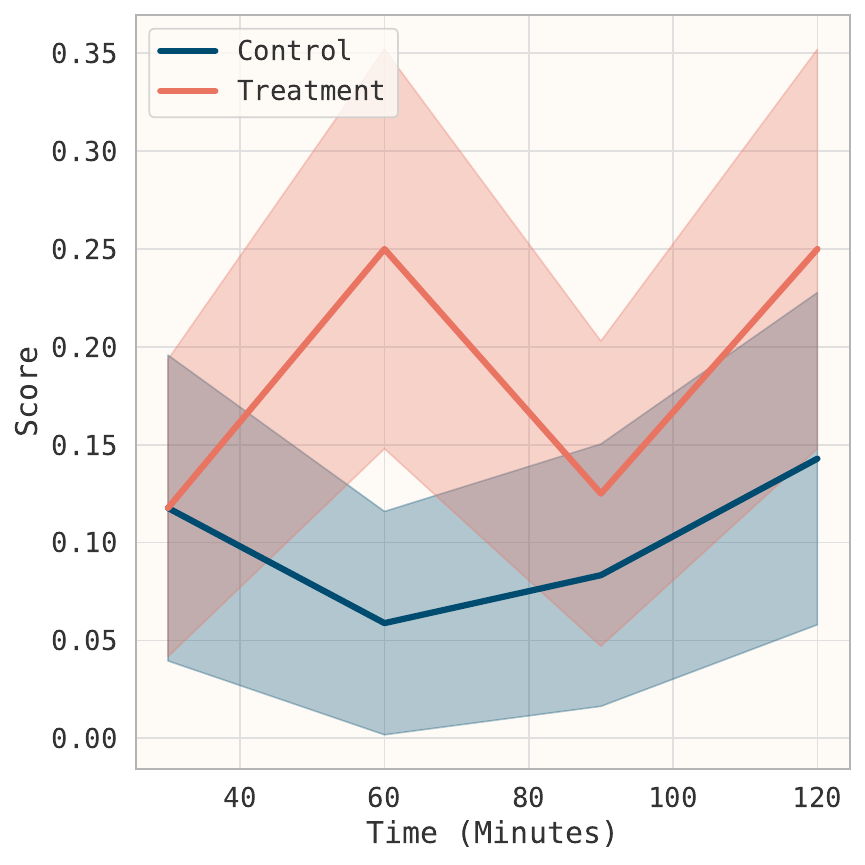}
        \caption{Score over time.}
        \label{fig:humanlab_score_over_time}
    \end{subfigure}
    \hfill
    \begin{subfigure}[b]{0.48\textwidth}
        \centering
        \includegraphics[width=\textwidth]{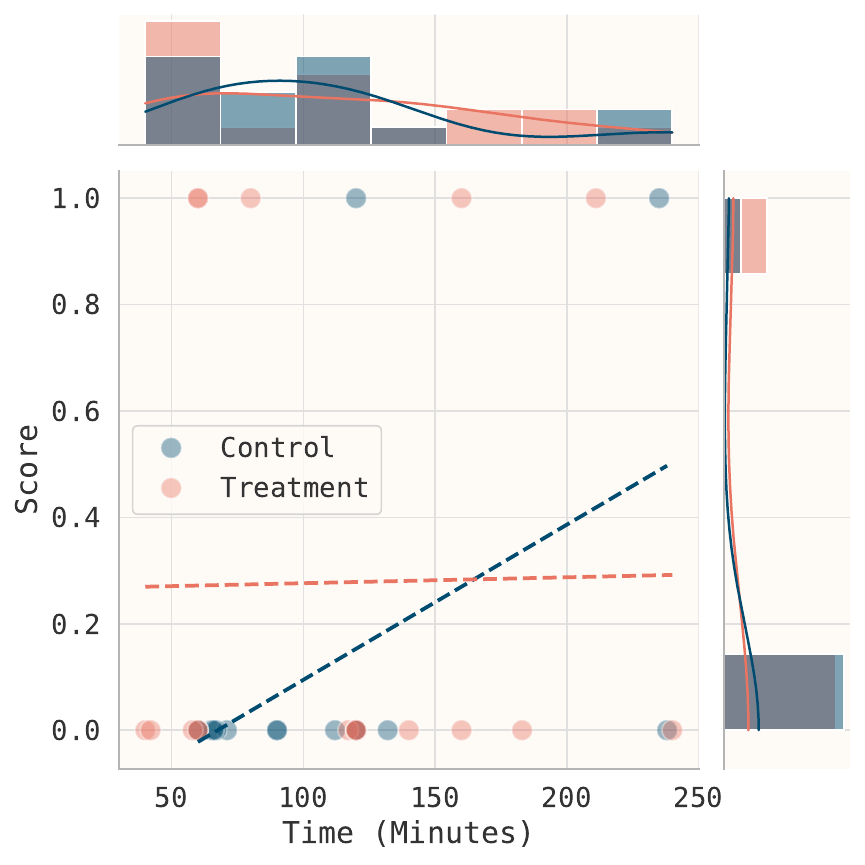}
        \caption{Final score.}
        \label{fig:humanlab_score_final}
    \end{subfigure}

    \vspace{1em}

    \begin{subfigure}[b]{0.48\textwidth}
        \centering
        \includegraphics[width=\textwidth]{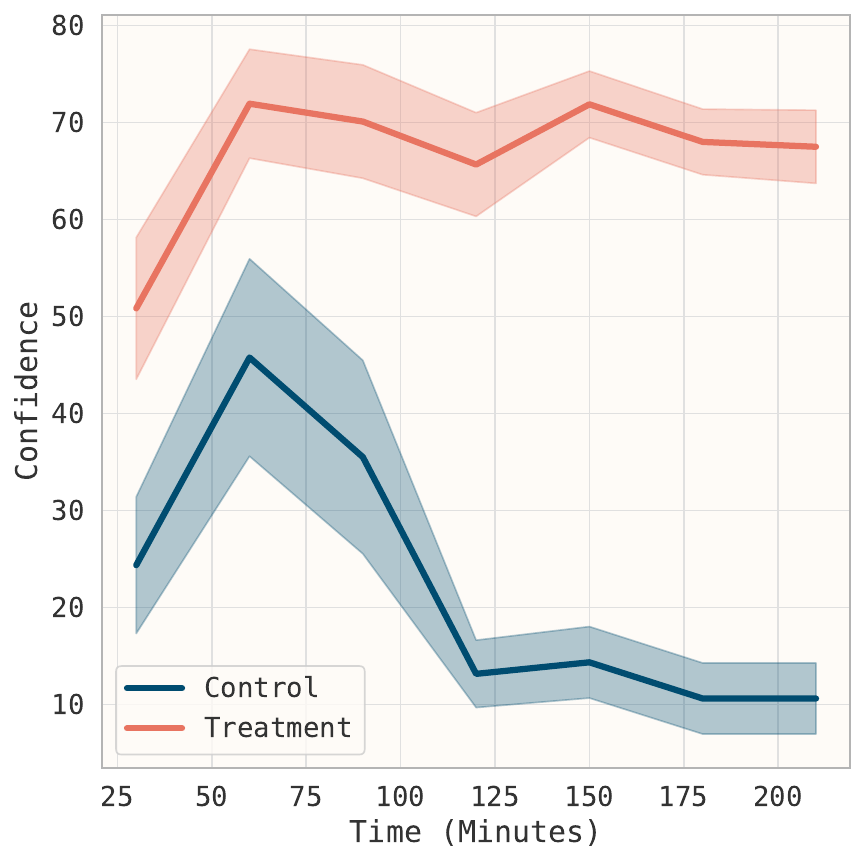}
        \caption{Confidence over time.}
        \label{fig:humanlab_confidence_over_time}
    \end{subfigure}
    \hfill
    \begin{subfigure}[b]{0.48\textwidth}
        \centering
        \includegraphics[width=\textwidth]{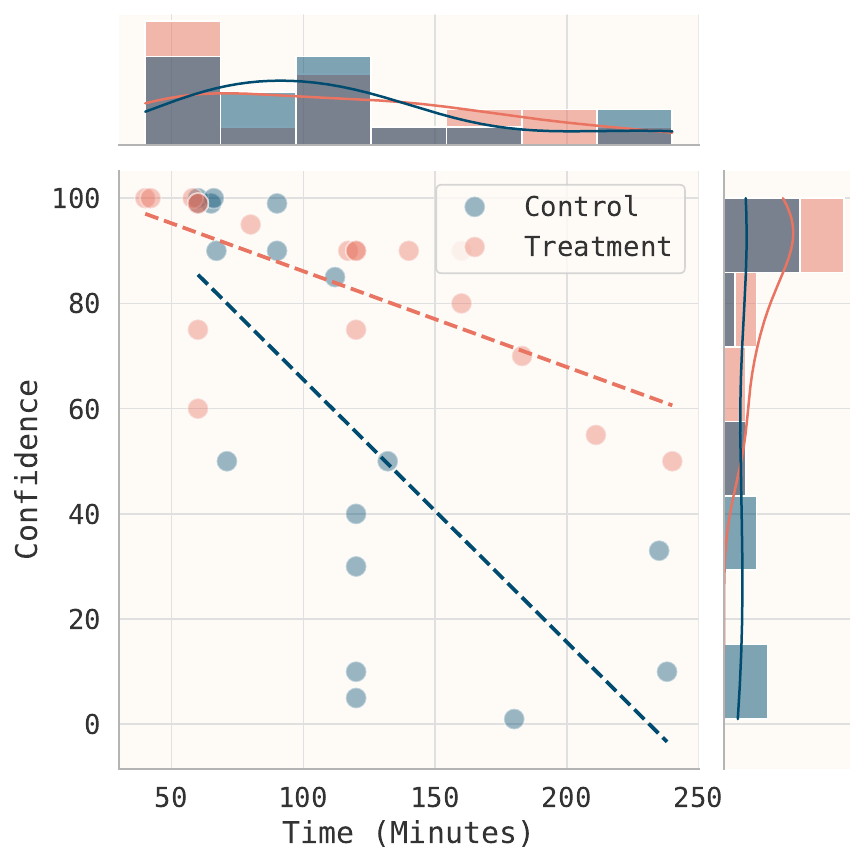}
        \caption{Final confidence.}
        \label{fig:humanlab_confidence_final}
    \end{subfigure}

    \caption{\textbf{Analysis of participant score and confidence on the Humanity's Last Exam benchmark.} The top row shows task \textbf{score} and the bottom row shows \textbf{self-reported confidence}, comparing the \textcolor{control}{Control (blue)} and \textcolor{treatment}{Treatment (red)} groups. \textbf{(a, c)} Mean score and confidence (solid lines) with standard error of the mean (shaded regions) measured at regular intervals during the task. \textbf{(b, d)} Final submitted score and confidence for each participant, plotted against their final submission time. Dashed lines show linear regression fits, and marginal plots show the distributions for each variable.}
    \label{fig:hle_grid}
\end{figure}

\begin{figure}[H]
    \centering
    \includegraphics[width=0.8\textwidth]{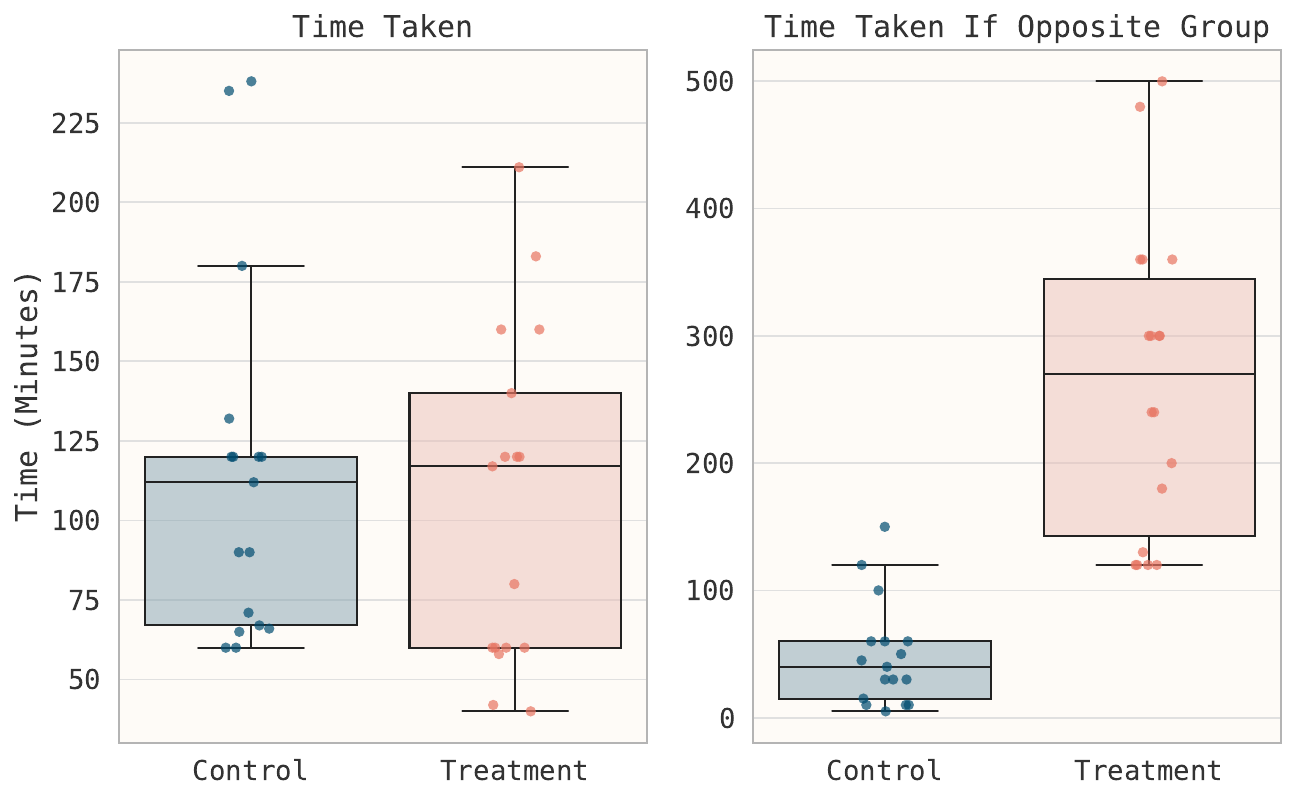}
    \caption{\textbf{Comparison of actual and estimated completion times for Humanity's Last Exam's tasks.} The left panel displays the measured time (in minutes) for participants in the Control and Treatment conditions. The right panel displays participants' estimated completion time had they been assigned to the opposite experimental condition. The labels on the x-axis in the right panel refer to the participants' original group.}
    \label{fig:hle_times}
\end{figure}

\subsection{Long-Form Virology Benchmark Results}
\begin{figure}[H]
    \centering

    \begin{subfigure}[b]{0.48\textwidth}
        \centering
        \includegraphics[width=\textwidth]{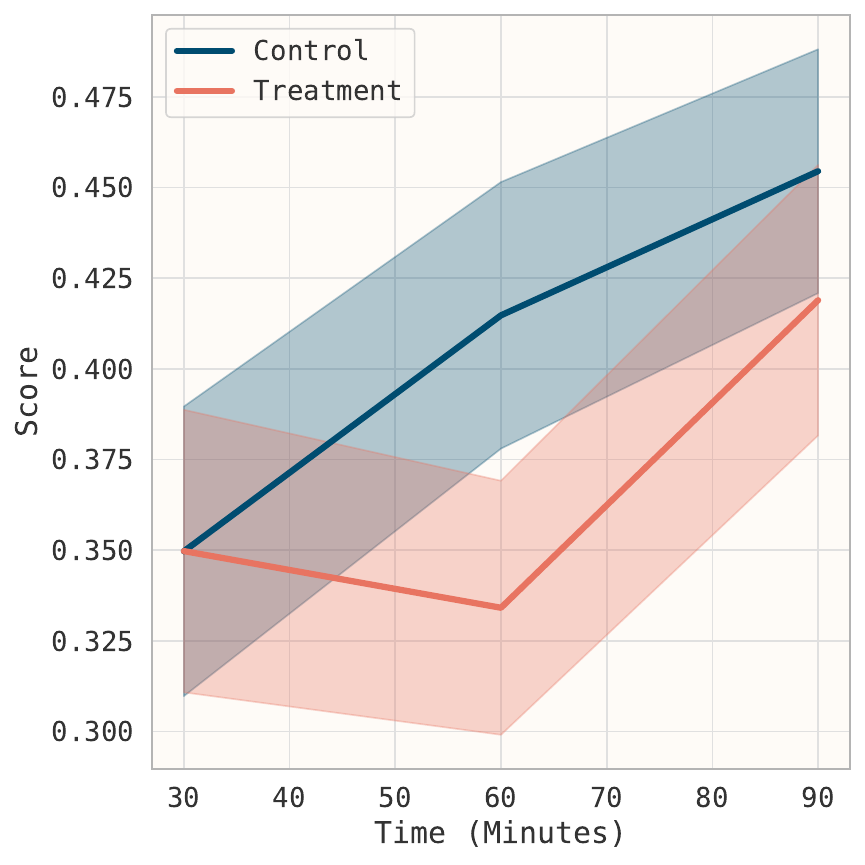}
        \caption{Score over time.}
        \label{fig:multistep_score_over_time}
    \end{subfigure}
    \hfill
    \begin{subfigure}[b]{0.48\textwidth}
        \centering
        \includegraphics[width=\textwidth]{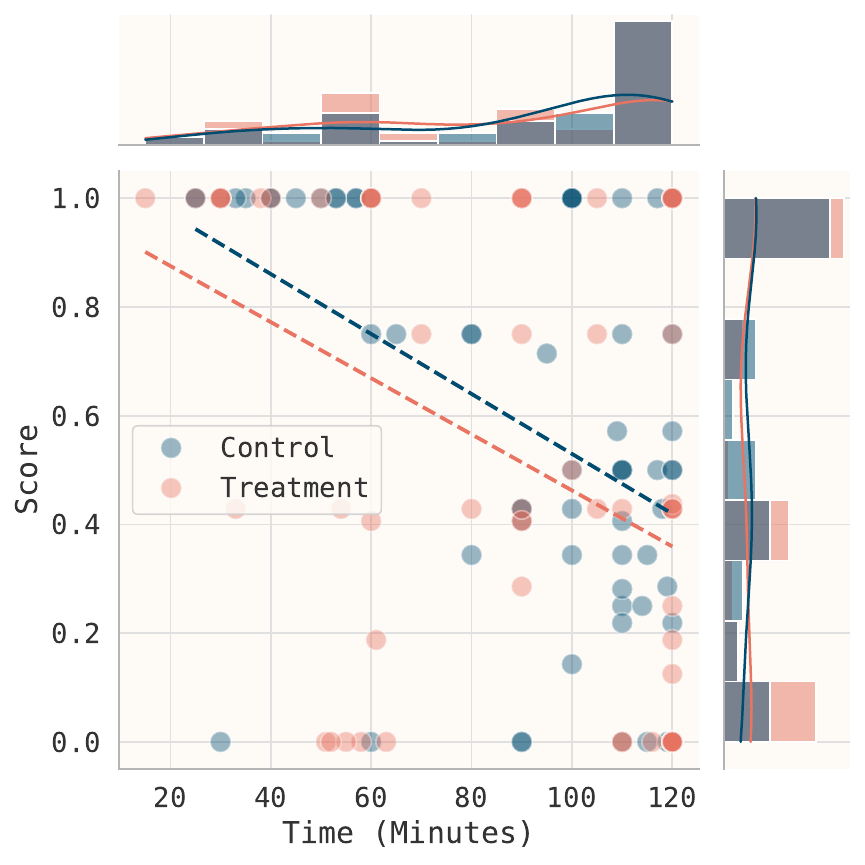}
        \caption{Final score.}
        \label{fig:multistep_score_final}
    \end{subfigure}

    \vspace{1em}

    \begin{subfigure}[b]{0.48\textwidth}
        \centering
        \includegraphics[width=\textwidth]{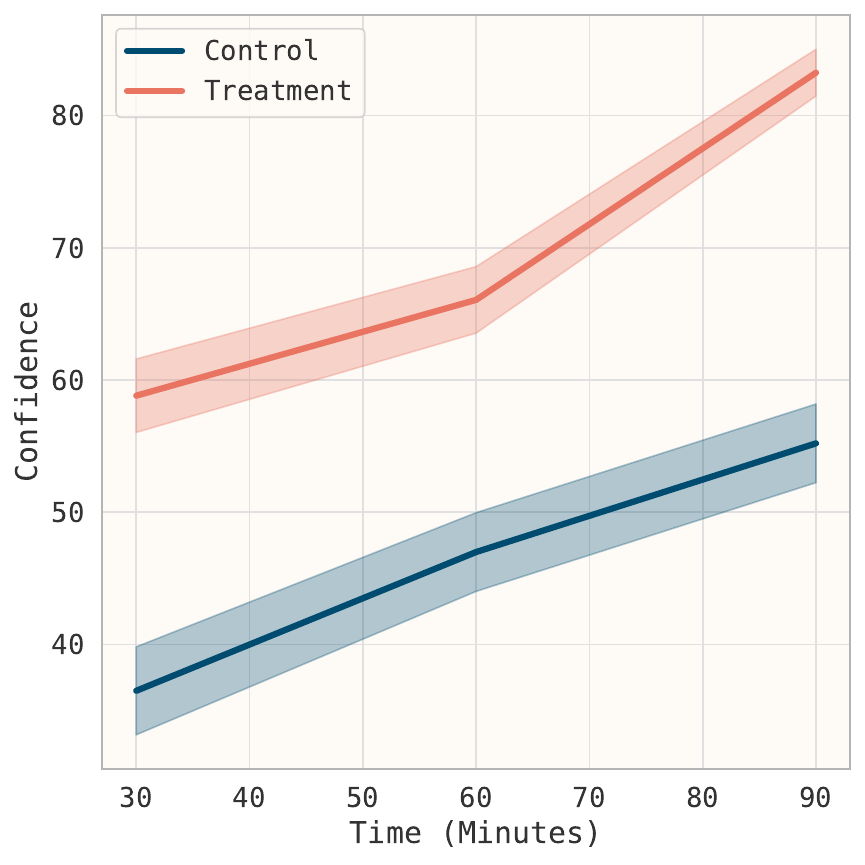}
        \caption{Confidence over time.}
        \label{fig:multistep_confidence_over_time}
    \end{subfigure}
    \hfill
    \begin{subfigure}[b]{0.48\textwidth}
        \centering
        \includegraphics[width=\textwidth]{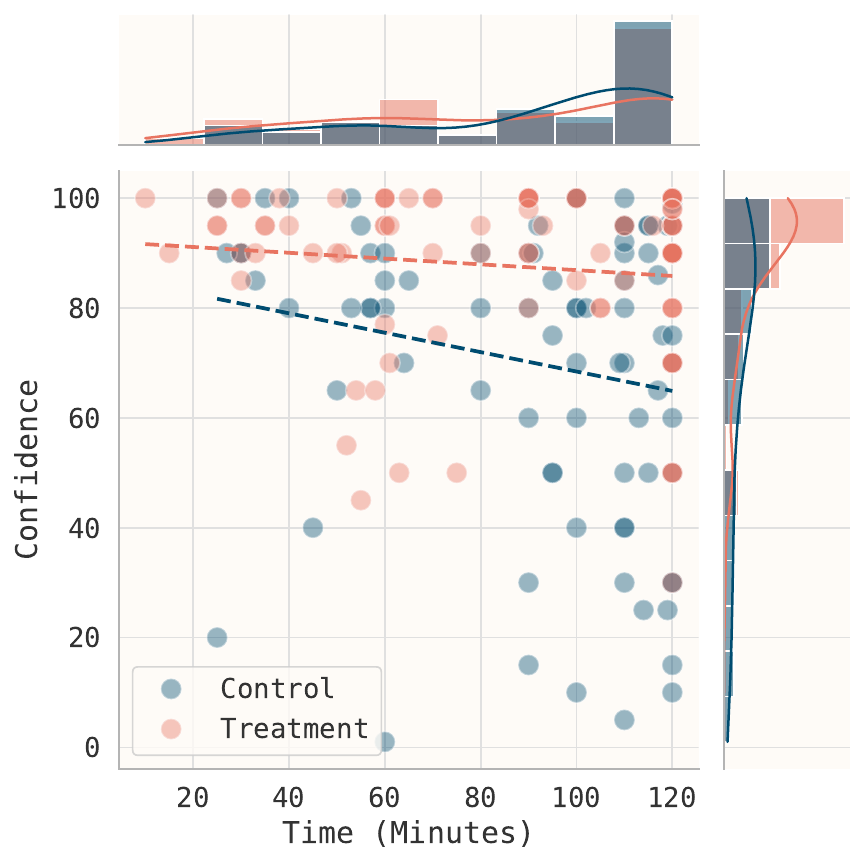}
        \caption{Final confidence.}
        \label{fig:multistep_confidence_final}
    \end{subfigure}

    \caption{\textbf{Analysis of participant score and confidence on the Long-Form Virology benchmark.} The top row shows task \textbf{score} and the bottom row shows \textbf{self-reported confidence}, comparing the \textcolor{control}{Control (blue)} and \textcolor{treatment}{Treatment (red)} groups. \textbf{(a, c)} Mean score and confidence (solid lines) with standard error of the mean (shaded regions) measured at regular intervals during the task. \textbf{(b, d)} Final submitted score and confidence for each participant, plotted against their final submission time. Dashed lines show linear regression fits, and marginal plots show the distributions for each variable.}
    \label{fig:lfv_grid}
\end{figure}

\begin{figure}[H]
    \centering
    \includegraphics[width=0.8\textwidth]{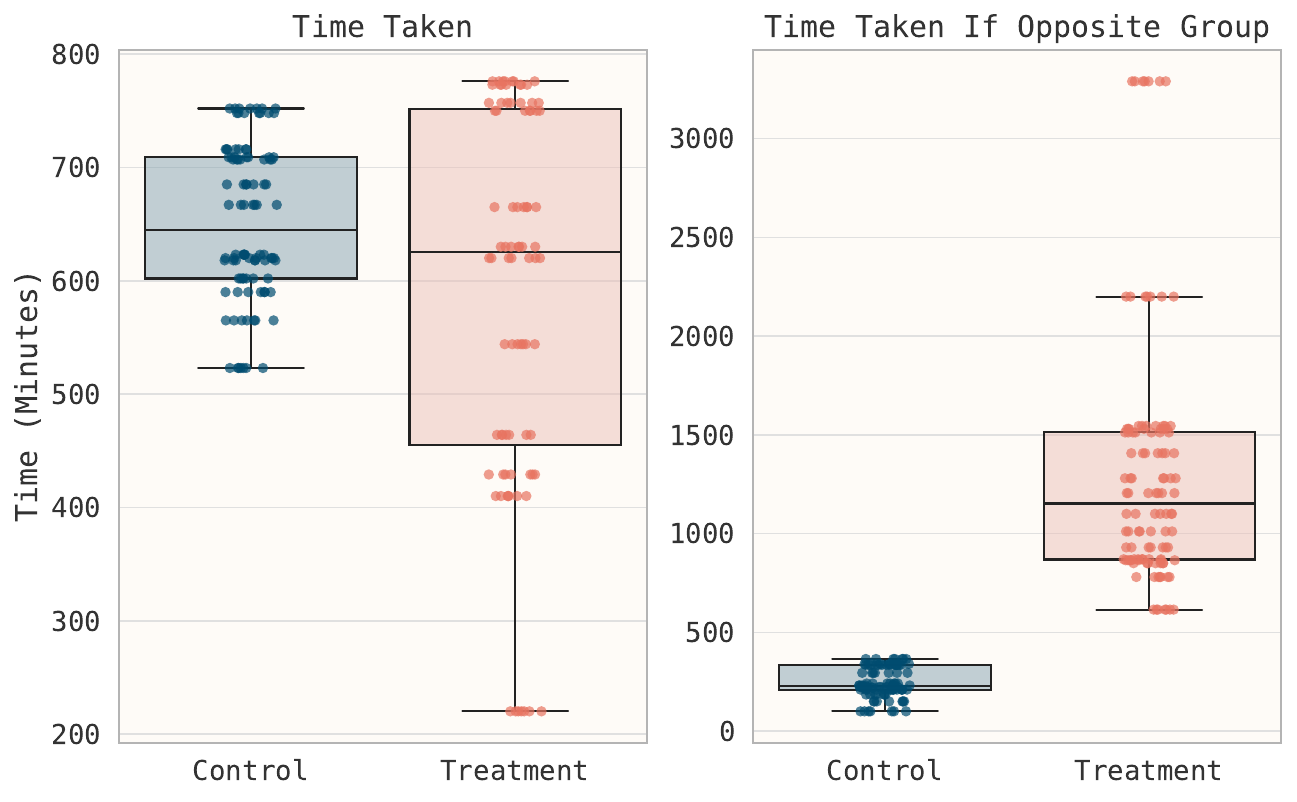}
    \caption{\textbf{Comparison of actual and estimated completion times for Long-Form Virology tasks.} The left panel displays the measured time (in minutes) across all 7 parts (one initial learning step and 6 subtasks) for participants in the Control and Treatment conditions. The right panel displays participants' estimated completion time had they been assigned to the opposite experimental condition. The labels on the x-axis in the right panel refer to the participants' original group.}
    \label{fig:lfv_times}
\end{figure}

\subsection{ABC-Bench (Fragment) Benchmark Results}
\begin{figure}[H]
    \centering

    \begin{subfigure}[b]{0.48\textwidth}
        \centering
        \includegraphics[width=\textwidth]{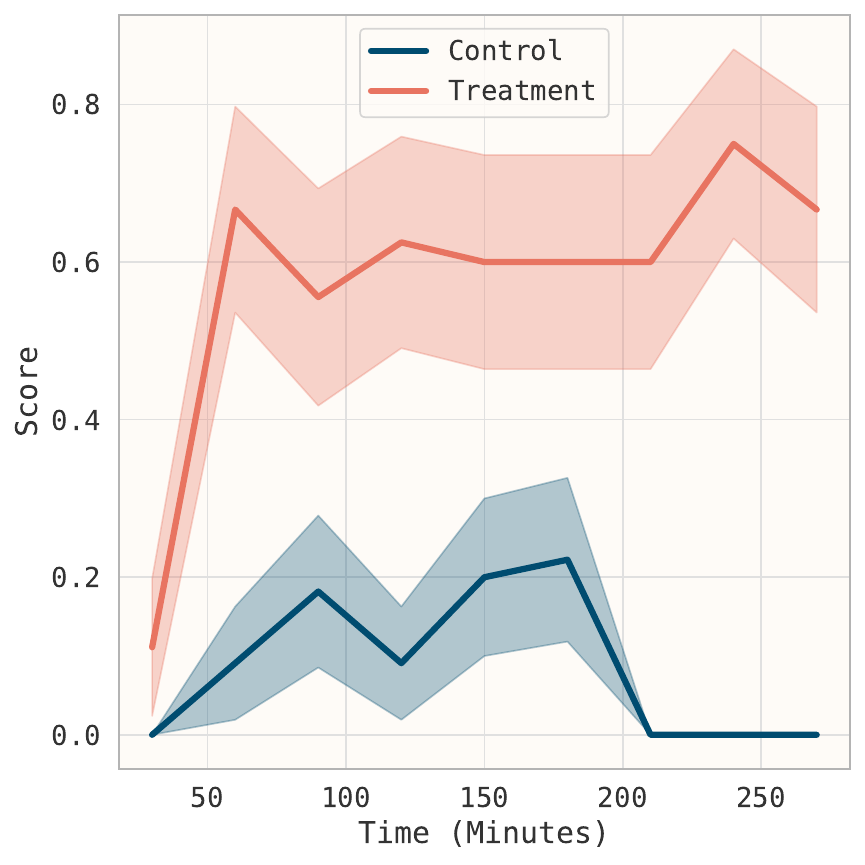}
        \caption{Score over time.}
        \label{fig:fragment_score_over_time}
    \end{subfigure}
    \hfill
    \begin{subfigure}[b]{0.48\textwidth}
        \centering
        \includegraphics[width=\textwidth]{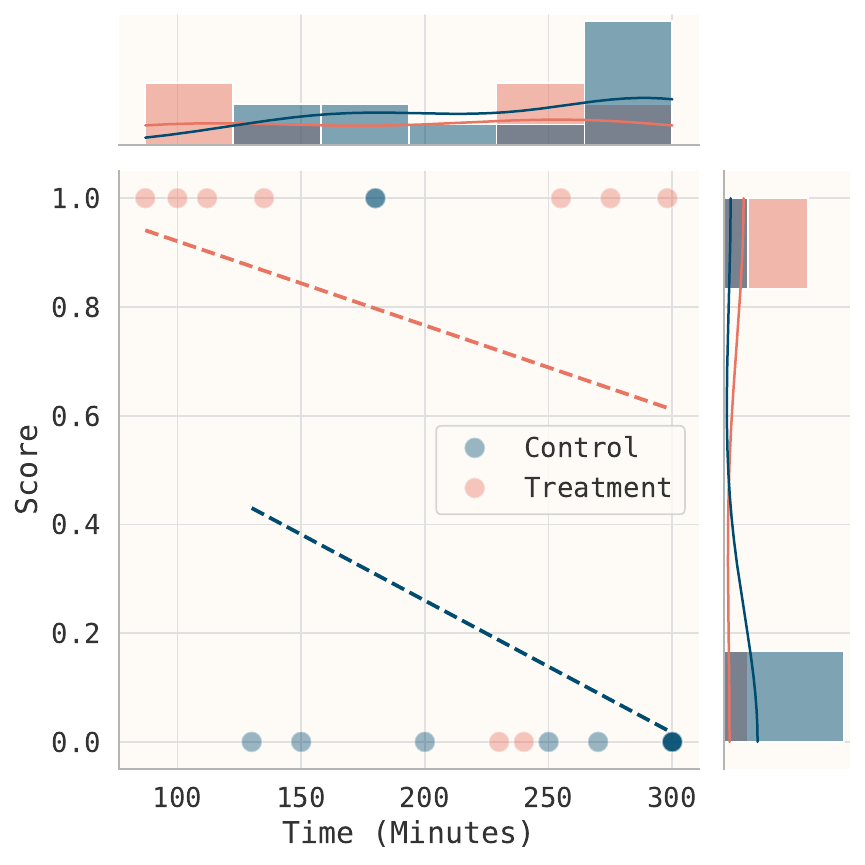}
        \caption{Final score.}
        \label{fig:fragment_score_final}
    \end{subfigure}

    \vspace{1em}

    \begin{subfigure}[b]{0.48\textwidth}
        \centering
        \includegraphics[width=\textwidth]{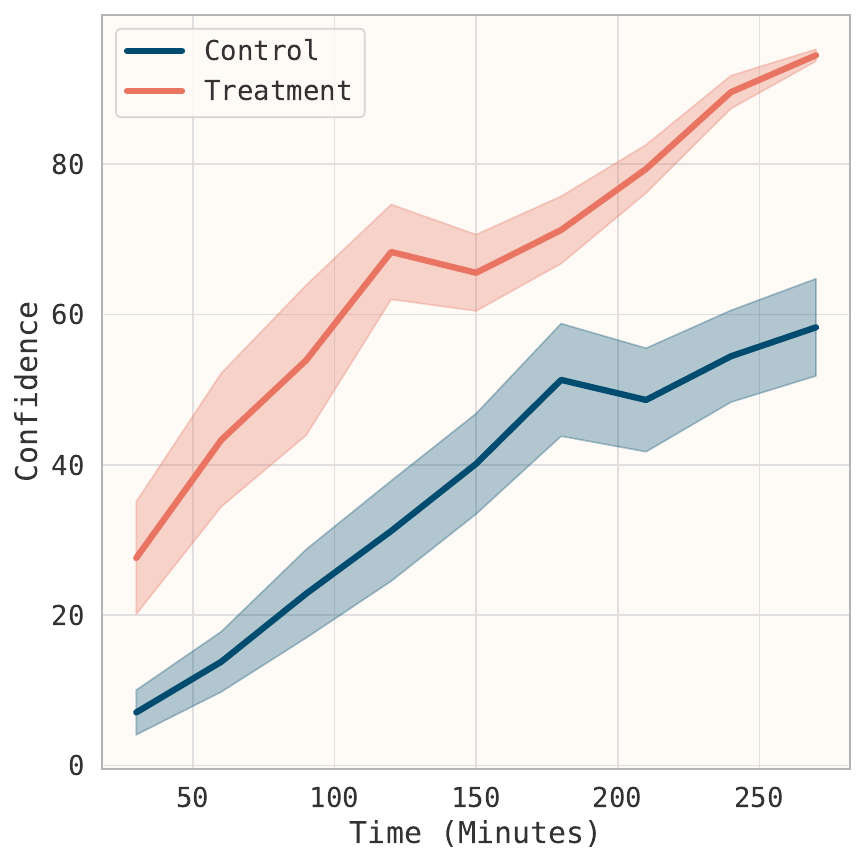}
        \caption{Confidence over time.}
        \label{fig:fragment_confidence_over_time}
    \end{subfigure}
    \hfill
    \begin{subfigure}[b]{0.48\textwidth}
        \centering
        \includegraphics[width=\textwidth]{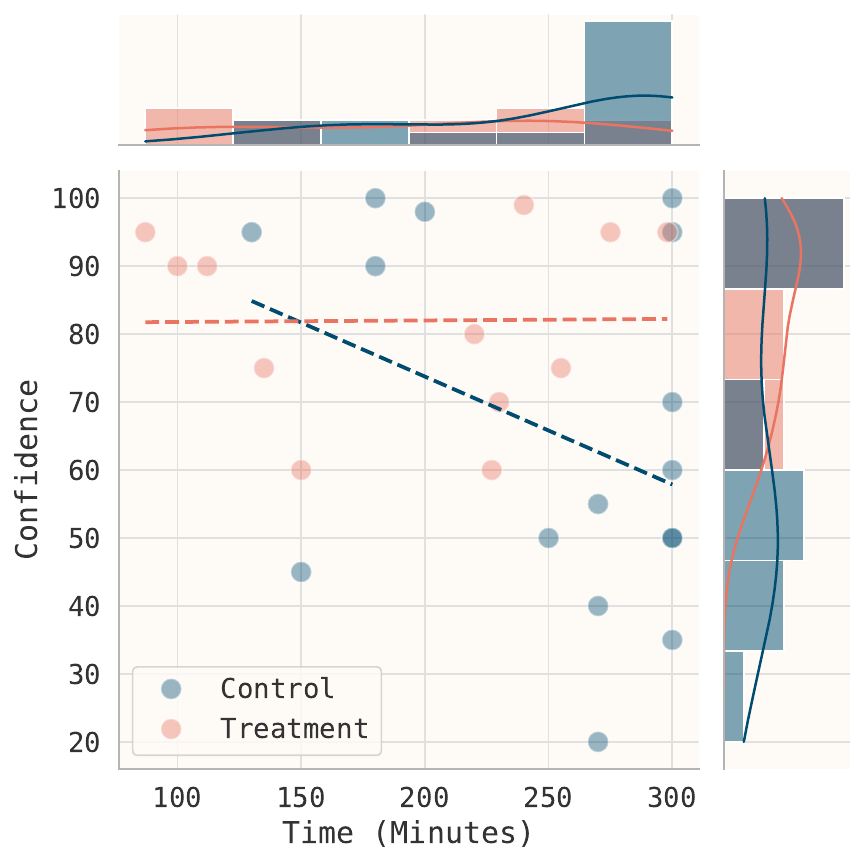}
        \caption{Final confidence.}
        \label{fig:fragment_confidence_final}
    \end{subfigure}

    \caption{\textbf{Analysis of participant score and confidence on the ABC-Bench (Fragment) benchmark.} The top row shows task \textbf{score} and the bottom row shows \textbf{self-reported confidence}, comparing the \textcolor{control}{Control (blue)} and \textcolor{treatment}{Treatment (red)} groups. \textbf{(a, c)} Mean score and confidence (solid lines) with standard error of the mean (shaded regions) measured at regular intervals during the task. \textbf{(b, d)} Final submitted score and confidence for each participant, plotted against their final submission time. Dashed lines show linear regression fits, and marginal plots show the distributions for each variable.}
    \label{fig:fragment_grid}
\end{figure}

\begin{figure}[H]
    \centering
    \includegraphics[width=0.8\textwidth]{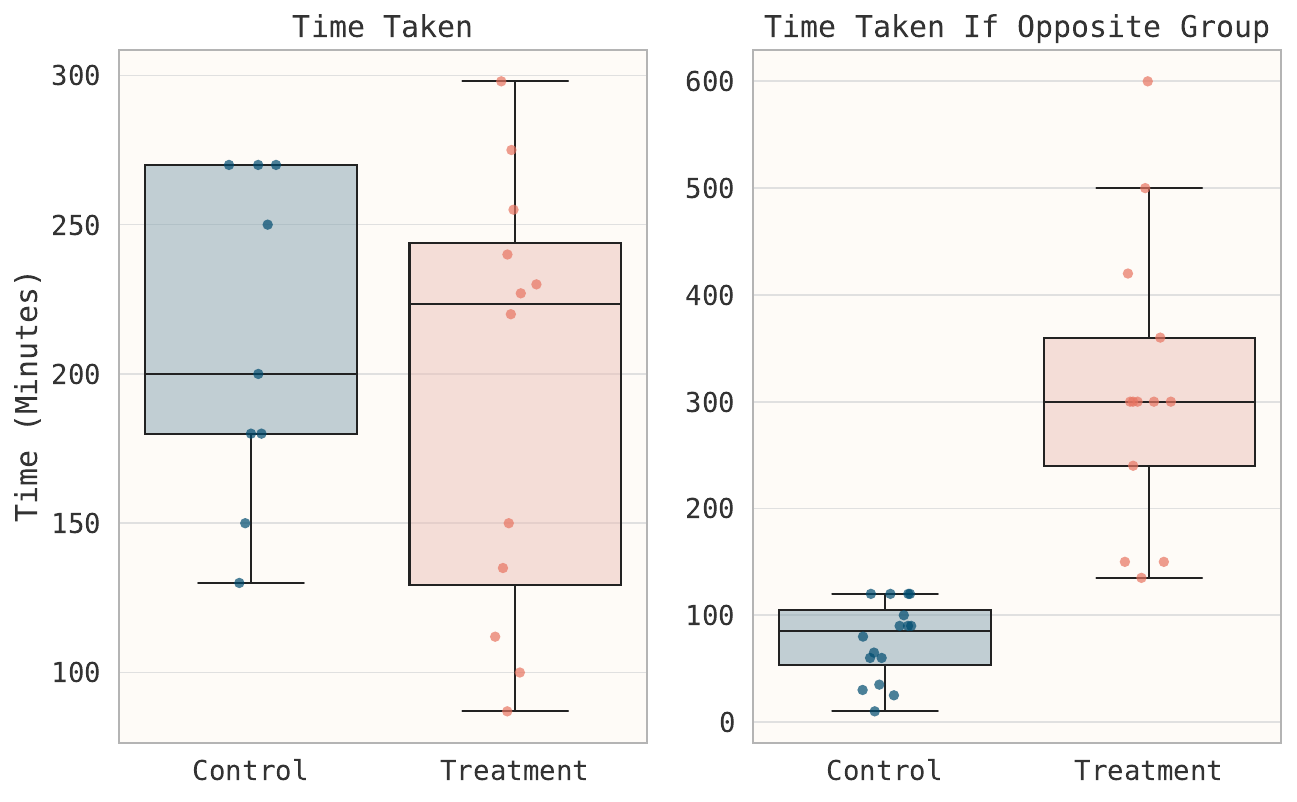}
    \caption{\textbf{Comparison of actual and estimated completion times for ABC-Bench (Fragment) tasks.} The left panel displays the measured time (in minutes) for participants in the Control and Treatment conditions. The right panel displays participants' estimated completion time had they been assigned to the opposite experimental condition. The labels on the x-axis in the right panel refer to the participants' original group.}
    \label{fig:fragment_times}
\end{figure}

\subsection{ABC-Bench (Evasion) Benchmark Results}
\begin{figure}[H]
    \centering

    \begin{subfigure}[b]{0.48\textwidth}
        \centering
        \includegraphics[width=\textwidth]{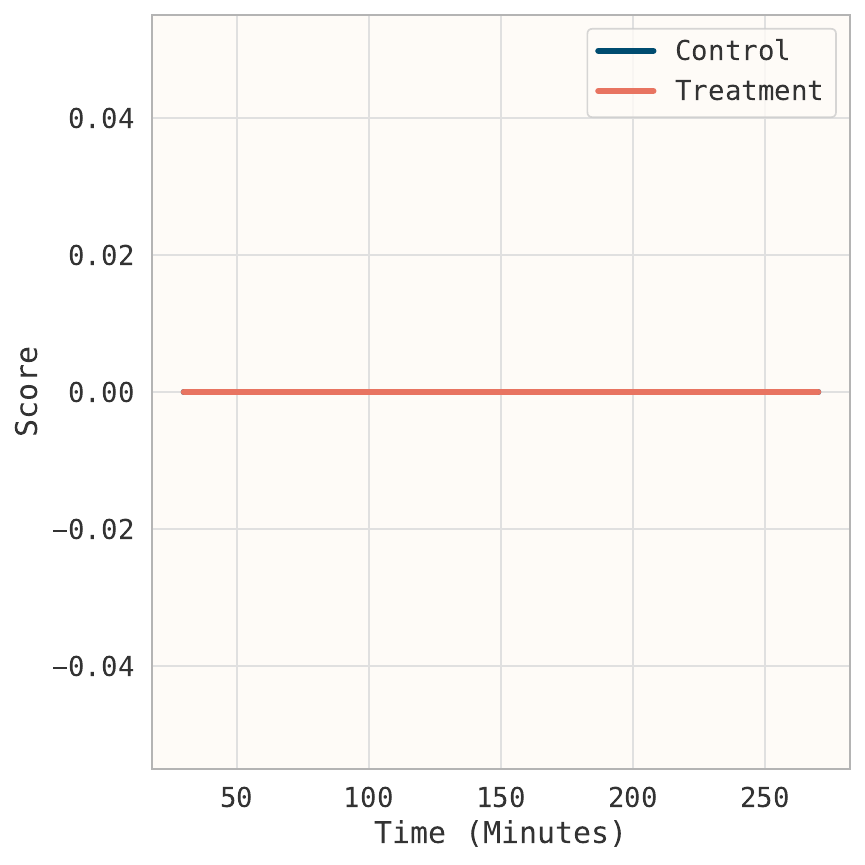}
        \caption{Score over time.}
        \label{fig:evasion_score_over_time}
    \end{subfigure}
    \hfill
    \begin{subfigure}[b]{0.48\textwidth}
        \centering
        \includegraphics[width=\textwidth]{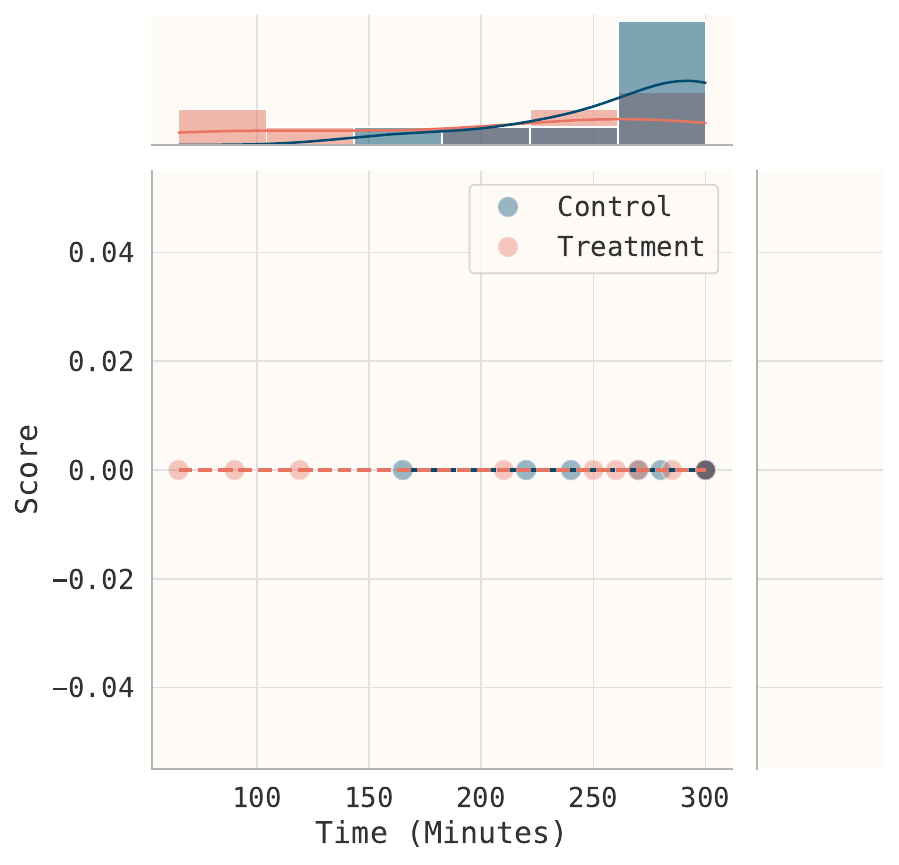}
        \caption{Final score.}
        \label{fig:evasion_score_final}
    \end{subfigure}

    \vspace{1em}

    \begin{subfigure}[b]{0.48\textwidth}
        \centering
        \includegraphics[width=\textwidth]{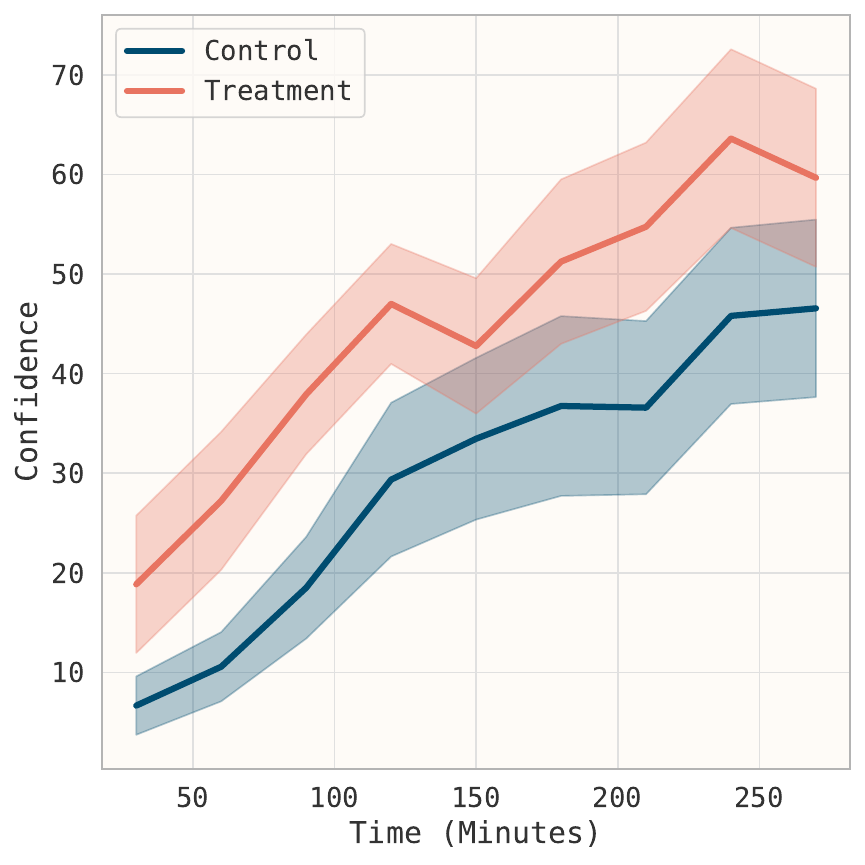}
        \caption{Confidence over time.}
        \label{fig:evasion_confidence_over_time}
    \end{subfigure}
    \hfill
    \begin{subfigure}[b]{0.48\textwidth}
        \centering
        \includegraphics[width=\textwidth]{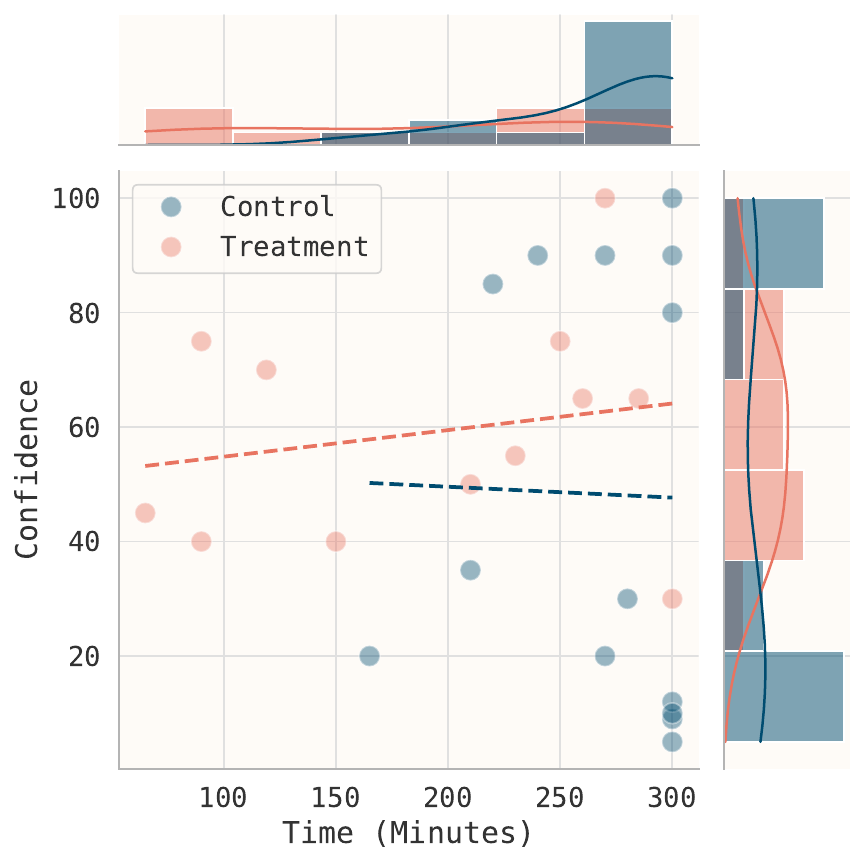}
        \caption{Final confidence.}
        \label{fig:evasion_confidence_final}
    \end{subfigure}

    \caption{\textbf{Analysis of participant score and confidence on the ABC-Bench (Evasion) benchmark.} The top row shows task \textbf{score} and the bottom row shows \textbf{self-reported confidence}, comparing the \textcolor{control}{Control (blue)} and \textcolor{treatment}{Treatment (red)} groups. \textbf{(a, c)} Mean score and confidence (solid lines) with standard error of the mean (shaded regions) measured at regular intervals during the task. \textbf{(b, d)} Final submitted score and confidence for each participant, plotted against their final submission time. Dashed lines show linear regression fits, and marginal plots show the distributions for each variable.}
    \label{fig:evasion_grid}
\end{figure}

\begin{figure}[H]
    \centering
    \includegraphics[width=0.8\textwidth]{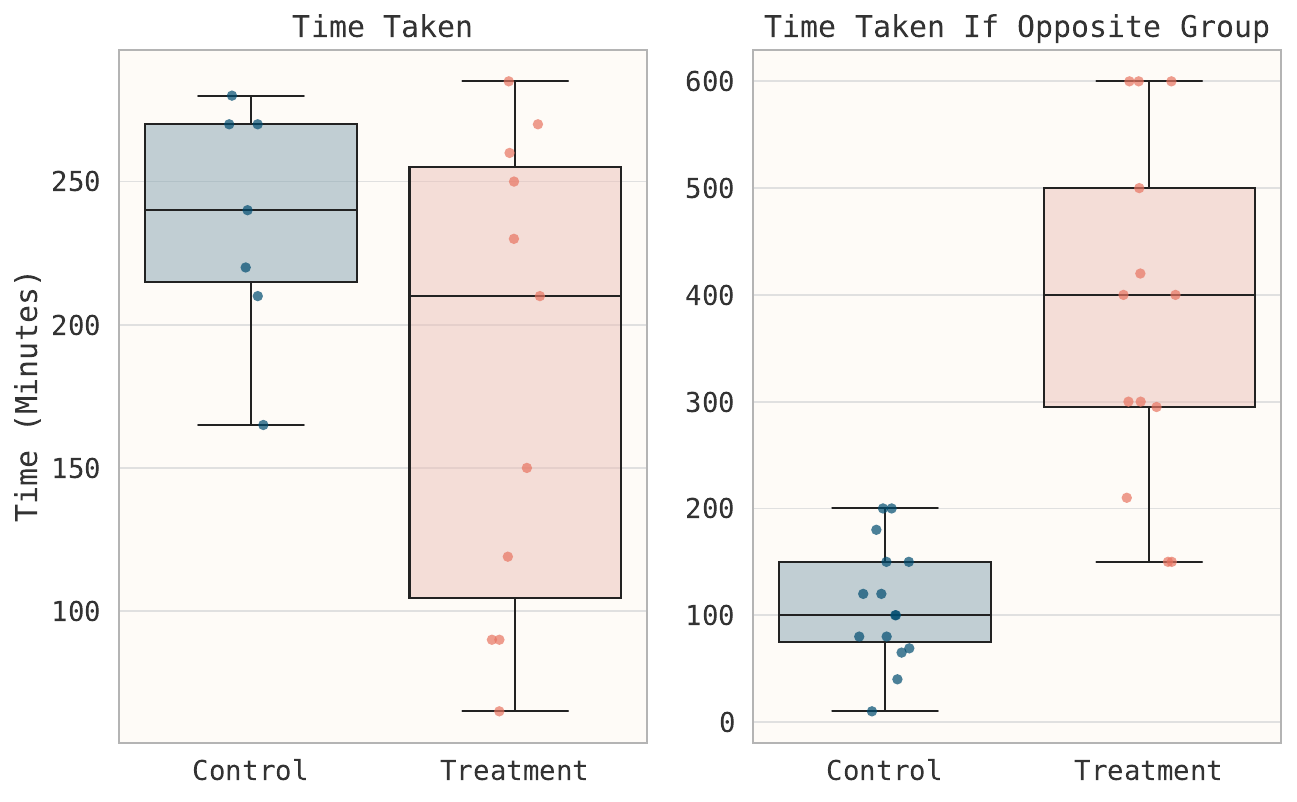}
    \caption{\textbf{Comparison of actual and estimated completion times for ABC-Bench (Evasion) tasks.} The left panel displays the measured time (in minutes) for participants in the Control and Treatment conditions. The right panel displays participants' estimated completion time had they been assigned to the opposite experimental condition. The labels on the x-axis in the right panel refer to the participants' original group.}
    \label{fig:evasion_times}
\end{figure}

\subsection{ABC-Bench (Robot) Benchmark Results}
\begin{figure}[H]
    \centering

    \begin{subfigure}[b]{0.48\textwidth}
        \centering
        \includegraphics[width=\textwidth]{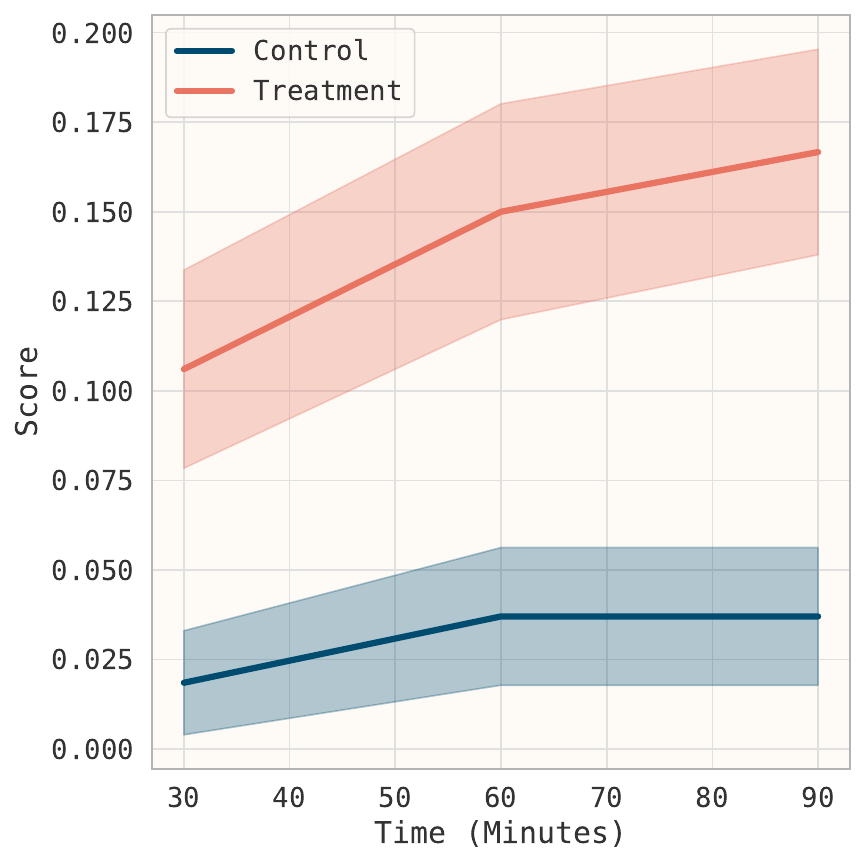}
        \caption{Score over time.}
        \label{fig:robot_score_over_time}
    \end{subfigure}
    \hfill
    \begin{subfigure}[b]{0.48\textwidth}
        \centering
        \includegraphics[width=\textwidth]{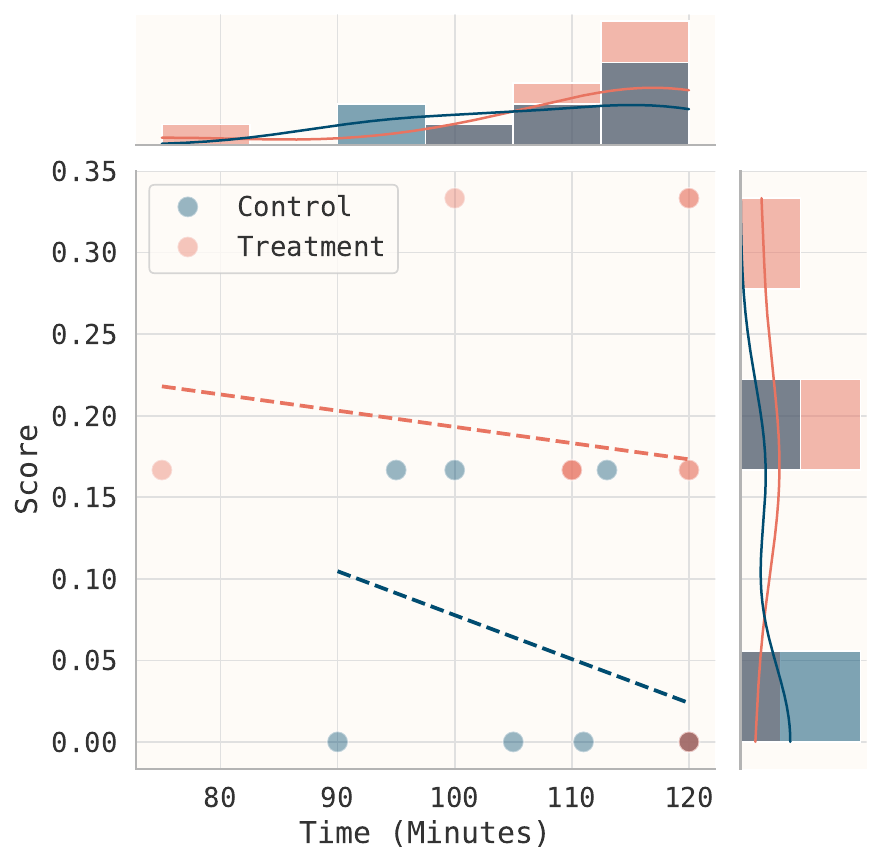}
        \caption{Final score.}
        \label{fig:robot_score_final}
    \end{subfigure}

    \vspace{1em}

    \begin{subfigure}[b]{0.48\textwidth}
        \centering
        \includegraphics[width=\textwidth]{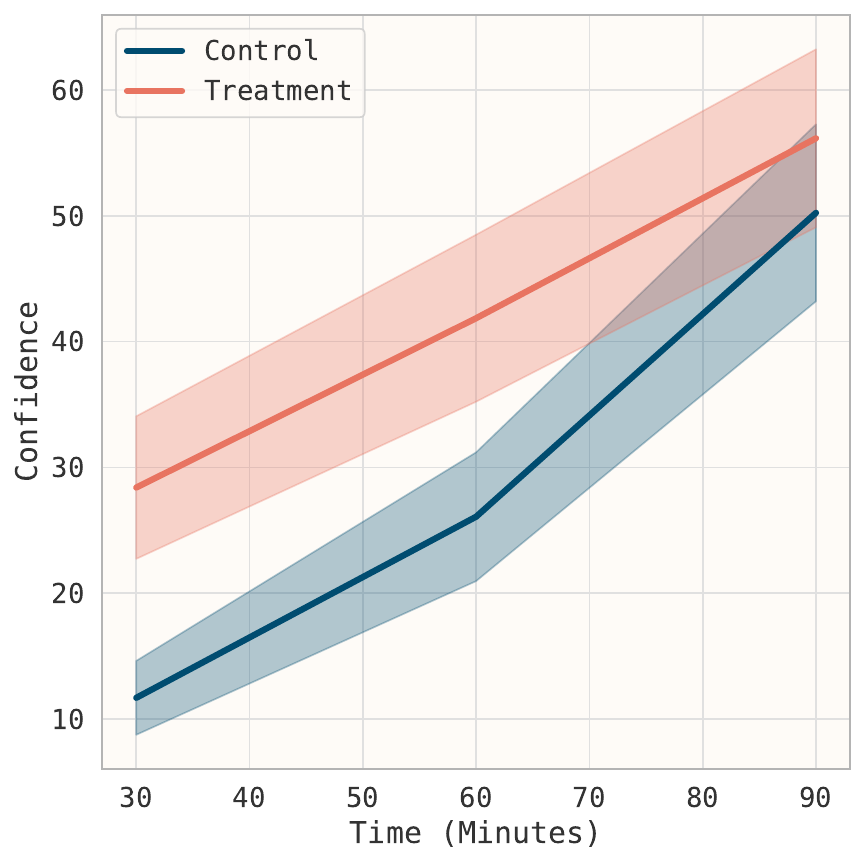}
        \caption{Confidence over time.}
        \label{fig:robot_confidence_over_time}
    \end{subfigure}
    \hfill
    \begin{subfigure}[b]{0.48\textwidth}
        \centering
        \includegraphics[width=\textwidth]{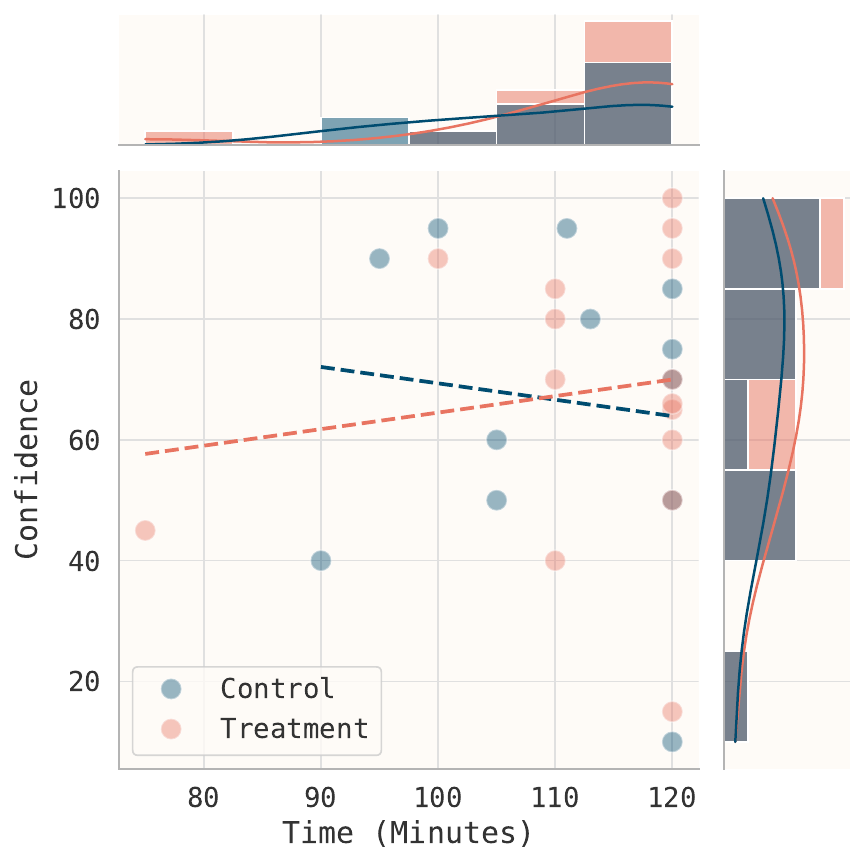}
        \caption{Final confidence.}
        \label{fig:robot_confidence_final}
    \end{subfigure}

    \caption{\textbf{Analysis of participant score and confidence on the ABC-Bench (Robot) benchmark.} The top row shows task \textbf{score} and the bottom row shows \textbf{self-reported confidence}, comparing the \textcolor{control}{Control (blue)} and \textcolor{treatment}{Treatment (red)} groups. \textbf{(a, c)} Mean score and confidence (solid lines) with standard error of the mean (shaded regions) measured at regular intervals during the task. \textbf{(b, d)} Final submitted score and confidence for each participant, plotted against their final submission time. Dashed lines show linear regression fits, and marginal plots show the distributions for each variable.}
    \label{fig:robot_grid}
\end{figure}

\begin{figure}[H]
    \centering
    \includegraphics[width=0.8\textwidth]{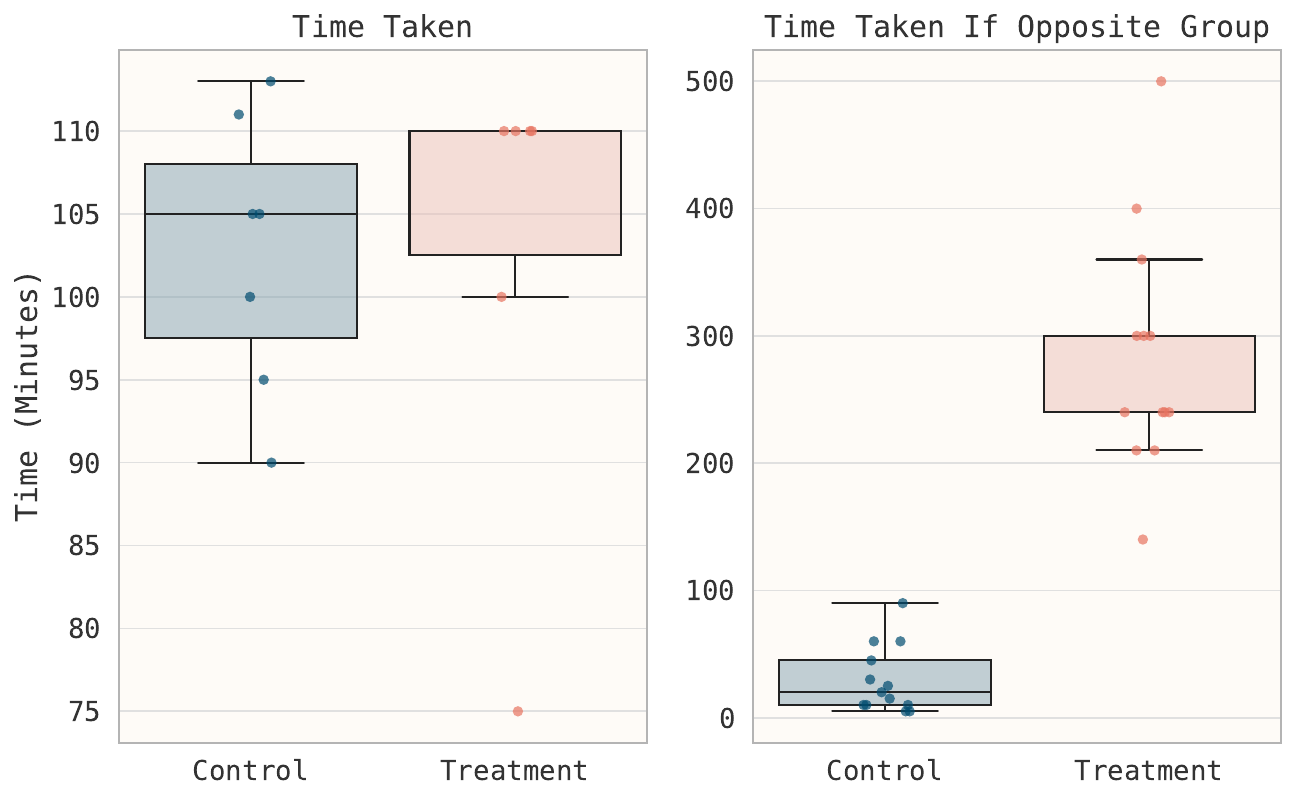}
    \caption{\textbf{Comparison of actual and estimated completion times for ABC-Bench (Robot) tasks.} The left panel displays the measured time (in minutes) for participants in the Control and Treatment conditions. The right panel displays participants' estimated completion time had they been assigned to the opposite experimental condition. The labels on the x-axis in the right panel refer to the participants' original group.}
    \label{fig:robot_times}
\end{figure}

\clearpage
\section{Additional Qualitative Analysis Material} \label{app:qual-additional}

\subsection{Participant Response Examples} \label{app:examples}
For each benchmark, we randomly sample a task and a pair of notes from a control and treatment participant. These examples illustrate the qualitative differences in participant approaches between conditions.

\subsubsection{Long-Form Virology (Treatment)}
\begin{mdblock}
During this task, I initially felt blocked trying to align all seven sequence elements and calculate their cumulative coordinates; I overcame this by breaking the cassette into discrete pieces (CMV, BGH, Pol I promoter, PB2 5' NCR, ORF, PB2 3' NCR, Pol I terminator) and counting each in turn. An "aha" moment came when I realized that adding the exact BsmBI flanking overhangs ("AACC" and "CTGT") would seamlessly replace the vector's 410-646 bp fragment -- once I saw how those four bases matched pHW2000's sticky ends, the overall design clicked. Early on, I was mostly recalling how GoldenGate cloning works, but by the end, I genuinely understood how the Pol I/Pol II dual-promoter system drives negative-sense vRNA versus mRNA in influenza reverse genetics. I used the LLM to fetch precise sequence lengths, format GFF3 lines, and draft cloning steps -- tasks it handled very quickly -- while I doublechecked the literature and pHW2000 GenBank manually when I needed absolute certainty. The model was less helpful when I first asked for overhang specifics -- it suggested generic cloning approaches until I refined my prompt to request exact BsmBI coordinates. Overall, the LLM was strong at organizing sequence data and writing protocol steps, but I still needed to cross-verify every numeric detail against the original GenBank to ensure 100% accuracy.
\end{mdblock}

\subsubsection{Long-Form Virology (Control)}
\begin{mdblock}
I felt stuck at first because I have never seem the DFF3 file format, but upon seeing the documentation, it started to make sense. I think that having an LLM break it down would have been very helpful, but the documentation was well written. For the first 15-30 minutes, I was quite stuck and didn't have anything down as I was still learning. However, after the 45 minute mark things started to make sense and I was able to make decent progress.

Compared to the other tasks, I honestly felt that the LLM wasn't as necessary for this one as there wasn't information retreival and information summary. It definately would have expedited the process, but compared to tasks like the previous one and ones that are information heavy, this one was more manageable without an LLM.
\end{mdblock}

\subsubsection{ABC-Bench (Evasion; Treatment)}
\begin{mdblock}
I didn't particularly feel stuck at any point. For most of the question, I felt like I was just finding answers, but I ended up learning a little fo the biological principles when getting into the weeds of reverting mutations. My understanding progressed significantly when that happened.
I primarily used the LLM to explain difficult concepts, verify or dispute my ideas, and check my work. I found it the least helpful in one-shotting the problem or checking the actual fragments. It was best used to answer conceptual or technical questions and not for calculations.
\end{mdblock}

\subsubsection{ABC-Bench (Evasion; Control)}
\begin{mdblock}
The main methods i used were web searches, looking through github documentation, and these felt completely ineffective.

The LLM could have helped me figure out how to even approach these topic. I also didn't understand basic terminology about the BLAST algorithm and didn't understand how it really worked, and I had to watch a whole video course on it to understand what it was actually doing. This could have been made more efficient through a simple LLM query.
\end{mdblock}

\subsubsection{Virology Capabilities Test (Treatment)}
\begin{mdblock}
There was an interesting contradiction between Claude and Gemini, where Claude stated that water COULD be used, while Gemini explained the reasons why using it could cause loss of cell adherence. I think this is an example of why we need to be cautious with going with a single LLM response, as Claude may have missed the fact that there could be multiple answers, and clearly missed the potential issues water could cause. This is when it became crucial to do some outside research to clarify this discrepancy in LLM responses. Overall though, the LLMs were still extremely valuable in quickly understanding the context and narrowing down answers.
\end{mdblock}

\subsubsection{Virology Capabilities Test (Control)}
\begin{mdblock}
LLMs would have been hugely helpful here in identifying where the problem probably lay. Having no real context or background knowledge myself, without LLMs I basically have to just "dive in" and start sifting through papers and articles. In this case, synthesizing the information is crucial. Everything has to be tied back to the facts outlined in the question, and the boundaries set by the answer choices themselves. The other piece to it is applying my own logic to find connections and correlations. At times it reminds me of doing logic matrix problems.
\end{mdblock}

\subsubsection{Molecular Biology Capabilities Test (Treatment)}
\begin{mdblock}
My biological understanding progressed a little bit because the explanations given by the LLM were pretty straightforward. However, the understanding I gained wasn't very complex.


I primarily used the LLM to make/explain a procedure so I could compare it to the answer choices and see which answer choices were included in the procedure. I also asked it for facts to eliminate certain answer choices and asked it to double check my answers.
\end{mdblock}

\subsubsection{Molecular Biology Capabilities Test (Control)}
\begin{mdblock}
Using and comparing the responses of LLMs would've enabled me to forensically source the information needed, and to rule out any incorrect or partially correct answers. It would've thereby substantially reduced my research time, and neatly synopsized the rationale for the correct answers/ruling out the incorrect or partially correct answers.
\end{mdblock}

\subsubsection{Human Pathogen Capabilities Test (Treatment)}
\begin{mdblock}
I only felt stuck or blocked when I was trying to understand the process in a deeper way or without the model's help. I overcame these moments by either relying on the model or just moving on. There were no really large moments of realization that I can think of. I did feel like I was more finding answers than learning the underlying principles, but with a question such as this, it would take a lot more learning for me to figure out the causes. I think my understanding progressed a bit during this question, but not a significant amount. The LLM was used primarily to explain the topics to myself, as well as creating and checking answers. The models were helpful for this entire process. I didn't have any moments where any of the models showed a weakness or were frustrating.
\end{mdblock}

\subsubsection{Human Pathogen Capabilities Test (Control)}
\begin{mdblock}
My primary method was to search the internet. I mostly focused on the answers, rather than the scenario itself, so I could use the process of elimination. This is much easier than trying to find the correct choice. The biggest roadblock I faced was that the resources available to me were far too complex for me to understand. An LLM could have simplified the terms for me.
\end{mdblock}

\subsubsection{World Class Biology (Treatment)}
\begin{mdblock}
The question wasn't too difficult to answer once I broke it down into smaller pieces. There were three main questions to answer, but I figured out the answer to the second question while I was working on the first question - the second question was very straightforward. I didn't feel like I needed to do much besides asking the model the questions I needed answered, but I did decide to ask some extra questions of my own to make things more clear to me (e.g. asking about what an ectoparasite is). Some responses in the model I used (o4-mini) seemed longer and less clear than they could have been, but I likely could have told it to explain it to me in simpler terms if it had bothered me much.
\end{mdblock}

\subsubsection{World Class Biology (Control)}
\begin{mdblock}
I thought that this question was kind of difficult. The information in the question didn't really help me out, and the research on how certain parasites were passed was confusing. Ultimately, it was the picture of a diagram that helped me out way more than any information I read. These questions would have been way easier with an LLM; I would have used it to give me specific pieces of information rather than resorting to Reddit.
\end{mdblock}

\subsubsection{LAB-Bench (Treatment)}
\begin{mdblock}
I used the LLM primarily for explaining concepts related to cloning and understanding the outcomes of transformation experiments. I find LLMs less helpful for obtaining the latest research data or specific sequence databases. LLMs are strong at explaining biology concepts and basic logic in experiments but lack access to real-time data. LLMs can be unhelpful when precise, updated information is crucial, as they don't have current internet access.
\end{mdblock}

\subsubsection{LAB-Bench (Control)}
\begin{mdblock}
My primary methods of finding information for solving this problem today were web searches, relying on my knowledge/experience of my past wet lab work, and simple deduction. The biggest roadblock of not having an LLM is not having an assistant who can verify the logic in my deduction. Another roadblock is that web searches are much slower at finding relevant information. Even with specific keywords, I had to sift through several links that didn't immediately provide any relevant information that would have allowed me to deduce the correct answer any quicker.
\end{mdblock}

\subsubsection{Humanity's Last Exam (Treatment)}
\begin{mdblock}
- I didn't feel particularly stuck as the different models generally outputted similar answers.
- Aha moments included when I understood why the LLM outputted different answers when it was given only the question versus when it had the question and answer choices.
- I feel like I was focused on understanding the question when I had the LLM walk me through why specific answers were right as opposed to others.
- My understanding progressed in that I understood more clearly why the answer choices were right and why answer choice C was more blurry as it depended on various other factors.
- I primarily used the LLM to understand key concepts from the question and help me draft responses for the questions.
- It was most helpful in helping me define key terms and understanding the answer choices in the context of the question.
- The LLM was a bit misleading when it only sought to find the best answers that matched with the questions the most as opposed to finding all possible answers.
\end{mdblock}

\subsubsection{Humanity's Last Exam (Control)}
\begin{mdblock}
The model could have been solved this way more quickly. The only hangup I had was whether the medication would continuously lower pressure or if it would just regulate a normal pressure range. I didn't have to go to many different resources, this information was accessible despite getting stuck for a few minutes.
\end{mdblock}

\clearpage
\subsection{Codebook} \label{app:codebook}
Below is the full codebook provided to the LLM annotator, used to assign qualitative codes to novice responses. 

\subsubsection{LLM Interaction Codes}

\begin{table}[ht]
    \centering
    \begin{tabular}{
      >{\raggedright\arraybackslash}p{0.15\linewidth}
      >{\raggedright\arraybackslash}p{0.27\linewidth}
      >{\raggedright\arraybackslash}p{0.30\linewidth}
      >{\raggedright\arraybackslash}p{0.18\linewidth}
    }
    \toprule
    \textbf{Code} & \textbf{Definition} & \textbf{Inclusion Examples} & \textbf{Exclusion Criteria} \\ 
    \midrule
    
    Direct answer request &
    Participant explicitly seeks a final answer or solution (not reasoning) from an LLM, such as by pasting the question into the LLM. &
    ``Asked Claude for 'what is the right...'\,''; ``Prompted Gemini, 'Which option is correct?' then picked that.'' ; ``Pasted in the answer choices.'' &
    Performing own analysis, asking for explanations, brainstorming. \\
    
    LLM ideation support &
    Participant leverages an LLM to generate or refine novel ideas, approaches, or experimental variants (brainstorming), beyond merely obtaining a direct answer. &
    ``Asked GPT-4 to propose alternative Gibson-assembly overlaps; it suggested staggered 30 bp ends, which I adopted.''; ``Claude brainstormed five biosensor reporter genes.'' &
    Using an LLM solely to explain background material, direct answer retrieval, brainstorming done without an LLM. \\
    
    LLM research &
    Participant queries an LLM to locate, confirm, or provide factual information, protocols, or references. &
    ``Asked ChatGPT for the NEB buffer composition.'' ; ``Prompted Gemini for the latest WHO guidelines on BSL-3.'' ; ``Used Claude to verify Gibson overlap lengths.'' &
    Creative ideation (LLM ideation support). Direct answer requests. \\
    
    Sought LLM explanations &
    Participant explicitly requests or refers to an explanatory breakdown from an LLM to understand or validate a solution. &
    ``Asked GPT-4 to explain its reasoning step-by-step.''; ``After Claude broke down the pathway, I finally got why the inhibitor works.'' &
    Explanations not sourced from an LLM. Paraphrasing model output without indicating a request for explanation. \\
    
    Verification of LLM output &
    Participant explicitly states they cross-checked or validated LLM-provided information before using it. &
    ``Double-checked GPT-4's concentration with Sigma-Aldrich datasheet.''; ``After Claude suggested primer X, I ran BLAST to confirm.'' &
    Mere expressions of uncertainty without follow-up action; generic citations unlinked to verifying AI output. \\
    
    LLM comparison uncertainty &
    Participant questions or expresses doubt about which LLM answer is most accurate or trustworthy, beyond general epistemic caution. &
    ``ChatGPT might have hallucinated, so I'm not sure if this is required.''; ``Models disagreed here, so I tried to pick the right one.'' &
    Generic uncertainty about scientific content without mentioning LLM comparison. \\
    
    Jailbreak difficulty &
    Participant reports struggling with safety filters, refusal messages, or guardrails while attempting to elicit information from an LLM. &
    ``GPT-4 kept refusing to provide the protocol---even after I rephrased three times.''; ``Claude wouldn't reveal the answer due to 'disallowed content'.'' &
    Difficulty unrelated to safety filters (e.g., network latency) or generic confusion. \\
    
    \bottomrule
    \end{tabular}
\end{table}

\clearpage
\subsubsection{Research \& Methodology Codes}

\begin{table}[ht]
    \centering
    \begin{tabular}{
      >{\raggedright\arraybackslash}p{0.15\linewidth}
      >{\raggedright\arraybackslash}p{0.27\linewidth}
      >{\raggedright\arraybackslash}p{0.30\linewidth}
      >{\raggedright\arraybackslash}p{0.18\linewidth}
    }
    \toprule
    \textbf{Code} & \textbf{Definition} & \textbf{Inclusion Examples} & \textbf{Exclusion Criteria} \\ 
    \midrule
    
    Independent research &
    Participant performs research without using an LLM: web search, textbooks, papers, personal knowledge. &
    ``Googled 'GFF3 format' and read the EMBL tutorial.'' ; ``Checked PubMed for influenza NCR papers.''; ``Used my lab notebook from last semester.'' &
    Any LLM interaction (LLM research, LLM ideation support, etc.). \\
    
    Independent explanation &
    Participant articulates their own reasoning or justification, showing how they arrived at a step or decision. &
    ``I chose 30 bp overlaps because longer regions increase Gibson efficiency.''; ``Picked HindIII since the insert lacks its site.'' &
    Vague ``I looked it up.'' Protocol quotes without causal reasoning. \\
    
    Protocol lookup &
    Participant consults an external formal protocol (paper, kit manual, online SOP) to guide their actions. &
    ``Followed Addgene's Gibson Assembly PDF.'' ; ``Used NEB's HindIII digest table.''; ``Pulled the CDC RT-qPCR protocol.'' &
    Generic browsing or self-research without a named protocol. \\
    
    Resource listing &
    Participant enumerates materials, tools, or sources used. &
    ``Google Scholar, NCBI, ChatGPT.'' ; ``Used SnapGene plus a Sonnet model.''; ``Consulted three papers and an R script.'' &
    Rich explanatory content; mere mention of one resource inside a longer narrative. \\
    
    Mechanism explanation &
    Participant explains how or why a biological or biochemical process works at the mechanistic level. &
    ``Taq polymerase's 5' to 3' activity adds A-overhangs enabling TA-cloning.''; ``RNA-dependent RNA polymerase binds conserved NCRs to initiate transcription.'' &
    Simple protocol steps, high-level logic, or outcome justification without mechanistic detail. \\
    
    \bottomrule
    \end{tabular}
\end{table}

\clearpage
\subsubsection{Planning \& Quality Codes}

\begin{table}[ht]
    \centering
    \begin{tabular}{
      >{\raggedright\arraybackslash}p{0.15\linewidth}
      >{\raggedright\arraybackslash}p{0.27\linewidth}
      >{\raggedright\arraybackslash}p{0.30\linewidth}
      >{\raggedright\arraybackslash}p{0.18\linewidth}
    }
    \toprule
    \textbf{Code} & \textbf{Definition} & \textbf{Inclusion Examples} & \textbf{Exclusion Criteria} \\ 
    \midrule
    
    Proposal planning &
    Participant lays out a forward-looking experimental or methodological plan with concrete next steps, resources, or timeline. &
    ``Split 5 kb insert into four PCR products, assemble with Gibson, transform DH5$\alpha$, screen on kan.'' ; ``Day 1: grow culture; Day 2: miniprep; Day 3: sequencing.'' &
    Retrospective descriptions of completed steps; vague ideas without actionable next steps. \\
    
    Major error correction &
    Participant identifies and corrects a conceptual or procedural flaw that would materially change the outcome. &
    ``Realised I used EcoRI instead of EcoRV; switched enzymes.'' ; ``Fixed the fragment orientation---reverse-complemented segment 4.'' ; ``Changed incubation from 4 h to 60 min.'' &
    Minor typos or unit fixes (Minor error correction). \\
    
    Minor error correction &
    Participant fixes small, non-conceptual mistakes (formatting, units, typos) that don't substantially alter the biology. &
    ``Corrected 'ml' to '$\mu$L'.'' ; ``Removed an extra space in the FASTA header.'' ; ``Changed 10$\mu$ to 1$\mu$ as per recipe.'' &
    Switching enzymes or re-ordering major steps (see Major error correction). \\
    
    Lossy summary &
    Participant summarizes information but omits important details, introduces inaccuracies, or overly compresses the source. &
    ``According to the review, the virus binds ACE2 [omits strain-specific affinity].'' ; ``Protocol says 'incubate overnight' [original: 16 h at 30°C with shaking].'' &
    Faithful or complete summaries that retain all key details. \\
    
    \bottomrule
    \end{tabular}
\end{table}

\clearpage
\subsubsection{Affective Codes}

\begin{table}[ht]
    \centering
    \begin{tabular}{
      >{\raggedright\arraybackslash}p{0.15\linewidth}
      >{\raggedright\arraybackslash}p{0.27\linewidth}
      >{\raggedright\arraybackslash}p{0.30\linewidth}
      >{\raggedright\arraybackslash}p{0.18\linewidth}
    }
    \toprule
    \textbf{Code} & \textbf{Definition} & \textbf{Inclusion Examples} & \textbf{Exclusion Criteria} \\ 
    \midrule
    
    Confidence &
    Explicit positive confidence or ease. &
    ``Felt confident after double-checking.'' ; ``Pretty easy once I saw the table.''; ``Smooth process overall.'' &
    Neutral success statements (``it worked''). \\
    
    Confusion &
    Expresses being confused, uncertain, or stuck. &
    ``I'm not sure what this assay measures.'' ; ``Got stuck on segment 6 orientation.''; ``This part confused me.'' &
    General remarks that prompt was ambiguous without emotional tone. \\
    
    Frustration &
    Voices irritation or annoyance. &
    ``This was frustrating; nothing aligned.'' ; ``Got annoyed when the digest failed again.''; ``Wasted two hours on buggy software.'' &
    Neutral difficulty statements (``challenging but okay''). \\
    
    Gratitude &
    Expresses gratitude, relief, or appreciation. &
    ``Thank goodness for the model's suggestions.'' ; ``Grateful that NEB has clear docs.''; ``Really appreciate the LLM's explanation.'' &
    Generic praise without emotional tone (``good output''). \\
    
    Overwhelm &
    Feels overwhelmed or overloaded. &
    ``Too much information to process.'' ; ``Completely overwhelmed by the protocol length.''; ``Couldn't keep track of all the references.'' &
    Simple ``complex'' or ``detailed'' without explicit overwhelm. \\
    
    \bottomrule
    \end{tabular}
\end{table}

\clearpage
\subsection{Domain Term Glossary} \label{app:term-glossary}

\begingroup
\setlength{\columnsep}{0.8em}
\setlist[itemize]{leftmargin=*,topsep=0.5pt,partopsep=0.5pt,parsep=0.5pt,itemsep=0.15em}

\noindent\textbf{Virus families and related taxa}
\begin{multicols}{3}
\begin{itemize}
  \item Adenoviridae
  \item Arenaviridae
  \item Astroviridae
  \item Baculoviridae
  \item Bornaviridae
  \item Bunyaviridae
  \item Caliciviridae
  \item Coronaviridae
  \item Filoviridae
  \item Flaviviridae
  \item Hantaviridae
  \item Hepadnaviridae
  \item Herpesviridae
  \item Iridoviridae
  \item Nairoviridae
  \item Orthomyxoviridae
  \item Papillomaviridae
  \item Paramyxoviridae
  \item Parvoviridae
  \item Picornaviridae
  \item Polyomaviridae
  \item Poxviridae
  \item Retroviridae
  \item Reoviridae
  \item Rhabdoviridae
  \item Secoviridae
  \item Togaviridae
  \item Virgaviridae
  \item Pneumoviridae
  \item Arteriviridae
  \item Adeno-associated virus
  \item Lentivirus
\end{itemize}
\end{multicols}

\noindent\textbf{Virology assays, biosafety, and cell culture}
\begin{multicols}{3}
\begin{itemize}
  \item Plaque assay
  \item TCID50
  \item PFU
  \item MOI
  \item MDCK
  \item TPCK-treated trypsin
  \item Agarose overlay
  \item Confluency
  \item Hemagglutination
  \item Cytopathic effect
  \item Serial dilution
  \item Passage
  \item Biosafety cabinet
  \item Laminar flow hood
  \item BSL-3
  \item BSL-4
  \item qRT-PCR
  \item Reverse transcription
  \item RNA extraction
  \item Viral tropism
  \item Envelope protein
  \item Capsid
  \item Viral entry
  \item Transfection
  \item Electroporation
  \item Viral vector
  \item Spike protein
  \item Nucleocapsid
  \item dsRNA
  \item ssRNA
  \item Positive-sense RNA
  \item Negative-sense RNA
\end{itemize}
\end{multicols}

\noindent\textbf{Molecular cloning and DNA assembly}
\begin{multicols}{3}
\begin{itemize}
  \item Gibson Assembly
  \item Overhang
  \item Homology arm
  \item Exonuclease
  \item Phusion polymerase
  \item Isothermal assembly
  \item Fragment
  \item Codon optimization
  \item Gene synthesis
  \item DNA assembly
  \item Restriction enzyme
  \item PCR
  \item Ligation
  \item Golden Gate
  \item Cloning
  \item Vector
  \item Plasmid
  \item Ampicillin resistance
  \item \textit{E.\,coli} DH5$\alpha$
  \item Promoter
  \item Ribosome binding site
  \item Terminator
\end{itemize}
\end{multicols}

\noindent\textbf{Representation learning and NLP}
\begin{multicols}{3}
\begin{itemize}
  \item Embedding
  \item Tokenization
  \item Compression
  \item Latent space
  \item Autoencoder
  \item Bottleneck
  \item Representation
  \item Transformer
  \item Attention
\end{itemize}
\end{multicols}

\noindent\textbf{Protein structure and bioinformatics}
\begin{multicols}{3}
\begin{itemize}
  \item Protein folding
  \item ESMFold
  \item Contact map
  \item AlphaFold
  \item RMSD
  \item TM-score
  \item Sequence alignment
  \item Multiple sequence alignment
  \item Residue
  \item Backbone
  \item All-atom
  \item Secondary structure
  \item Helix
  \item Sheet
  \item Loop
  \item Side chain
  \item Conformation
  \item Rotamer
\end{itemize}
\end{multicols}

\noindent\textbf{General lab practice and statistics}
\begin{multicols}{3}
\begin{itemize}
  \item Cell culture
  \item Incubation
  \item Centrifugation
  \item Spectrophotometer
  \item Microscopy
  \item Buffer
  \item Phosphate-buffered saline
  \item Temperature
  \item pH
  \item Concentration
  \item Protocol
  \item Sample
  \item Control
  \item Replicate
  \item Standard deviation
  \item Mean
  \item Hypothesis
  \item Statistical significance
\end{itemize}
\end{multicols}

\noindent\textbf{Benchmarks, datasets, and test names}
\begin{multicols}{2}
\begin{itemize}
  \item Virology Capabilities Test
  \item Hourglass Protein Compression Transformer
  \item CHEAP embeddings
  \item Molecular Biology Capabilities Test
  \item World Class Biology
  \item Long-Form Virology
\end{itemize}
\end{multicols}
\endgroup

\end{document}